\documentclass{article}
\usepackage[square,sort,comma,numbers]{natbib} 

\usepackage[final]{neurips_2024}

\usepackage[utf8]{inputenc} 
\usepackage[T1]{fontenc}    
\usepackage{hyperref}       
\usepackage{url}            
\usepackage{booktabs}       
\usepackage{amsfonts}       
\usepackage{nicefrac}       
\usepackage{microtype}      
\usepackage{xcolor}         

\usepackage{microtype}
\usepackage{graphicx}
\usepackage{subfig}
\usepackage{booktabs} 
\usepackage{hyperref}
\usepackage{algorithm}
\usepackage{algpseudocode}
\usepackage{colortbl}
\usepackage{wrapfig}
\usepackage{float}
\usepackage{subfig}
\usepackage{lipsum}
\usepackage{multicol}
\usepackage{multirow}
\usepackage{amsmath}
\usepackage{amssymb}
\usepackage{mathtools}
\usepackage{amsthm}

\usepackage{amsmath,amsfonts,bm}

\def\eqref#1{equation~\ref{#1}}

\def\1{\bm{1}}

\DeclareMathAlphabet{\mathsfit}{\encodingdefault}{\sfdefault}{m}{sl}
\SetMathAlphabet{\mathsfit}{bold}{\encodingdefault}{\sfdefault}{bx}{n}

\newcommand{\E}{\mathbb{E}}

\usepackage{bm}
\usepackage{url}
\usepackage{enumitem}
\def\E{\mathbb{E}}
\def\H{\mathbf{H}}
\def\barH{\bar{\H}}
\def\D{\mathcal{D}}
\def\F{\mathbf{F}}

\def\A{\mathbf{A}}

\def\B{\mathbf{B}}
\def\W{\mathbf{W}}

\def\xlikeli{p(s|\bm{\theta})}
\def\samples{s\in\D}

\def\vec{\mathrm{vec}}
\def\Si{\mathbf{\Sigma}}
\def\I{\mathbf{I}}

\def\M{\mathbf{M}}
\def\Z{\mathbf{Z}}

\title{FedLPA: One-shot Federated Learning with Layer-Wise Posterior Aggregation}

\author{Xiang Liu$^{1,\dag}$, Liangxi Liu$^{2,\dag}$, Feiyang Ye$^3$, Yunheng Shen$^4$, Xia Li$^5$, Linshan Jiang$^{1,\ddag}$, Jialin Li$^{1,\ddag}$\\ 
$^1$National University of Singapore, 
$^2$Northeastern University, \\
$^3$University of Technology Sydney,
$^4$Tsinghua University,
$^5$ETH Zurich\\
\small{\texttt{\{liuxiang,lijl\}@comp.nus.edu.sg} \qquad
\texttt{liu.liangx@northeastern.edu}\qquad
\texttt{feiyang.ye.uts@gmail.com}} \\
\small{\texttt{shenyh19@mails.tsinghua.edu.cn}\qquad
\texttt{ethlixia@gmail.com} \qquad
\texttt{linshan@nus.edu.sg}}\\
$^\dag$\textit{Equal Contribution}\\
$^\ddag$\textit{Correspondence Author}
}

\begin{document}

\maketitle

\begin{abstract}
Efficiently aggregating trained neural networks from local clients into a global model on a server is a widely researched topic in federated learning. Recently, motivated by diminishing privacy concerns, mitigating potential attacks, and reducing communication overhead, one-shot federated learning (i.e., limiting client-server communication into a single round) has gained popularity among researchers. However, the one-shot aggregation performances are sensitively affected by the non-identical training data distribution, which exhibits high statistical heterogeneity in some real-world scenarios. To address this issue, we propose a novel one-shot aggregation method with layer-wise posterior aggregation, named FedLPA. FedLPA aggregates local models to obtain a more accurate global model without requiring extra auxiliary datasets or exposing any private label information, e.g., label distributions. To effectively capture the statistics maintained in the biased local datasets in the practical non-IID scenario, we efficiently infer the posteriors of each layer in each local model using layer-wise Laplace approximation and aggregate them to train the global parameters. Extensive experimental results demonstrate that FedLPA significantly improves learning performance over state-of-the-art methods across several metrics. 
\end{abstract}

\section{Introduction}
Data privacy issues in Deep Learning~\cite{dl0, dl1,ml2, ml1, ml3,  dl2,dls} have grown to be a major global concern~\cite{yang2019federated}. To safeguard data privacy, the conventional federated learning algorithm will use the aggregation methods and follow the data management rules of different institutions, which implies that the distribution of data exhibits variations among clients~\cite{yang2019federated}. In the domain of machine learning, federated learning (FL)~\cite{fedavg1, kairouz2021advances, li2021survey} has emerged as a prominent paradigm. The fundamental tenet of federated learning revolves around sharing machine learning models derived from decentralized data repositories, as opposed to divulging user raw data. This approach effectively preserves the confidentiality of individual data.

The standard federated learning framework, FedAvg~\cite{fedavg1, fedavg2}, applies local model training. These local models are then aggregated into a global model through parameter averaging. Existing FL algorithms, however, require many communication rounds to effectively train a global model, leading to substantial communication overhead, increased privacy concerns, and higher demand for fault tolerance throughout the rounds. One-shot FL, which reduces client-server communication into a single round as explored by prior work~\cite{guha2019one,li2020practical,diao2023towards}, is a promising yet challenging scheme to address these issues. One-shot FL proves particularly practical in scenarios where iterative communication is not feasible. Moreover, a reduction in communication rounds translates to fewer opportunities for any potential eavesdropping attacks.

While one-shot FL shows promises, existing approaches often grapple with challenges such as inadequate handling of high statistical heterogeneity information~\cite{al2020federated,liu2023bayesian} or non-independent and non-identically distributed (non-IID) data~\cite{zhou2020distilled,zhang2021practical}. Moreover, some prior methods rely on an auxiliary public dataset to achieve satisfactory performance in one-shot FL~\cite{guha2019one,li2020practical}, or even on pre-trained large models~\cite{yang2023exploring}, which may not be practical~\cite{zhu2021data} in some sensitive scenarios. Additionally, some approaches, such as those~\cite{shin2020xor,zhang2021practical,heinbaugh2022data,diao2023towards}), might expose private label information to both local and global models, e.g., the client label distribution, potentially violating General Data Protection Regulation (GDPR) rules. Furthermore, some prior methods~\cite{li2020practical,zhou2020distilled,zhang2022dense} require substantial computing resources for dataset distillation, model distillation, or even training a generator capable of generating synthetic data for second-stage training on the server side, making them less practical.

Besides, the performance of one-shot FL often falls short when dealing with non-IID data. Non-IID data biases global updates, reducing the accuracy of the global model and slowing down convergence. In extreme non-IID cases, clients may be required to address distinct classes solely on their side. Several approaches to federated learning are proposed in multi-round settings to tackle this heterogeneity among clients. In the work~\cite{smith2017federated}, it allows each client to use a personalized model instead of a shared global model. With the personalized approach, a multi-round framework benefits from joint training while allowing each client to keep its unique model. However, one-shot aggregation on a local model is far from being resolved to address the concern of non-i.i.d data distributions.

In this paper, we introduce a novel one-shot aggregation approach to address these issues, named FedLPA (Federated Learning with Layer-wise Posterior Aggregation). FedLPA infers the posteriors of each layer in each local model using the empirical Fisher information matrix obtained by layer-wise Laplace Approximation. Laplace Approximations are widely used to compute the empirical Fisher information matrix for neural networks, conveying the data statistics in non-i.i.d settings. However, computing empirical Fisher information matrices of multiple local clients and aggregating their Fisher information matrices remains an ongoing challenge~\cite{liu2023bayesian}. To mitigate it,  FedLPA aggregates the posteriors of local models using the accurately computed block-diagonal empirical Fisher information matrices to measure the parameter space. This matrix captures essential parameter correlations and distinguishes itself from prior methods by being non-diagonal and non-low-rank, thereby conveying the statistics of biased local datasets. After that, the global model parameters are aggregated without any need for server-side knowledge distillation~\cite{lin2020ensemble}.

Our extensive experiments verify the efficiency and effectiveness of FedLPA, highlighting that FedLPA markedly enhances the test accuracy when compared to existing one-shot FL baseline approaches across various datasets. Our main contributions are summarized as follows:

\begin{itemize}
\item To the best of our knowledge, we are the first to propose an effective one-shot federated learning approach that trains global models using block-diagonal empirical Fisher information matrices. Our approach is data-free without any need for any auxiliary dataset and label information and significantly improves system performance, including negligible communication cost and moderate computing overhead. 
\item We are the first to train global model parameters via constructing a multi-variate linear objective function and optimizing its quadratic form, which allows us to formulate and solve the problem in a convex form efficiently, which has a linear convergence rate, ensuring good performance.
\item We conduct extensive experiments to illustrate the effectiveness of FedLPA. Our approach consistently outperforms the baselines, showcasing substantial improvement across various settings and datasets. Even in some extreme scenarios where label skew is severe, e.g., each client has only one class, we achieve satisfactory results while other existing one-shot federated learning algorithms struggle.
\end{itemize}

\vspace{-.5em}

\section{Background and related works}
\subsection{Federated learning on non-iid data}
Previous work FedAvg~\cite{fedavg1} first introduced the concept of FL and presented the algorithm, which\\

achieved competitive performance on i.i.d data, in comparison to several centralized techniques. However, it was observed in previous works~\cite{zhao2018federated,li2019convergence} that the convergence rate and ultimate accuracy of FedAvg on non-IID data distributions were significantly reduced, compared to the results observed with homogeneous data distributions.

Several methods have been developed to enhance performance in federated learning against non-IID data distributions. The SCAFFOLD method~\cite{scaffold} leveraged control variates to reduce objective inconsistency in local updates. It estimated the drift of directions in local optimization and global optimization and incorporated this drift into local training to align the local optimization direction with the global optimization. FedNova~\cite{fednova} addressed objective inconsistency while maintaining rapid error convergence through a normalized averaging method. It scaled and normalized the local updates of each client based on the number of local optimization steps. FedProx~\cite{fedprox} enhanced the local training process by introducing a global prior in the form of an $L2$ regularization term within the local objective function. Researchers introduced PFNM~\cite{yurochkin2019bayesian,wang2020federated}, a Bayesian probabilistic framework specifically tailored for multilayer perceptrons. PFNM employed a Beta-Bernoulli process (BBP)~\cite{BBP} to aggregate local models, quantifying the degree of alignment between global and local parameters. The framework~\cite{liu2023bayesian} proposed utilized a multivariate Gaussian product method to construct a global posterior by aggregating local posteriors estimated using an online Laplace approximation. FedPA~\cite{al2020federated} also applied the Gaussian product method but employed stochastic gradient Markov chain Monte Carlo for approximate inference of local posteriors. DAFL (Data-Free Learning)~\cite{chen2019data} introduced an innovative framework based on generative adversarial networks. ADI~\cite{yin2020dreaming} utilized an image synthesis method that leveraged the image distribution to train deep neural networks without real data. The pFedHN method~\cite{shamsian2021personalized} incorporated HyperNetworks~\cite{krueger2017bayesian} to address federated learning applications.

However, all of these methods encountered challenges in the one-shot federated learning setting, as they required aggregating the model by multiple rounds and might be inaccurate due to the omission of critical information, such as posterior joint probabilities between different parameters.

\subsection{One-shot federated learning}
One-shot Federated Learning (FL) is an emerging and promising research direction characterized by its minimal communication cost. In the first study on one-shot FL~\cite{guha2019one}, the approach involved the aggregation of local models, forming an ensemble to construct the final global model. Subsequently, knowledge distillation using public data was applied in the following step. FedKT~\cite{li2020practical} brought forward the concept of consistent voting to fortify the ensemble. Recent research endeavors~\cite{zhang2021practical, zhang2022dense} proposed data-free knowledge distillation schemes tailored for one-shot FL. These methods adopted the basic ensemble distillation framework as FedDF~\cite{lin2020ensemble}. XorMixFL~\cite{shin2020xor} introduced the use of exclusive OR operation (XOR) for encoding and decoding samples in data sharing. It is important to note that XorMixFL assumed the possession of labeled samples from a global class by all clients and the server, which might not align with practical real-world scenarios. A noteworthy innovation of DENSE~\cite{zhang2022dense} was its utilization of a generator to create synthetic datasets on the server side, circumventing the need for a public dataset in the distillation process. Co-Boosting~\cite{dai2024enhancing} improves the ensemble when doing the distillation to improve the performance. FedOV~\cite{diao2023towards} delved into addressing comprehensive label skew cases. FEDCVAE~\cite{heinbaugh2022data} confronted this challenge by transmitting all label distributions from clients to servers. These schemes~\cite{shin2020xor,li2020practical,zhang2021practical,zhang2022dense,heinbaugh2022data,diao2023towards} exposed some client-side private information, leading to additional communication overhead and potential privacy leakage, e.g., FEDCVAE~\cite{heinbaugh2022data} needed all the client label distribution to be transmitted to the server side and FedOV~\cite{diao2023towards} needed the clients to know the labels which were unknown. Instead, MA-Echo~\cite{su2023one} adopted a unique approach by emphasizing the addition of norms among layer-wide parameters during the aggregation of local models. The project~\cite{jhunjhunwala2024fedfisher} focused on the theoretic analysis of the error in its approximation method. However, their method grappled with limited experiments and lacked detailed explanations of the approach. FedDISC~\cite{yang2023exploring}, on the other hand, relied on the pre-trained model CLIP from OpenAI, where their reliance might not always align with practicality or suitability for diverse scenarios.

While some of these techniques are orthogonal to FedLPA and can be integrated with it, it is worth noting that none of the previously mentioned algorithms possess the capability to train global model parameters using empirical Fisher information matrices on extensive experiment settings. Some of them~\cite{guha2019one,li2020practical} may require additional information, and may potentially entail the risk of label distribution leakages.

\section{Methodology}
\subsection{Objective formulation}
Generally, federated learning is defined as a optimization problem~\cite{fedprox,scaffold,fednova,fl8} for maximizing a global objective function $\mathbb{F(\bm{\theta})}$ which is a mixture of local objective functions $\mathbb{F}_k(\bm{\theta}, \D_k)$:
\begin{equation}
    \mathbb{F(\bm{\theta})} = \sum_{k=1}^K \mathbb{F}_k(\bm{\theta}, \D_k)
    \label{EQ:flgoal}
\end{equation}
where  $\bm{\theta} = [\vec(\W_{1}), \dots,\vec(\W_{l}), \dots,\vec(\W_{L})]$ is the parameter vector of global model and $\W_{l}$ is the weight and bias of layer $l$ for a $L$-layers neural network; $\D_k$ is the local dataset $k$-th client. $\mathbb{F}_k(\bm{\theta}, \D_k)$ is the expectation of the local objective function, which is proportional to the logarithm of likelihood $\log p(\D_k | \bm{\theta})$.

Previous works~\cite{al2020federated,liu2023bayesian} give a common formula of the global posterior which consists of local posteriors $p(\bm{\theta} | \D_k)$ under variational inference formulation. 


\begin{equation}
    p(\bm{\theta}|\D) \propto \prod_{k=1}^K p(\D_k | \bm{\theta}) \propto \prod_{k=1}^K p(\bm{\theta}|\D_k)
    \label{EQ:gp2}
\end{equation}

\begin{equation}
    \begin{aligned}
        \max_{\bm{\theta}}  \mathbb{F(\bm{\theta})} 
        = \sum_{k=1}^K \frac{|D_k|}{|\D|} \cdot \E_{s \in \D_k} \left[ \log p(s | \bm{\theta}) \right]
         \equiv  \max_{\bm{\theta}} \prod_{k=1}^K p(\bm{\theta}|\D_k)
    \end{aligned}
    \label{EQ.fleq}
\end{equation}
As we know, the objective function is the expectation of the likelihood, and the sum of the logarithms is equal to the logarithms of the product as Eq. \ref{EQ.fleq}. Therefore, globally variational inference using Eq. \ref{EQ:gp2} is equivalent to optimization for Eq. \ref{EQ:flgoal}. Correspondingly, we have:

\begin{equation}
        \max_{\bm{\theta}} \mathbb{F}_k(\bm{\theta}, \D_k) \equiv  \max_{\bm{\theta}} p(\bm{\theta}|\D_k)
    \label{EQ:fleql}
\end{equation}
Following the same training pattern of federated learning, each client infers the local posterior $p(\bm{\theta}|\D_k)$ by using the local dataset $\D_k$. As a result, the server obtains the global posterior $p(\bm{\theta}|\D)$ by aggregating local posteriors using Eq. \ref{EQ:gp2}.

However, both the global and local posterior are usually intractable because modern neural networks are usually non-linear and have a large number of parameters.
Therefore, it is necessary to design an efficient and accurate aggregation method for one-shot federated learning.

\subsection{Approximating posteriors}

Although the posterior is usually intractable, the posterior can be approximated as a Gaussian distribution by performing a Taylor expansion on the logarithm of the posterior~\cite{ritter2018scalable}:
\begin{equation}
    \log p(\bm{\theta} | \D) \approx \log p\left(\bm{\theta}^{*} | \D\right)-\frac{1}{2}\left(\bm{\theta}-\bm{\theta}^{*}\right)^{\top} \barH\left(\bm{\theta}-\bm{\theta}^{*}\right)
    \label{EQ:Taylor}
\end{equation}
where $\bm{\theta}^{*}$ is the optimal parameter vector, $\barH = \mathbb{E}_{\samples}[\mathbf{H}]$ is the average Hessian of the negative log posterior over a dataset $\D$. It is reasonable to approximate global and local posteriors as multi-variates Gaussian distributions with expectations $\bar{\bm{\mu}} = \bm{\theta}^*$ and $\bm{\mu_k} = \bm{\theta}_k^*$; co-variances $\bar{\Si}=\barH^{-1}$ and  $\Si_k=\barH_k^{-1}$~\cite{daxberger2021laplace}. 

\begin{equation}
    \begin{aligned}
        p(\bm{\theta}|\D) &\equiv \bm{\theta} \sim \mathcal{N}(\bar{\bm{\mu}}, \bar{\Si}), 
        p(\bm{\theta}|\D_k) &\equiv \bm{\theta} \sim \mathcal{N}(\bm{\mu_k}, \Si_k)
    \end{aligned}
    \label{EQ:laplace}
\end{equation}

As a result, if given local expectation $\bm{\mu}_k$ and local co-variance $\Si_k$, the global posterior is determined by Eq. \ref{EQ:gp2} as below: 
\begin{equation}
    \bar{\bm{\mu}} = \bar{\mathbf{\Sigma}} \sum_k^K \mathbf{\Sigma}_k^{-1} \bm{\mu}_k, 
    \bar{\mathbf{\Sigma}}^{-1} = \sum_k^K \mathbf{\Sigma}_k^{-1}
    \label{EQ:Framework}
\end{equation}
Modern algorithms~\cite{bp1,KFAC1} allow the local training process to obtain an optimal, regarded as the expectation $\bm{\mu}_k$ in the above equations. However, $\barH_k$ is intractable to compute due to a large number of parameters in modern neural networks. An efficient method is to approximate $\barH_k$ using the empirical Fisher information matrix~\cite{KP-1}.

\subsection{Inferring the local layer-wise posteriors with the block-diagonal empirical Fisher information matrices}
A empirical Fisher $\tilde{\F}$ is defined as below:
\begin{equation}
    \tilde{\F}= \sum_{s \in \D}\left[\nabla \log \xlikeli \nabla \log \xlikeli^{\top}\right]
    \label{EQ:FIM}
\end{equation}
where $\xlikeli$ is the likelihood on data point $s$. It is an approximate of the Fisher information matrix, the empirical Fisher information matrix is equivalent to the expectation of the Hessian of the negative log posterior if assuming $\xlikeli$ is identical for each $s \in \D$.

Therefore, the local co-variance $\Si_k$ can be approximated by the empirical Fisher $\tilde{\F}_k$~\cite{martens2015optimizing,KFAC2}.

\begin{equation}
    \Si_k^{-1} \approx \tilde{\F}_k+\lambda \I
    \label{EQ:SigmaFIM}
\end{equation}

The works~\cite{EWC,matena2022merging,liu2023bayesian} ignore co-relations between different parameters and only consider the self-relations of parameters as computing all co-relations is impossible. Thus, their methods are inaccurate. Detailed discussions and the novelty compared to previous works are in Appendix~\ref{appendix:related}.

In order to capture co-relations between different parameters efficiently, previous works~\cite{KFAC1,ritter2018scalable} estimate a block empirical Fisher information matrix $\F$ instead of assuming parameters are independent and approximating the co-variance by the diagonal of the empirical Fisher. As pointed out, co-relations inner a layer are much more significant than others~\cite{KFAC1,benzing2022gradient,zhang2022scalable}, while computing the co-relations between different layers brings slight improvement but much more computation~\cite{martens2016second,ritter2018scalable}. Therefore, assuming parameters are layer-independent is a good trade-off. As a result, the approximated layer-wise empirical Fisher is block-diagonal. For layer $l$ on client $k$, its empirical Fisher $\F_{k_l}$ is one of the diagonal blocks in the whole empirical Fisher for the local model and is factored into two small matrices as below,


\begin{equation}
   \Si^{-1}_{k_l} \approx \F_{k_l} = \A_{k_l} \otimes \B_{k_l}
    \label{EQ:blockFIM}
\end{equation}
where $\otimes$ is the Kronecker product; $\A_{k_l} = \mathbb{E} \left[ \hat{\mathbf{a}}_{k_{l-1}} \hat{\mathbf{a}}_{k_{l-1}}^{\top}\right]+\pi_{l} \sqrt{\lambda} \I$ and $\B_{k_l} = \mathbb{E} \left[\hat{\mathbf{b}}_{k_l} \hat{\mathbf{b}}_{k_l}^{\top}\right] + \frac{1}{\pi_{l}} \sqrt{\lambda} \I$ are two expectation factor  matrices over the data samples; $\hat{\mathbf{a}}_{k_l}$ is the activations and $\hat{\mathbf{b}}_{k_l}$ is the gradient of the pre-activations of layer $l$ on client $k$, $\lambda$ is the hyperparameter and $\pi_l$ is a factor minimizing approximation error in $\F_{k_l}$~\cite{KFAC1,KFAC2,KFRA}. $\A_{k_l}$ and $\B_{k_l}$ are symmetric positive definite matrices~\cite{bp1,KFAC1}. 

We use $\bm{\theta}_{k_l}$ to denote the parameter vector of layer $l$ and $\M_{k_l}=vec^{-1}(\bm{\mu}_{k_l})$ is the vectorized optimal weight matrix of layer $l$ on client $k$. Thus, the resulting local layer-wise posterior approximation is $ \bm{\theta}_{k_l} \sim \mathcal{N}(\bm{\mu}_{k_l}, \F_{k_l}^{-1})$.

\subsection{Estimating the global expectation}
Given the local posteriors, the global expectation could be aggregated by Eq. \ref{EQ:Framework}. With Eq. \ref{EQ:blockFIM}, the $l$-th layer's global expectation $\bar{\bm{\mu}}_l$ consists of Kronecker products:    
\begin{equation}
    \begin{aligned}
        \bar{\bm{\mu}}_l &=\bar{\Si}_l \sum_k^K \Si_{k_l}^{-1} \bm{\mu}_{k_l} 
        = \bar{\Si}_l \sum_k^{K}{ (\A_{k_l} \otimes \B_{k_l}) \bm{\mu}_{k_l}} \\
        &= \bar{\Si}_l \sum_k^{K} \vec(\B_{k_l} \M_{k_l} \A_{k_l}) 
        = \bar{\Si}_l \sum_k^{K}{ \mathbf{z}_{k_l}} = \bar{\Si}_l \bar{\mathbf{z}}_l
    \end{aligned}
    \label{eq:gmk}
\end{equation}

where $\bar{\mathbf{z}}_l =  \sum_k^{K}{\mathbf{z}_{k_l}}$ and $\mathbf{z}_{k_l} = \vec(\B_{k_l} \M_{k_l} \A_{k_l})$ is a immediate notations for simplification. For the global expectation, we have $\bar{\bm{\mu}} = \bar{\mathbf{\Sigma}} \cdot \bar{\mathbf{z}}$. The corresponding global co-variance is an inverse of the sum of Kronecker products:
\begin{equation}
    \bar{\Si}_l = (\sum_k^{K}{ \A_{k_l} \otimes \B_{k_l}})^{-1}
    \label{eq:gsc}
\end{equation}
As shown in Eq. \ref{eq:gmk}, obtaining the global expectation $\bar{\bm{\mu}}_l$ requires calculating the inverse of $\bar{\Si}_l^{-1}$ as Eq. \ref{eq:gsc}, which is unacceptable and the details are in Appendix~\ref{append:unacceptable}. Thus, we propose our method to directly train the parameters of the global model on the server side.

\subsection{Train the parameters of the global model}
We use $\mathbb{E} \left[ \A \right]$ denotes $\sum_k^{K}(\A_{k})$, $\mathbb{E} \left[ \B \right]$ denotes $\sum_k^{K}(\B_{k})$, $\mathbb{E} \left[ \A \otimes \B \right]$ denotes $\sum_k^{K}(\A_{k} \otimes \B_{k})$. Previous works~\cite{KFAC1, KFAC2} approximate the expectation of Kronecker products by a Kronecker product of expectations $\mathbb{E} \left[ \A \otimes \B \right] \approx \mathbb{E} \left[ \A \right]\otimes \mathbb{E} \left[ \B \right]$ with an assumption of $\A_{k_l}$ and $\B_{k_l}$ are independent, which is called Expectation Approximation (EA). However, it may lead to a biased global expectation. The details are discussed in Appendix~\ref{append:EA}.
Instead, we could construct a linear objective after aggregating the approximation of local posteriors via using block-diagonal empirical Fisher information matrices. We denotes $\bar{\M}$ as the matrix formula of $\bar{\bm{\mu}} = \vec(\bar{\M})$, and the optimal solution of $f(\bar{\bm{\mu}})$ is $\bar{\bm{\mu}}^* = \vec(\bar{\M}^*)$. We construct $f(\bar{\bm{\mu}})$ as a multi-variates linear objective function. When $\bar{\bm{\mu}}=\bar{\bm{\mu}}^*$ is optimal solution, $f(\bar{\bm{\mu}})=\mathbf{o}$, where $\mathbf{o}$ is a vector with all zero. Note that 
\begin{equation}
    \begin{aligned}
        f(\bar{\bm{\mu}}) &= \bar{\mathbf{\Sigma}}^{-1} \bar{\bm{\mu}} - \bar{\mathbf{z}}
        = \sum_k^K \vec(\mathbf{B}_k \bar{\mathbf{M}} \mathbf{A}_k) - \bar{\mathbf{z}} \\
        &= \vec(\mathbb{E}\left[ \mathbf{B} \bar{\mathbf{M}} \mathbf{A}\right]) - \bar{\mathbf{z}}
    \end{aligned}
    \label{EQ:MU:obj}
\end{equation}
To obtain the optimal solution, we minimize the following problem to obtain an approximate solution $\bar{\mathbf{M}}^*$ of $\bar{\mathbf{M}}$:
\begin{equation}
    \begin{aligned}
        \bar{\mathbf{M}}^* &=  \min_{\bar{\mathbf{M}}} \frac{1}{2} \left\|\sum_k^K \vec(\mathbf{B}_k \bar{\mathbf{M}} \mathbf{A}_k) - \bar{\mathbf{z}}\right\|_2^2
    \end{aligned}
    \label{EQ:MU:obj:solution}
\end{equation}

The above equation is a quadratic objective, and it can be solved by modern optimization tools efficiently and conveniently. Since the main objective of the above problem is both convex and  Lipschitz smooth w.r.t $\vec(\Bar{\mathbf{M}})$, we can use the gradient descent method to solve it with a linear convergence rate. Here, we use automatic differentiation to calculate the gradient w.r.t. $\Bar{\mathbf{M}}$.


 
\begin{algorithm}[]
    \caption{FedLPA Global Aggregation}
    \label{algorithm:fedlpa}
    \begin{multicols}{2}
    \begin{algorithmic}[1]
    \State {\bfseries Input:} clients $K$,  layers $L$
    \State Initialize global weight $\bar{\W}_l$ of layer $l = 1, ..., L$
    \State {\bfseries{clients executes:}}
    \State Initialize local model
    \For{k = 1, ..., K}
        \State $\{\M_{k_l}, \A_{k_l}, \B_{k_l} | l = 1, ..., L\}$ $\leftarrow $ local training
    \EndFor
    \State {\bfseries{Server executes:}}
    \For{l = 1, ..., L}
        \State $\bar{\A}_l \leftarrow \sum_k^K \A_{k_l}$
        \State $\bar{\B}_l \leftarrow \sum_k^K \B_{k_l}$
        \State $\bar{\Z}_l \leftarrow \sum_k^K \B_{k_l} \M_{k_l} \A_{k_l}$
        \State $\bar{\M}_l \leftarrow $ Train the parameter of the global model
        \State $\bar{\W}_l \leftarrow \bar{\M}_l$
    \EndFor
    \end{algorithmic}
    \end{multicols}
\end{algorithm}

\subsection{Overall FedLPA algorithm and discussions}

In summary, the proposed algorithm FedLPA follows the same paradigm as the standard one-shot federated learning framework. In FedLPA, the clients locally train their models to get $\M_k$ and calculate the local co-variance over its training dataset using the layer-wise Laplace approximation to compute $\A_k, \B_k$. Subsequently, each client transmits their local $\A_k, \B_k, \M_k$ to the server. Following Algorithm~\ref{algorithm:fedlpa}, the server aggregates these contributions to obtain the global expectation, as described in Eq. \ref{EQ:Framework}, then trains the global model parameters, as outlined in Eq. \ref{EQ:MU:obj:solution}.
 Thus, the transmitted data between the clients and the server is solely $\A_k, \B_k, \M_k$ without any extra auxiliary dataset and label information.

 Note that FedLPA can be directly adopted in most common scenarios.  For the special case that the neural model has enormous single-layer weight parameters, how to extend our proposed FedLPA is discussed in Appendix~\ref{appendix:extendlarge}. 
 

 \subsection{t-SNE observation and discussions}
To quickly demonstrate the effectiveness of FedLPA, we show the t-SNE visualization of our FedLPA global model on the MNIST dataset as an example with a biased training data setting among 10 local clients. The experiment details, t-SNE visualizations of the local models and the global models of other algorithms and discussions are in Appendix~\ref{append:tsne}. As shown in Figure~\ref{fig:tsne_our}, FedLPA generates the global model which can clearly distinguish these classes, meanwhile, the classes are separate.

\begin{wrapfigure}{htr}{0.4\textwidth}
  \vspace*{-0.5cm}
    \includegraphics[width=0.4\textwidth]{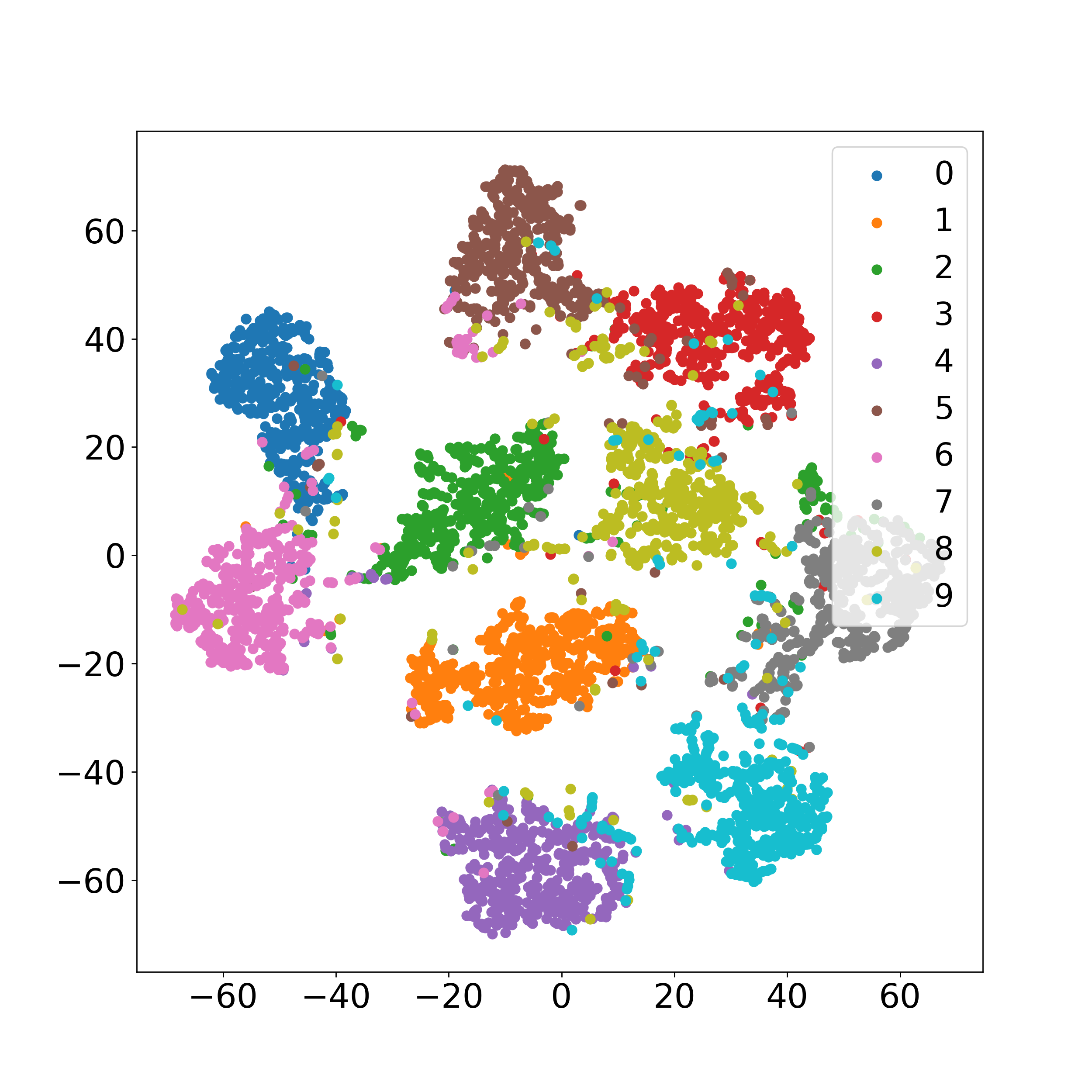}
  \caption{t-SNE visualization for our FedLPA global model.}
  \label{fig:tsne_our}
\end{wrapfigure}

 \subsection{Privacy Discussions}
FedLPA is intuitively compatible with existing privacy-preserving techniques, such as differential privacy (DP) \cite{dwork2006calibrating,lyu2022privacy}, secure multiparty computation (SMC) \cite{yao1982protocols,demmler2015aby}, and homomorphic encryption (HE) \cite{elgamal1985public,paillier1999public,gentry2009fully}. In Appendix~\ref{appendix:dpexp}, we propose a naive DP-FedLPA with two different mechanisms to show the compatibility with differential privacy. Meanwhile, we mention that our proposed FedLPA has the same privacy-preserving level as the conventional federated learning algorithms (i.e, FedAvg, FedProx, FedNova and Dense). Compared with FedAvg, we have conducted a detailed analysis from a privacy attack perspective to show that our proposed FedLAP exhibits a security level consistent with FedAvg against several types of privacy attacks, where the details are shown in Appendix~\ref{appendix:privacyexample}. Note that the main focus of FedLPA is to improve the learning performance on the one-shot FL settings, thus, we leave the integration with other privacy-preserving techniques beyond DP as an open problem.


\section{Experiments}
\begin{table*}[ht!]\tiny
\centering
\caption{ Comparison with various FL algorithms in one round.}
\label{table:main}
\setlength\tabcolsep{3.6mm}
\begin{tabular}{c|c|c|c|c|c|c|c}
\toprule[0.15em]
Dataset&Partition & FedLPA & FedNova & SCAFFOLD & FedAvg & FedProx &DENSE \\
\midrule[0.1em]
\multirow{9}{*}{FMNIST} &$\beta$=0.01 & \cellcolor{green!25} 21.20$\pm$0.67&10.13$\pm$0.00 &15.97$\pm$0.12 &18.17$\pm$0.15 & 13.37$\pm$0.19&15.23$\pm$0.14  \\
&$\beta$=0.05  &\cellcolor{green!25} 54.27$\pm$0.38 & 18.67$\pm$0.41&18.67$\pm$0.41 &18.67$\pm$0.41 &22.03$\pm$0.14 & 47.77$\pm$0.20\\
& $\beta$=0.1 &\cellcolor{green!25} 55.33$\pm$0.06 &30.47$\pm$0.59&31.40$\pm$0.25 &30.93$\pm$0.58 & 31.00$\pm$0.52&52.93$\pm$0.67 \\
&$\beta$=0.3 &\cellcolor{green!25} 68.20$\pm$0.04 & 49.40$\pm$0.26&46.00$\pm$0.02 & 45.17$\pm$0.05&44.30$\pm$0.08 &64.27$\pm$0.08 \\
& $\beta$=0.5&\cellcolor{green!25} 73.33$\pm$0.06 & 57.03$\pm$0.28&56.03$\pm$0.28 &59.10$\pm$0.63 & 58.10$\pm$0.47&72.87$\pm$0.13 \\
&$\beta$=1.0 &\cellcolor{green!25} 76.03$\pm$0.05 & 63.63$\pm$0.33& 66.10$\pm$0.02&62.13$\pm$0.43 &63.10$\pm$0.29 & 72.97$\pm$0.01\\
&\#C=1 &\cellcolor{green!25} 13.20$\pm$0.02 &10.37$\pm$0.00 &10.40$\pm$0.00 & 10.37$\pm$0.00& 13.03$\pm$0.18&10.00$\pm$0.00 \\
&\#C=2 & \cellcolor{green!25} 46.13$\pm$0.15&21.00$\pm$0.10 & 23.53$\pm$0.22&23.20$\pm$0.08 &19.97$\pm$0.10 &38.90$\pm$0.45 \\
&\#C=3  &\cellcolor{green!25}  57.90$\pm$0.06&27.47$\pm$0.02 &27.37$\pm$0.36 &29.20$\pm$0.03 &23.93$\pm$0.33 &53.40$\pm$0.07 \\
\midrule[0.1em]
\multirow{9}{*}{CIFAR-10}  &$\beta$=0.01 &\cellcolor{green!25} 16.17$\pm$0.00 &11.57$\pm$0.02 &11.47$\pm$0.01 &11.53$\pm$0.05 &10.47$\pm$0.00 & 12.30$\pm$0.03\\
&$\beta$=0.05  &\cellcolor{green!25} 18.37$\pm$0.00 &10.30$\pm$0.00 &10.73$\pm$0.01 &10.23$\pm$0.00 &10.97$\pm$0.02 &17.87$\pm$0.31 \\
& $\beta$=0.1 &\cellcolor{green!25} 19.97$\pm$0.02 &12.30$\pm$0.04 & 10.87$\pm$0.01&12.83$\pm$0.06 &11.97$\pm$0.04 &19.93$\pm$0.07 \\
&$\beta$=0.3 &\cellcolor{green!25} 26.60$\pm$0.01 & 11.77$\pm$0.02& 10.93$\pm$0.01&10.53$\pm$0.00 & 10.97$\pm$0.00& 25.57$\pm$0.84\\
& $\beta$=0.5&\cellcolor{green!25} 24.20$\pm$0.02 &11.07$\pm$0.00 & 11.77$\pm$0.02&10.97$\pm$0.00 &11.33$\pm$0.00 &20.17$\pm$0.73 \\
&$\beta$=1.0 &\cellcolor{green!25} 29.33$\pm$0.00 & 12.00$\pm$0.00& 13.00$\pm$0.00&13.23$\pm$0.00 &13.63$\pm$0.01 &28.23$\pm$0.34 \\
&\#C=1 &\cellcolor{green!25} 10.70$\pm$0.01 & 10.50$\pm$0.00& 10.27$\pm$0.00& 10.23$\pm$0.00&10.37$\pm$0.01 &10.00$\pm$0.00 \\
&\#C=2 &\cellcolor{green!25} 16.40$\pm$0.00 &10.07$\pm$0.00 &12.03$\pm$0.08 & 10.07$\pm$0.00&10.03$\pm$0.00 &14.13$\pm$0.22 \\
&\#C=3  &\cellcolor{green!25} 18.97$\pm$0.01 & 11.30$\pm$0.01&11.00$\pm$0.01 &11.53$\pm$0.01 & 10.77$\pm$0.00&14.77$\pm$0.11 \\
\midrule[0.1em]
\multirow{9}{*}{MNIST}  &$\beta$=0.01 &\cellcolor{green!25} 39.17$\pm$1.16 & 13.53$\pm$0.20&8.87$\pm$0.01 & 9.37$\pm$0.00&9.33$\pm$0.00 &15.80$\pm$0.24 \\
&$\beta$=0.05  &\cellcolor{green!25}70.07$\pm$0.05  &31.60$\pm$0.71 &41.07$\pm$0.46 &38.57$\pm$0.28 & 32.23$\pm$0.18& 57.83$\pm$1.55\\
& $\beta$=0.1 &\cellcolor{green!25} 77.43$\pm$0.14 & 48.07$\pm$0.28&47.73$\pm$0.22 & 48.63$\pm$0.15 &47.40$\pm$0.00 & 70.33$\pm$0.02\\
&$\beta$=0.3 &\cellcolor{green!25}85.77$\pm$0.02  & 67.6$\pm$0.40&67.07$\pm$0.15 &66.17$\pm$0.21 &63.40$\pm$0.41 & 84.50$\pm$0.01\\
& $\beta$=0.5&\cellcolor{green!25} 88.73$\pm$0.07 &79.27$\pm$0.08 & 78.57$\pm$0.29&77.37$\pm$0.07 &79.60$\pm$0.24 &86.33$\pm$0.36 \\
&$\beta$=1.0 &\cellcolor{green!25}93.37$\pm$0.08  &84.93$\pm$0.18 & 85.33$\pm$0.15&85.10$\pm$0.13 &86.50$\pm$0.16 &91.43$\pm$0.02 \\
&\#C=1 &\cellcolor{green!25}11.43$\pm$0.01  &10.27$\pm$0.02 &10.10$\pm$0.01 &10.10$\pm$0.01 & 10.13$\pm$0.01& 9.93$\pm$0.00\\
&\#C=2 &\cellcolor{green!25} 69.63$\pm$0.29 &20.90$\pm$0.49 & 25.23$\pm$1.08& 16.47$\pm$0.23&14.30$\pm$0.34 &52.73$\pm$0.46 \\
&\#C=3  &\cellcolor{green!25}77.13$\pm$0.24  &29.53$\pm$1.65 &31.83$\pm$2.45 & 33.13$\pm$2.60&29.00$\pm$2.05 & 58.90$\pm$0.31\\
\midrule[0.1em]
\multirow{9}{*}{SVHN}  &$\beta$=0.01 &\cellcolor{green!25}19.20$\pm$0.00  &13.73$\pm$0.14 &9.83$\pm$0.00 & 12.13$\pm$0.04&11.43$\pm$0.12 &17.33$\pm$0.28 \\
&$\beta$=0.05  &\cellcolor{green!25}22.93$\pm$0.38 & 14.90$\pm$0.43&15.77$\pm$0.14 &16.60$\pm$0.23 &15.90$\pm$0.12 &21.47$\pm$0.20 \\
& $\beta$=0.1 &\cellcolor{green!25} 39.77$\pm$0.69 & 25.97$\pm$0.13 & 25.70$\pm$0.08& 22.17$\pm$0.02&24.50$\pm$0.06 &19.43$\pm$0.45 \\
&$\beta$=0.3 &\cellcolor{green!25}52.23$\pm$0.26  &34.40$\pm$0.28 &34.03$\pm$0.06 &33.93$\pm$0.26 & 34.70$\pm$0.20&47.13+7.14 \\
& $\beta$=0.5&\cellcolor{green!25}54.27$\pm$0.02  & 38.53$\pm$0.07&40.07$\pm$0.13 & 38.53$\pm$0.15&36.93$\pm$0.09 &53.70$\pm$0.07 \\
&$\beta$=1.0 &\cellcolor{green!25} 67.80$\pm$0.01 &55.60$\pm$0.08 &54.03$\pm$0.14 & 55.97$\pm$0.04& 55.23$\pm$0.12&54.40+9.43 \\
&\#C=1 &\cellcolor{green!25} 19.60$\pm$0.00 & 10.43$\pm$0.00&13.73$\pm$0.18 &13.77$\pm$0.17 &18.27$\pm$0.03 &7.70$\pm$0.03 \\
&\#C=2 &\cellcolor{green!25} 47.03$\pm$4.63 &12.90$\pm$0.27 &24.47$\pm$0.08 & 20.17$\pm$0.04& 17.47$\pm$0.13&37.67$\pm$0.76 \\
&\#C=3  &\cellcolor{green!25}48.00$\pm$0.22 &20.87$\pm$0.12 & 28.37$\pm$0.09&27.60$\pm$0.03 &24.93$\pm$0.10 & 47.43$\pm$0.40\\
\bottomrule[0.1em]
\end{tabular}
\end{table*}

\subsection{Experiments settings}
\textbf{Datasets. }We conduct experiments on MNIST~\cite{mnist}, Fashion-MNIST~\cite{fmnist}, CIFAR-10~\cite{cifar10}, and SVHN~\cite{svhn} datasets. In most of the previous works and the most popular benchmark, the majority of their experiments use these datasets and these models. We choose these datasets and models to do the majority of our experiments following these established methods and benchmarks to fairly compare our method with the baselines. We use the data partitioning methods for non-IID settings of the benchmark \footnote{https://github.com/Xtra-Computing/NIID-Bench} to simulate different label skews. Specifically, we try two different kinds of partition: 1) \#C = $k$: each client only has data from $k$ classes. We first assign $k$ random class IDs for each client. Next, we randomly and equally divide samples of each class to their assigned clients; 2) $p_k$ - Dir($\beta$): for each class, we sample from Dirichlet distribution $p_k$ and distribute $p_{k,j}$ portion of class $k$ samples to client $j$. In this case, smaller $\beta$ denotes worse skews.

Here's a brief overview of these datasets. MNIST Dataset: The MNIST dataset comprises binary images of handwritten digits. It consists of 60,000 28x28 training images and 10,000 testing images. FMNIST Dataset: Similar to MNIST, the FMNIST dataset also contains 60,000 28x28 training images and 10,000 testing images. SVHN Dataset: The SVHN dataset includes 73,257 32x32 color training images and 10,000 testing images. CIFAR-10 Dataset: CIFAR-10 consists of 60,000 32x32 color images distributed across ten classes, with each class containing 6,000 images. The input dimensions for MNIST, FMNIST, SVHN, and CIFAR-10 are 784, 784, 3,072, and 3,072, respectively.

\textbf{Training Details. }By default, we follow FedAvg~\cite{fedavg2} and other existing studies~\cite{wang2020principled,li2022federated,diao2023towards} to use a simple CNN with 5 layers in our experiments. The experiments with more complex neural network structures are in Appendix~\ref{appendix:structure}. We set the batch size to 64, the learning rate to 0.001, and the $\lambda=0.001$ for FedLPA. By default, we set 10 clients and run 200 local epochs for each client. For the various settings of the number of clients and local epochs, we refer to Section~\ref{sec:sca} and  Section~\ref{sec:ab}. For results with error bars, we run three experiments with 5 different random seeds. Note that all methods were evaluated under fair comparison settings. Due to the page limit, representative results are represented in the main paper. Refer to Appendix~\ref{appendx:exp} for more experimental details and additional results.

\textbf{Baselines. }To ensure fair comparisons, we neglect the comparison with methods that require to download auxiliary models or datasets, such as FedBE~\cite{chen2020fedbe}, FedKT~\cite{li2020practical} and FedGen~\cite{zhu2021data}, or even pretrained large model, like  FedDISC~\cite{yang2023exploring}. FedOV~\cite{diao2023towards} and FEDCAVE~\cite{heinbaugh2022data} entail sharing more client-side label information or transmitting client label information to the server, which could jeopardize label privacy and are beyond the scope of this study. XorMixFL~\cite{shin2020xor} may not be practical, as we mentioned before. FedFisher~\cite{jhunjhunwala2024fedfisher} is not publicly available. FedDF~\cite{lin2020ensemble}, DAFL~\cite{chen2019data} and ADI~\cite{yin2020dreaming} are compared with the state-of-the-art data-free method DENSE~\cite{zhang2022dense}. Co-Boosting~\cite{dai2024enhancing} requires too many computational resources\footnote{The experiments with more models, FedOV and Co-Boosting, are in Appendix~\ref{append:fedovcoboosting}.}. In conclusion, we include one-shot FL algorithms as baselines including FedAvg~\cite{fedavg2}, FedProx~\cite{fedprox}, FedNova~\cite{fednova}, SCAFFOLD~\cite{scaffold} and DENSE~\cite{zhang2022dense}. All the methods are fairly compared, and our implementation is available and the experiment details can be viewed in Appendix~\ref{appendx:exp:codebase}.

\subsection{An overall comparison}
We compare the accuracy between FedLPA and the other baselines as shown in Table~\ref{table:main}, the data in the green shadow shows the best results. FedLPA can achieve the best performance in all the dataset and partition settings. In extreme cases such as $\beta=\{0.01,0.05\}$, \#C = 1, \#C = 2, FedLPA exhibits a significant performance advantage over the baseline algorithms. This demonstrates our framework's ability to effectively aggregate valuable information from local clients for global weight training. In summary, the state-of-the-art DENSE could be comparable with FedLPA when the skew level is small. However, with the increment of skewness, FedLPA shows significantly superior results.

\begin{table*}[t]\tiny
\centering
\caption{Experimental results of varying number of clients on FMNIST dataset.}
\label{table:clients}
\setlength\tabcolsep{3.4mm}
\begin{tabular}{c|c|c|c|c|c|c|c}
\toprule[0.15em]
\# of Clients &Partition & FedLPA & FedNova & SCAFFOLD & FedAvg & FedProx &DENSE \\
\midrule[0.1em]
\multirow{9}{*}{20 Clients} &$\beta$=0.01 & \cellcolor{green!25}33.57$\pm$0.38 &10.00$\pm$0.00 &13.13$\pm$0.24 & 13.23$\pm$0.21&13.93$\pm$0.08 & 10.30$\pm$0.00\\
&$\beta$=0.05  &\cellcolor{green!25}47.30$\pm$0.74 &21.30$\pm$0.08 &20.53$\pm$0.56 &21.20$\pm$0.64 &19.40$\pm$0.46 &46.13$\pm$0.36 \\
& $\beta$=0.1 &\cellcolor{green!25}57.37$\pm$0.05& 31.50$\pm$0.29&29.23$\pm$0.60 & 32.43$\pm$0.99& 28.80$\pm$1.26&57.20$\pm$0.12 \\
&$\beta$=0.3 &\cellcolor{green!25} 71.30$\pm$0.03&53.87$\pm$0.33 &50.63$\pm$0.10 &52.83$\pm$0.08 & 52.13$\pm$0.40& 71.17$\pm$0.04\\
& $\beta$=0.5&\cellcolor{red!25}74.07$\pm$0.00 & 62.83$\pm$0.03&58.60$\pm$0.08 &60.17$\pm$0.03 & 59.47$\pm$0.06& \cellcolor{green!25}74.10$\pm$0.04\\
&$\beta$=1.0 &\cellcolor{green!25}76.07$\pm$0.01 & 68.63$\pm$0.08&69.13$\pm$0.12 &68.33$\pm$0.08 &69.33$\pm$0.10 & 75.47$\pm$0.04\\
&\#C=1 &\cellcolor{green!25} 21.50$\pm$0.30&10.00$\pm$0.00 & 10.00$\pm$0.00&10.00$\pm$0.00 & 10.33$\pm$0.00&10.00$\pm$0.00 \\
&\#C=2 &\cellcolor{green!25} 59.17$\pm$0.45& 19.23$\pm$0.23&19.47$\pm$0.49 & 18.53$\pm$0.46& 13.53$\pm$0.26&33.07$\pm$0.27 \\
&\#C=3  &\cellcolor{green!25}66.37$\pm$0.01 & 27.30$\pm$0.20& 28.07$\pm$0.35&25.93$\pm$0.27 &24.63$\pm$0.26 &52.23$\pm$0.79 \\
\midrule[0.1em]
\multirow{9}{*}{50 Clients} & $\beta=0.01$ & \cellcolor{green!25}15.91$\pm$0.01 &10.00$\pm$0.00 &10.00$\pm$0.00 &10.00$\pm$0.00 & 10.27$\pm$0.00&10.00$\pm$0.00 \\
&$\beta$=0.05  &\cellcolor{green!25}28.43$\pm$0.80&15.50$\pm$0.43 & 17.77$\pm$0.25&17.37$\pm$0.24 &18.10$\pm$0.01 & 25.03$\pm$0.47\\
& $\beta$=0.1 &\cellcolor{green!25}57.03$\pm$0.00 &34.33$\pm$0.04 &30.17$\pm$0.03 &28.90$\pm$0.05 &31.00$\pm$0.27 & 55.83$\pm$0.49\\
&$\beta$=0.3 &\cellcolor{green!25}66.70$\pm$0.23 &46.70$\pm$0.12 & 43.97$\pm$0.02&45.40$\pm$0.12&45.07$\pm$0.11 &59.23$\pm$1.90 \\
& $\beta$=0.5&\cellcolor{green!25}71.13$\pm$0.00 &57.93$\pm$0.40 &52.93$\pm$0.22 &53.67$\pm$0.26 &53.80$\pm$0.20 &69.57$\pm$0.02 \\
&$\beta$=1.0 &\cellcolor{green!25}71.07$\pm$0.04 &60.00$\pm$0.20 &57.67$\pm$0.22 &56.30$\pm$0.45 &56.90$\pm$0.41 &70.33$\pm$0.03 \\
&\#C=1 &\cellcolor{green!25}15.93$\pm$0.02 &10.00$\pm$0.00 &10.00$\pm$0.00 &10.00$\pm$0.00 & 10.27$\pm$0.00&10.00$\pm$0.00 \\
&\#C=2 &\cellcolor{green!25} 49.60$\pm$0.37&18.03$\pm$0.11 &17.20$\pm$0.00 & 20.50$\pm$0.26&15.70$\pm$0.03 & 44.57$\pm$0.92\\
&\#C=3  &\cellcolor{green!25}65.50$\pm$0.05 &38.03$\pm$0.99 & 40.53$\pm$1.41& 40.97$\pm$1.51&38.93$\pm$1.34 &56.10$\pm$0.38 \\
\bottomrule[0.1em]
\end{tabular}
\end{table*}

\subsection{Scalability}
\label{sec:sca}
We assess the scalability of FedLPA by varying the number of clients. In this section, we show results on FMNIST in Table~\ref{table:clients}. From the table, we can observe that FedLPA still almost always achieves the best accuracy when increasing the number of clients. Notably, there is a slight exception highlighted in red, where DENSE outperforms us when we have $20$ clients and $\beta=0.5$, this may be attributed to the dataset being less biased and the DENSE only getting a marginal 0.03\% higher test accuracy. Our method is generally much more robust in all kinds of settings.

\begin{table}[!h]
\centering
\caption{Experiments with different proportions of data samples.}
\label{table:proportion}
\resizebox{0.7\textwidth}{!}{
\begin{tabular}{c|c|c|c}
\toprule[0.15em]
Data sample proportion	& Accuracy($\beta$=0.1)  & Accuracy($\beta$=0.3) & Accuracy($\beta$=0.5) \\
\midrule[0.1em]
100\%&	55.33$\pm$0.06&	68.20$\pm$0.04&	73.33$\pm$0.06\\
80\%&	53.88$\pm$1.14&	65.47$\pm$0.02&	73.17$\pm$0.05\\
60\%&	53.15$\pm$0.82&	64.80$\pm$0.71&	72.40$\pm$0.29\\
40\%&	53.20$\pm$0.21&	64.10$\pm$0.40&	70.02$\pm$0.17\\
20\%&	45.71$\pm$0.13&	62.15$\pm$0.03&	68.54$\pm$2.02\\
\bottomrule[0.1em]
\end{tabular}
}
\end{table}

\subsection{Experiments with different proportions of data samples}

We have added the experiments with our method on the same experiment setting with 10 clients. We conducted experiments on FMNIST datasets with $\beta$=0.1, 0.3 and 0.5. The performance changes w.r.t the number of data samples are shown in Table~\ref{table:proportion}. We could see that our method FedLPA could yield satisfactory results even with only 20\% data samples under multiple settings.

\subsection{Ablation study}
\label{sec:ab}
The hyper-parameter of our approach is $\lambda$, which controls variances of a priori normal distribution and guarantees $\A_k$ and $\B_k$ are positive semi-definite. In this part, we show results on FMNIST. All other Laplace Approximations are sensitive to the hyper-parameter $\lambda$ based on their experimental results, Table~\ref{table:lambda} shows that our approach is relatively robust. Based on our numerical results, we set $\lambda=0.001$ by default for our method FedLPA.

We also conduct the experiments when the local epochs are 10,20,50,100. More experiments are available in Appendix~\ref{appendx:exp:epochs}, which shows that our methods outperform all the baselines in all kinds of scenarios without requiring extensive tuning.

\begin{table*}
\begin{minipage}[t]{0.45\textwidth}
\centering
\scriptsize
\caption{Experimental results of different hyper-parameter $\lambda$ on FMNIST dataset.}
\label{table:lambda}
\begin{tabular}{c|c|c|c}
\toprule[0.15em]
value of $\lambda$ & 0.01 & 0.001 &0.0001 \\
\midrule[0.1em]
$\beta$=0.01 &18.63$\pm$0.78 &21.20$\pm$0.67&22.50$\pm$1.84\\
$\beta$=0.05  &54.33$\pm$0.54 &54.27$\pm$0.38&53.30$\pm$0.01\\
$\beta$=0.1&56.83$\pm$0.19 &55.33$\pm$0.06&54.60$\pm$0.15\\
$\beta$=0.3 &66.83$\pm$0.02 &68.20$\pm$0.04&67.53$\pm$0.03\\
$\beta$=0.5 &73.20$\pm$0.03 &73.33$\pm$0.06&72.17$\pm$0.04\\
$\beta$=1.0 &76.53$\pm$0.02 &76.03$\pm$0.05&73.47$\pm$0.19\\
\#C=1&12.73$\pm$0.01 &13.20$\pm$0.02&14.17$\pm$0.02\\
\#C=2  &45.20$\pm$0.21 &46.13$\pm$0.15&44.80$\pm$0.03\\
\#C=3&58.97$\pm$0.07 &57.90$\pm$0.06&55.60$\pm$0.06\\
\bottomrule[0.1em]
\end{tabular}   
\end{minipage}
\hspace{7mm}
\begin{minipage}[t]{0.45\textwidth}
\centering
\scriptsize
\caption{Communication and computation overhead evaluation.}
\label{table:runcom}
\begin{tabular}{c|c|c}
\toprule[0.15em]
         & \begin{tabular}[c]{@{}c@{}}Overall \\ Computation (mins)\end{tabular} & \begin{tabular}[c]{@{}c@{}}Overall\\ Communication (MB)\end{tabular} \\
\midrule[0.1em]
FedLPA   & 65                         & 4.98                       \\
FedNova  & 50                         & 2.47                       \\
SCAFFOLD & 50                         & 4.94                      \\
FedAvg   & 50                         & 2.47                        \\
FedProx  & 75                         & 2.47                     \\
DENSE    & 400                        & 2.47                        \\
\bottomrule[0.1em]
\end{tabular}
\end{minipage}
\end{table*}

\subsection{Communication and computation overhead}
We conduct experiments on CIFAR-10 on a single 2080Ti GPU to estimate the overall communication and computation overhead. We set the number of clients is 10. Table~\ref{table:runcom} shows the numerical results on FedLPA and baselines. Details of the overhead evaluation are referred to Appendix~\ref{append:com} and~\ref{append:run}. Our observations reveal that FedLPA is slightly slower than FedNova, SCAFFOLD, FedAvg, and FedProx, while much faster than DENSE. FedLPA also has significantly improved the one-shot learning performance of the above four approaches. Similarly, FedLPA performs moderately incremental communication overhead while outperforming other baseline approaches on learning performance, as one-shot FL introduces heavy computation overhead while communication overhead is usually small. It is noteworthy that FedLPA strikes a favorable balance between computation and communication overhead, making it the most promising approach for one-shot FL.

\begin{wrapfigure}{hr}{0.45\textwidth}
    \includegraphics[width=0.45\textwidth]{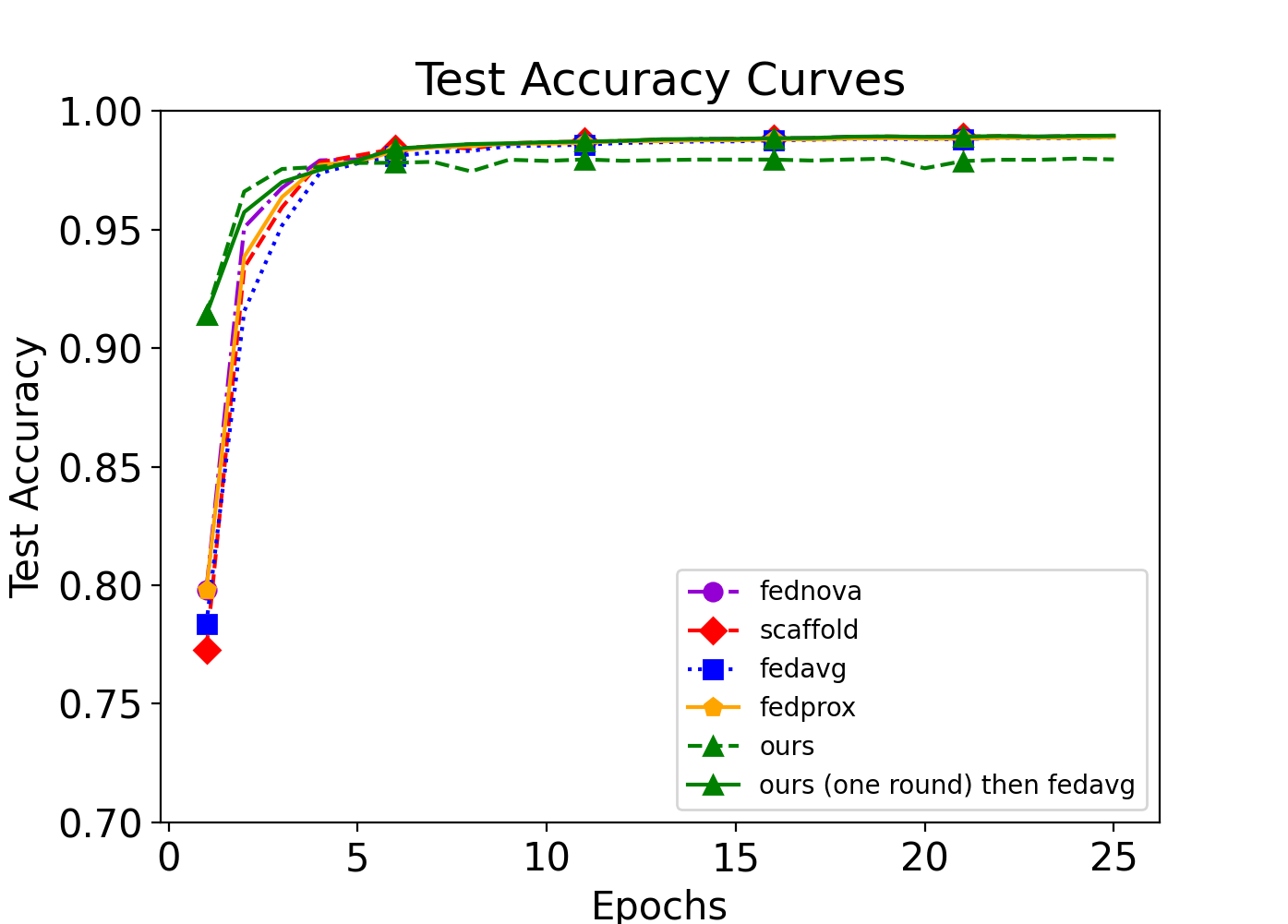}
  \caption{Extension to multiple rounds on MNIST dataset.}
  \label{fig:multiple_round}
\end{wrapfigure}

\subsection{Extension to multiple rounds}
\label{sec:extend}
We conduct experiments on MNIST with 10 clients and data partitioning $p_k$ - Dir($\beta=0.5$). The results are shown in Figure~\ref{fig:multiple_round}. As DENSE could not support multiple rounds, we compare our methods with FedAvg, FedNova, SCAFFOLD, and FedProx. FedLPA achieves the highest accuracy in the first round, denoting the strongest learning capabilities in a one-shot setting. With the increment in the number of rounds, the performances of FedLPA increase slower than the other baseline approaches. This figure shows that the joint approach (ours (one round) then FedAvg) that utilizes FedLPA in the first round and then adopts other baseline methods may be most promising to save communication and computation resources in the multiple-round federated learning scenario. 

\subsection{Supplementary experiments}

Experiments for privacy concerns, experiments on different local epoch numbers, experiments in extreme settings (the number of clients=5, $\beta=0.001$),  experiments with more methods, experiments with more complex network structures, experiments with more complex datasets, ablation experiments analyzing the number of approximation iterations of FedLPA can be found in Appendix. 


\section{Conclusions}

In this work, we design a novel one-shot FL algorithm FedLPA to better model the global parameters in effective one-shot federated learning. We propose a method that could aggregate the local clients in a layer-wise manner with their posterior approximation via block-diagonal empirical Fisher information matrices, which could effectively capture the accurate statistics of a locally biased dataset. Overall, FedLPA stands out as the most practical and efficient framework that conducts data-free one-shot FL, particularly well-suited for high data heterogeneity in various settings, considering it significantly outperforms other baselines with extensive experiments. Our FedLPA is available in https://github.com/lebronlambert/FedLPA\_NeurIPS2024.

\begin{ack}
We thank the anonymous reviewers. The authors would like to thank Yiqun Diao from National Univeristy of Singapore for his valuable comments to improve this work. Dr. Jialin Li is supported by the Singapore Ministry of Education Academic Research Fund Tier 1 (T1 251RES2104) and Tier 2 (MOE-T2EP20222-0016).
\end{ack}






\clearpage
\appendix

\section{The FedLPA algorithm}
\label{appex:fedlpaal}

The proposed algorithm follows the same paradigm as the standard one-shot federated learning framework. Each client follows the local training procedure as shown in the paper. The global aggregation is illustrated in Algorithm~\ref{algorithm:fedlpa}.

With the Algorithms, let us assume the dimensionality list of each layer in a fully connected neural network is ([$s_0$, $s_1$, $s_2$, ..., $s_l$,..., $s_L]$), which means the size of the weight $\W_{k_l}$ of layer $l$ is $s_{l-1}$x$s_l$. Consequently, the size of $\A_{k_l}$ for this layer would be $s_{l-1}$x$s_{l-1}$, and the size of $\B_{k_l}$ would be $s_l$x$s_l$. The size of $\F_{k_l}$ is ($s_{l-1}$x$s_l$)x($s_{l-1}$x$s_l$).

Then, we give a concrete example to show the dimensions of different matrices using a fully-connected neural network model with architecture 784-256-64-10 as in Appendix~\ref{appendx:exp}. Then, the $\M_{k_1}$ is 784x256+256, $\M_{k_2}$ is 256x64+64, $\M_{k_3}$ is 64x10+10. The $\A_{k_1}$ is 785x785, $\A_{k_2}$ is 257x257, $\A_{k_3}$ is 65x65. The $\B_{k_1}$ is 256x256, $\B_{k_2}$ is 64x64, $\B_{k_3}$ is 10x10. Then the  $\F_{k_1}$ is (785x785)x(256x256), $\F_{k_2}$ is (257x257)x(64x64), $\F_{k_3}$ is (65x65)x(10x10). The $\F_k$ is (785x785+257x257+65x65)x(256x256+64x64+10x10). 

However, in fact, we do not need to combine the $\A_{k_l}$, $\B_{k_l}$, $\F_{k_l}$ into $\A_k$, $\B_k$, $\F_k$. In this paper, we utilize the diagonal block property to compute each block in our method.

\section{Comparison with the previous methods}
\label{appendix:related}

 To the best of our knowledge, we are the first to consider the posterior inference problem in the one-shot scenario. Note that the approach~\cite{al2020federated} requires a lengthy burn-in period before conducting posterior inference, for instance, 400 rounds, and it updates global model parameters by modifying the covariance-aggregated local models. It means that the algorithm~\cite{al2020federated} necessarily requires multiple iterations and cannot be used in a one-shot scenario. In contrast, our method FedLPA only requires immediate variational inference after training the local model, ensuring higher flexibility and efficiency in the one-shot scenario.

Besides, in the algorithm~\cite{al2020federated}, obtaining statistical information to compute local covariances is of low rank. In reality, it fails to acquire the posterior of the aggregated model and cannot perform variational inference on the aggregated model. However, our method yields full-rank covariances, and after employing an expectation approximation method for variational inference on the aggregated model, we can achieve a usable global posterior.

In both the domain of natural gradient optimization~\cite{martens2015optimizing,grosse2016kronecker,al2020federated} and modeling output uncertainty in variational inference~\cite{botev2017practical}, using the Fisher approximation of the Hessian does not involve the issue of inverting covariance. However, in the context of federated learning, when performing variational inference on the aggregated model, the necessity of inverting covariance becomes unavoidable. To address this problem, we propose a novel algorithm that constructs a quadratic objective function. During aggregation, this algorithm directly trains the aggregated model using local covariances and expectations, thereby circumventing the need for inversion operations. 


Note that the previous methods~\cite{matena2022merging,liu2023bayesian} adopt the same core approach that utilizes the online Laplace approximation to obtain diagonal Fisher for model aggregation, in which they conduct experiments on different datasets and published on different venues. We mainly analyze our approach with the comparison of DiagonalFisher~\cite{liu2023bayesian}. DiagonalFisher assumes independence among parameters, neglecting inter-parameter correlations, resulting in inaccurate posterior approximations. However, strong correlations exist among parameters within each layer, such as matching patterns in convolutional kernels within convolutional networks. This is a crucial factor that cannot be overlooked; otherwise, aggregation of the posterior would result in lower posterior regions, as compared to our method. In complex environments, employing diagonal Fisher for aggregation would prove to be entirely ineffective, whereas our method effectively leverages inter-parameter correlations at each layer, rendering it more robust. To demonstrate, we present results comparing aggregation using diagonal Fisher and our method. We have added experiments using the settings of our paper and an MLP model (784-256-64-10) on the FMNIST dataset with five random seeds for one-shot FL, the client number is 10, and the $\beta$=0.01. The results are in Table~\ref{table:compare1}.

\begin{table}
\centering
\caption{Experiments with DiagonalFisher using MLP.}
\label{table:compare1}
\setlength\tabcolsep{3.5mm}
\resizebox{\linewidth}{!}{
\begin{tabular}{c|c|c|c|c|c}
\toprule[0.15em]
Initial &FedAvg	&FedProx&SCAFFOLD&DiagonalFisher&FedLPA \\
\midrule[0.1em]
Same	&42.35$\pm$0.16	&24.80$\pm$0.10	&42.10$\pm$0.15	&56.34$\pm$0.34	&76.63$\pm$0.04\\
Different & 10.00$\pm$0.00	&24.12$\pm$0.02&	10.16$\pm$0.70	&10.51$\pm$0.11	&73.73$\pm$0.07\\
\bottomrule[0.1em]
\end{tabular}
}
\end{table}

In the table, ``Initial" denotes whether the client models were initialized using the same parameter values or independently.

When ``Initial" is set to ``Same", all client models are trained on their respective datasets using identical parameter values for initialization. Consequently, there exists a strong correlation among the local models. Additionally, in this scenario, model aggregation is equivalent to aggregating updates of local models. Although DiagonalFisher performs reasonably well under this condition, our method demonstrates superior performance, exhibiting a 20.29\% increase in global test accuracy.

When ``Initial" is set to "Different", the models on different clients start training with distinct parameter values. Due to the high heterogeneity of local datasets, there is minimal correlation among local models. In this extreme scenario, DiagonalFisher completely fails, while our method maintains an accuracy of 73.73\%, showcasing remarkable robustness.

It is essential to consider the indispensability of parameter correlations, which is why we compute correlations among parameters within layers to ensure the robustness and accuracy of model aggregation.

Now, we discuss some related works which directly utilize K-FAC to approximate the Fisher matrix and make a comparison with our proposed approach FedLPA. The works~\cite{martens2015optimizing,grosse2016kronecker,botev2017practical} have provided us with significant inspiration. However, methods like K-FAC do not require computing the inverse of covariance. Nevertheless, in the context of federated learning, the necessity of inverting covariance becomes unavoidable during variational inference on the aggregated model.

Methods like K-FAC assume direct independence among data samples to utilize expectation approximation. They obtain the inverse of Fisher from individual samples and then directly compute the expectation, thereby avoiding inverse operations. However, the expectation approximation inevitably leads to biased results during model aggregation. Detailed analysis can be found in Appendix~\ref{append:EA}.

To address this issue, we propose a novel algorithm that constructs a quadratic objective function. During aggregation, this algorithm directly trains the aggregated model using local covariances and expectations, eliminating the need for inversion operations. This aims to minimize aggregation biases as much as possible. 


Here, we provide a comparative analysis of different methods.

FedAvg and FedProx minimize the Kullback-Leibler (KL) divergence between the local and global posteriors: $ \bar{\bm{\mu}}, \bar{\bm{\Sigma}}^{-1} = \min_{ \bar{\bm{\mu}}, \bar{\bm{\Sigma}}^{-1}} KL\left((\sum_{k}^{K} p(\bm{\theta} | \D_k)) | p(\bm{\theta} | \D)\right)$. SCAFFOLD computes the bias term, and FedNova computes the correction term. None of these four methods consider the correlations between parameters. DENSE and FedOV, on the other hand, employ distillation methods, attempting to extract the distribution of non-iid data among clients through distillation. However, this itself leads to information loss due to dimensionality reduction and introduces additional variance of data.

Although the work~\cite{angelino2016patterns} also uses the distributed Bayesian inference, however, it focuses on the dataset feature and could not be applied to train the global model parameters.

In conclusion, the reason our approach performs better in this scenario stems from our improved approximation of the global posterior. This approach signifies our novelty in addressing these challenges.

\subsection{The efficiency of FedLPA}
Although the number of uploaded bits increased per round of FedLPA, it resulted in a significant improvement in the final outcome. Additionally, the increase in transmitted bits enhanced the robustness of the aggregation method. Moreover, as indicated in Table~\ref{table:runcom} of the paper, we observe only a marginal increase in the amount of communication required.

A fully-connected neural network model with architecture 784-256-64-10, has $784 \cdot 256+256+256 \cdot64+64+64 \cdot10+10 = 217930$ floating point numbers, which is 6973760 bits or around 0.831 MB. For one communication from a client to the server, our approach needs to upload additional $\A_k$ and $\B_k$, which have $785 \cdot785+256 \cdot256+257 \cdot257+64 \cdot64+65 \cdot65+10 \cdot10 = 756231$ floating point numbers. Note, $\A_k$ and $\B_k$ are symmetric matrices, so we only need to upload the upper triangular part of $\A_k$ and $\B_k$, which is around 756231/2 = 378115.5 floating point numbers and 1.442 MB. Therefore, our approach costs 2.272 MB for the directed communication, which is only 1.367 times than DiagonalFisher while DiagonalFisher costs 0.831*2 = 1.662 MB. We show the following Table~\ref{table:compare} based on the previous experiment results. When ``Initial" is set to ``Same", the efficiency of every bit is almost the same. When ``Initial" is set to ``Different", the efficiency of every bit for our method is much higher than the DiagonalFisher.

\begin{table}
\centering
\caption{Experiments with DiagonalFisher considering efficiency.}
\label{table:compare}
\setlength\tabcolsep{3.4mm}
\resizebox{\linewidth}{!}{
\begin{tabular}{c|c|c}
\toprule[0.15em]
``Initial" Method & FedLPA (Global Test Acc / MB)  & DiagonalFisher (Global Test Acc / MB) \\
\midrule[0.1em]
``Same" &  76.63/(2.272*10) = 3.37 &56.34/(1.662*10) = 3.39\\
\midrule[0.1em]
``Different" & 73.73/(2.272*10) = 3.25 &10.51/(1.662*10) = 0.63 \\
\bottomrule[0.1em]
\end{tabular}
}
\end{table}

\section{Detailed discussion for the time complexity of Eq. \ref{eq:gsc}}
\label{append:unacceptable}

A fully-connected neural network model with architecture 784-256-64-10 as an example is shown in Appendix K. We use this example to further explain this question. The size of  $\A_{k_1}$ is 785x785 and the size of $\B_{k_1}$ are both 256x256. Then, we need to compute the inverse of the matrix (785x785)x(256x256), which is huge. The time complexity of calculating the inverse of a matrix is $O(n^3)$ (n is the dimension of the matrix), which is very slow. The accuracy of calculating it is decided by the condition number of the huge matrix~\cite{cheney1998numerical,belsley2005regression,pesaran2015time,trefethen2022numerical}. That’s why calculating Eq. \ref{eq:gsc} is unacceptable, considering the time complexity, the size of the huge matrix and the accuracy. 

Further, for example,  in the machine learning field, to accelerate the training of the neural network, they use the Newton method. However, using this method, they need to compute the inverse of the Hessian matrix, which is also huge and unacceptable. That is why they introduce the KFAC~\cite{KFAC1,KFAC2}, KFRA~\cite{KFRA} and KFLR~\cite{KFRA} methods to avoid computing the inverse of the huge Hessian matrix.

In this paper, we avoid computing the inverse of the huge matrix via our method, and the time complexity is linear.

\section{Expectation approximation (EA)}
\label{append:EA}
Previous works~\cite{KFAC1, KFAC2} approximate the expectation of Kronecker products by a Kronecker product of expectations $\mathbb{E} \left[ \A \otimes \B \right] \approx \mathbb{E} \left[ \A \right]\otimes \mathbb{E} \left[ \B \right]$ with an assumption of $\A_{k_l}$ and $\B_{k_l}$ are independent, which is called Expectation Approximation (EA). 

It is a simple and effective method to approximate the expectation of Kronecker products. As a result, the global co-variance $\bar{\mathbf{\Sigma}}$ is approximated by:
\begin{equation}
    \begin{aligned}
        \bar{\Si}_l 
        &\approx (\sum_k^{K} \A_{k_l})^{-1} \otimes (\sum_k^{K}  \B_{k_l})^{-1} = \bar{\A}_{l}^{-1} \otimes \bar{\B}_{l}^{-1}
    \end{aligned}
    \label{EQ:Global:Sigma:KP}
\end{equation}
where $\bar{\A}_{l} = \sum_k^{K} \A_{k_l}$ and $\bar{\B}_{l} = \sum_k^{K}  \B_{k_l}$.
Denoting $\bar{\Z}_l$ as matrix formula of $\bar{\mathbf{z}}_l = \vec(\bar{\Z}_l)$, then $\bar{\bm{\mu}}_l$ can be computed efficiently as below:

\begin{equation}
    \begin{aligned}
        \bar{\bm{\mu}}_l &= \bar{\Si_l} \cdot \bar{\mathbf{z}}_l 
        \approx (\bar{\A}_{l}^{-1} \otimes \bar{\B}_{l}^{-1}) \bar{\mathbf{z}}_l 
        = \vec(\bar{\B}_{l}^{-1}\bar{\Z}_l\bar{\A}_{l}^{-1})
    \end{aligned}
    \label{eq:ea}
\end{equation}


However, Eq. \ref{eq:ea} leads to a biased global expectation. The EA needs the independence assumption, but $\A_{k_l}$ and $\B_{k_l}$ are weakly related in back-propagation.  Besides, even if they are independent, Eq. \ref{eq:ea} still suffers from approximation error because the clients' number $K$ is finite and always a small number but statistical independence can only be demonstrated when the sampling number is large enough.
Eq. \ref{eq:eaerror} shows the approximation error directly:
\begin{equation}
    \begin{aligned}
         (\A_1 + \A_2) \otimes (\B_1 + \B_2) &= \A_1 \otimes \B_1 + \A_2 \otimes \B_2 \\
         &\qquad \quad + \A_1 \otimes \B_2 + \A_2 \otimes \B_1 \\
         &\neq \A_1 \otimes \B_1 + \A_2 \otimes \B_2
    \end{aligned}
    \label{eq:eaerror}
\end{equation}


\section{Extend FedLPA to the models with enormous single layer weight parameters}
\label{appendix:extendlarge}
This implies a Fisher matrix with a large dimension and it significantly increases communication costs. In such cases, the most intuitive approach is to explore the possibility of dimensionality reduction for its Fisher matrix. A promising approach to enhance the efficiency of our method may employ some low-rank factorization techniques~\cite{lee2020estimating}. As described~\cite{daxberger2021laplace}, the main idea involves performing an eigendecomposition on the Kronecker factors~\cite{george2018fast}, while preserving only the eigenvectors corresponding to the top k largest eigenvalues. As a result, this approach drastically reduces space complexity, enabling communication costs to be compared favorably with diagonal Fisher matrices.

\section{Privacay discussion of FedLPA}
\label{appendx:privacy}
In the FedLPA, $\A_k$ is computed via the activations while $\B_k$ is computed via the linear pre-activations of the layer. We note that $\A_k$, $\B_k$, and $\M_k$  do not carry any label information, thus the transmission of $\A_k$, $\B_k$, and $\M_k$ will not leak any label privacy.  As a comparison, FedCAVE, which transmits client label information to the server, requires training in label distribution to do the distillation. Several papers~\cite{sun2022label,wainakh2022user} have notified that label privacy, e.g., the concern of label distribution leakage and raw label leakage, is sensitive in federated learning. We believe that it has also been a concern in the one-shot FL scenario.

Besides, our t-SNE illustration in Fig~\ref{fig:tsne_our} shows the classification capability on the global model, which can separate the classes. However, our figures of the t-SNE illustrations on local models in Appendix~\ref{append:tsne} show that for the data belonging to the same class, their t-SNE illustrations are erratically distributed on different local nodes. For instance, for node 2, its training data only has 3 classes while most of the training data locates in class 5. However, it is hard for the server to infer that label distribution since the t-SNE illustration both on node 2 and other nodes also seems irregular.

$\A_k$, $\B_k$, and $\M_k$ are a function of data that may contain privacy-sensitive information of the local training data. However, in this case, our privacy-preserving level is similar to FedAvg, which means that FedLPA has the same privacy-preserving level as the conventional federated learning algorithms (i.e, FedAvg, FedProx, FedNova, and Dense), which are all vulnerable to some privacy attacks (e.g, membership inference~\cite{melis2019exploiting} or reconstruction attacks~\cite{geiping2020inverting}). Our approach FedLPA provides more information than FedAvg, However, the additional information we provide is the mean of each sample in each dimension, the mean of squares of each sample in each dimension, and the mean of square gradients. These solely marginally enrich the attack capability of several reconstruction attacks.

FedLPA is intuitively compatible with existing privacy-preserving techniques, such as differential privacy (DP)~\cite{dwork2006calibrating,lyu2022privacy}, secure multiparty computation (SMC)~\cite{yao1982protocols,demmler2015aby}, and homomorphic encryption (HE)~\cite{elgamal1985public,paillier1999public,gentry2009fully}. In Appendix~\ref{appendix:dpexp}, we propose a naive DP-FedLPA with two different mechanisms to show the compatibility with differential privacy. In Appendix~\ref{appendix:idlgexp}, using iDLG attack~\cite{geiping2020inverting}, we show that our proposed FedLPA has the same privacy-preserving level as the conventional federated learning algorithms (i.e, FedAvg, FedProx, FedNova and Dense). Compared with FedAvg, we have conducted a detailed analysis from a privacy attack perspective to show that our proposed FedLAP exhibits a security level consistent with FedAvg against several types of privacy attacks, where the details are shown in Appendix~\ref{appendix:privacyexample}. 
 Note that the main focus of FedLPA is to improve the learning performance on the one-shot FL settings, thus, we leave the integration with other privacy-preserving techniques beyond DP as an open problem.

\subsection{Experiments with differential privacy}
\label{appendix:dpexp}
We first list the definitions and techniques for differential privacy~\cite{dwork2011differential}. ($\epsilon$-DP) For $\epsilon>0$, a randomized function $f$ provides $\epsilon$-differential privacy if, for any datasets $D,D'$ that have only one single record different, for any possible output $O$,
\begin{equation}
    Pr[f(D)\in O] \leq e^{\epsilon} \cdot Pr[f(D')\in O]
\end{equation}

Suppose $f$ is a function and $D,D'$ have only one record different. The sensitivity of $f$ is defined as 
\begin{equation}
    \Delta f= \max_{D,D'} \|f(D)-f(D')\|_1
\end{equation}

Here one record different means a database has one more record than another.  We utilize the Laplace mechanism~\cite{dwork2014algorithmic} to achieve the $\epsilon-$DP.

Laplace Mechanism: For function $f:\mathcal{D}\rightarrow R^d$, function:

\begin{equation}
    F(D)=f(D)+Lap(0,\Delta f/\epsilon)
\end{equation}

provides $\epsilon$-DP, where $Lap(0,\Delta f/\epsilon)$ is sampled from Laplace distribution. 

Following the differential privacy (DP) mechanisms~\cite{dwork2006calibrating,geyer2017differentially,lyu2022privacy,diao2023exploiting} to protect privacy, we conduct the two mechanisms of DP-FedLPA: (1) adding Laplace random noise to the training data samples, (2) adding Laplace random noise to the parameters to be transmitted. DP is a rigorous and popular privacy metric, which guarantees that the output does not change with a high probability even though an input data record changes. Specifically, since the sensitivity of the data sample distribution after the normalization is 1, we add Laplacian noises with $\lambda=\frac{1}{\epsilon}$. We set $\epsilon=\{3,5,8\}$ that provides modest privacy guarantees since normally  $\epsilon \in (1,10)$ is viewed as a suitable choice. We have added the experiments using the same experiment setting in the paper with five random seeds and 10 clients on the FMNIST dataset. Results are shown in Table \ref{table:dp_all}. DP-FedLPA under both mechanisms outperforms FedAvg, which shows that it is compatible with combining our proposed FedLPA with DP to enhance privacy protection levels.   Note that the smaller $\epsilon$ is, the larger noises we add. We find that when the $\epsilon$ gets smaller, the performance drops simultaneously, while the privacy protection level is increased.

Besides, we have added the experiments using the same experiment setting to show the round results of how many rounds DP-FedAvg needs to achieve the same test performance with the first mechanism. The results in Table~\ref{table:dp_round} show that DP-FedAvg needs about 10 rounds of communication to achieve the same test performance, compared to our one-round FedLPA. Combined with our previous results in Table~\ref{table:runcom} and Table~\ref{table:dp_all}, our FedLPA could save the communication and computation overhead and combine with the DP method to mitigate the potential privacy leakage. Based on the above settings, DP-FedAvg needs at least 3x communication overhead and 5x computation overhead. While DP-FedAvg needs multiple rounds to get similar accuracy, DP-FedAvg maybe vulnerable to more privacy attack methods due to the multiple queries, such as curvature-based privacy attacks.


\begin{table}
\centering
\caption{Experiments with Differential Privacy using two mechanisms.}
\label{table:dp_all}
\setlength\tabcolsep{3.4mm}
\resizebox{\linewidth}{!}{
\begin{tabular}{c|c|c|c|c}
\toprule[0.15em]
$\epsilon$ & Partitions & FedAvg& DP-FedLPA (mechanism 1) & DP-FedLPA (mechanism 2)   \\
\midrule[0.1em]
\multirow{3}{*}{8} &$\beta$=0.1& 31.90$\pm$0.58& 50.01$\pm$0.07 & 57.15$\pm$1.23\\
&$\beta$=0.3 & 44.37$\pm$0.05& 68.30$\pm$0.41 & 66.21$\pm$0.14\\
& $\beta$=0.5& 57.92$\pm$0.63& 71.17$\pm$0.27 & 73.50$\pm$0.06\\
\midrule[0.1em]
\multirow{3}{*}{5} &$\beta$=0.1& 28.17$\pm$0.16& 48.51$\pm$0.07 & 55.87$\pm$ 0.88\\
&$\beta$=0.3 & 43.91$\pm$0.05& 67.34$\pm$0.92 & 66.02$\pm$0.71\\
& $\beta$=0.5& 57.14$\pm$0.63& 70.89$\pm$0.80 & 73.44$\pm$0.20\\
\bottomrule[0.1em]
\multirow{3}{*}{3} &$\beta$=0.1& 27.85$\pm$0.79& 48.39$\pm$0.07 & 54.31$\pm$0.44\\
&$\beta$=0.3 & 42.80$\pm$0.05& 65.08$\pm$0.45 & 65.22$\pm$0.46\\
& $\beta$=0.5& 54.80$\pm$0.63& 70.28$\pm$1.30 & 72.19$\pm$0.62\\
\bottomrule[0.1em]
\end{tabular}
}
\end{table}

\begin{table}
\centering
\caption{Experiments with Differential Privacy for Round Numbers.}
\label{table:dp_round}
\setlength\tabcolsep{3.5mm}
\begin{tabular}{c|c|c|c}
\toprule[0.15em]
$\beta$\ $\epsilon$   & 8&  5&3 \\
\midrule[0.1em]
0.1	& 	11 &11 &12 \\
0.3 & 11&9&8\\
0.5& 8&8&7\\
\bottomrule[0.1em]
\end{tabular}
\end{table}



\subsection{Experiments with iDLG attack}
\label{appendix:idlgexp}
We also add experiments with iDLG attack~\cite{geiping2020inverting} following the link (https://github.com/PatrickZH/Improved-Deep-Leakage-from-Gradients/blob/master/iDLG.py). We did the experiments with the setting of the paper~\cite{geiping2020inverting}: in each single experiment, the client is trained with one random picked image in FMNIST, then we use the iDLG attack to recover the image based on the model from FedAvg and FedLPA. We randomly selected 500 training examples to collect 500 MSEs between the recovered and the original image. The larger the MSE is, the better the privacy-preserving level for the method. Due to the rebuttal limitation, we cannot show the figure for the cumulative distribution function considering the MSE of the iDLG attack. We provide the results in Table~\ref{table:idlg} to show MSE considering the percentile for these 500 experiments.


\begin{table}
\centering
\caption{iDLG attack results of FedLPA and FedAvg.}
\label{table:idlg}
\setlength\tabcolsep{3.5mm}
\resizebox{\linewidth}{!}{
\begin{tabular}{c|c|c|c|c|c|c|c|c}
\toprule[0.15em]
Percentile &12.5 &25.0&37.5&50.0 &62.5& 75.0& 87.5& 100.0\\
\midrule[0.1em]
FedLPA & 0.60&1.00&1.10&1.39&2.75&50.71&40736.88&>=1e9 \\
\midrule[0.1em]
FedAvg & 0.09&0.16&0.71&1.56&26.24&950.89&54307.91&>=1e9 \\
\bottomrule[0.1em]
\end{tabular}
}
\end{table}

Based on the Table, we could see that from 12.5 to 50.0 percentile, regarding the privacy-preserving aspect, FedLPA behaves better than FedAvg on these samples. However, from 50.0 to 87.5 percentile, FedAvg behaves better than FedLPA on such samples. Thus, no clear evidence exists of which one performs better when referring to the privacy level.  Considering the overall 500 data samples, we roughly concluded that FedLPA and FedAvg share a similar privacy level.

\subsection{Concrete examples of privacy attack}
\label{appendix:privacyexample}
For privacy attacks, we start by assuming the simplest scenario where each client has only one sample, and the model comprises a single layer, such as a multi-layer perceptron.

Let $\mathbf{y}=\W*\mathbf{x}$, ($\mathbf{x}$ is $n+1$ dimensional, with the last dimension being a unit value 1), $\mathbf{g}=Df(\mathbf{y})/D\mathbf{x}$ (where $f$ is the loss function). In this case, $\A=\mathbf{x}\mathbf{x}^T, \B=\mathbf{g}\mathbf{g}^T$.

In this single-sample scenario, an attacker can directly obtain $\mathbf{x}$ from the last column of $\A$. With $\mathbf{x}$ and $\W$, the attacker can acquire the model's output. Furthermore, utilizing the Loss and $\mathbf{g}$, it's possible to get the label information.

FedAvg would also be vulnerable to a reconstruction attack in this scenario, allowing the attacker to obtain sample and label information.

When each client has two samples $(\mathbf{x}_1 \mathbf{x}_2 \in Dataset)$, then: $\A=1/2*\mathbf{x}_1*\mathbf{x}_1^T+1/2*\mathbf{x}_2*\mathbf{x}_2^T, \B=1/2*\mathbf{g}_1*\mathbf{g}_1^T+1/2*\mathbf{g}_2*\mathbf{g}_2^T$.  The last column $\mathbf{c}$ of $\A$ equals $1/2\mathbf{x}_1+1/2\mathbf{x}_2$. The diagonal elements $\mathbf{d}$ of $\A$ equal $1/2\mathbf{x}_1^2+1/2\mathbf{x}_2^2$. In the case of these two samples, an attacker can utilize the information from $\A$ and $\B$ to get the two samples $\mathbf{x}_1$ and $\mathbf{x}_2$. Using the same methodology, they can also obtain $\mathbf{g}_1$ and $\mathbf{g}_2$ . Consequently, the attacker can reverse-engineer the labels as well.

FedAvg could also potentially succumb to a reconstruction attack in this scenario, providing the attacker with sample and label information, although the obtained information might be more ambiguous.

When each client has three or more samples ($\mathbf{x} \in Dataset$), $\A=\mathbb{E}_{\mathbf{x} \in Dataset}(\mathbf{x}*\mathbf{x}^T), \B=\mathbb{E}_{\mathbf{x} \in Dataset}(\mathbf{g}*\mathbf{g}^T)$. In this situation, the last column $\mathbf{c}$ of $\A$, $\mathbf{c}=\mathbb{E}_{\mathbf{x} \in Dataset}(\mathbf{x})$ represents the average of the sample dataset, depicting the projection of the data distribution in the sample space on various coordinate axes. Furthermore, the diagonal elements of $\A (\mathbb{E}_{\mathbf{x} \in Dataset}(\mathbf{x}*\mathbf{x}^T))$ offer the attacker statistical information about this local dataset.

Generally, solely using the statistical information of these datasets cannot reconstruct the entire dataset. Similarly, it's not possible to obtain gradients for the output of each sample, thereby preventing the reconstruction of individual sample labels. The results obtained by using $\mathbf{c}$ and $\W$ to gather statistical label information are unreliable.

Additionally, for structures such as CNNs and RNNs/LSTMs, the difficulty of attacks increases due to weight sharing. For CNNs, since convolutional kernels only accept local samples as input, information in $\A$ encompasses statistical information from all localities of the samples. For RNNs/LSTMs, information in $\A$ includes statistics of each word vector in a sentence. These network structures make it possible for attackers to fail even in single-sample scenarios. For MLPs, the information contained in the intermediate layer $\A$ is almost equivalent to the information encoded in the parameters of the BN (Batch Normalization) layer. The mean output of the Batch Normalization (BN) layer is equivalent to the last column of $\A$, whereas the variances differ between the BN layer and $\A$'s diagonal but both contain statistical information related to squared values.

It's worth noting that the parameters acquired by the BN layer using the sliding-window average method are also frequently used during the computation of $\A$ and $\B$, as mentioned in the paper~\cite{martens2016second}.

FedAvg provides model parameter values, the average of gradients, and BN layer parameters. Compared to FedAvg, the additional information we offer is actually limited to: the mean of each sample in each dimension, the mean of squares of each sample in each dimension, and the mean of square gradients. Utilizing this information, attacking becomes highly challenging when the number of samples exceeds three. Although we don't rule out the possibility of successful methods in practice due to the data's own correlations, the limitations are significant based on our analysis, and our security level is quite close to that of FedAvg.

We discuss two common attacks here. Inferring class representatives: 

i) Model inversion attacks~\cite{fredrikson2015model} exploit the confidence information provided by machine learning applications or services. Our method does not provide confidence information, nor does it compute the information required for it. Therefore, our method's defense level against these attacks aligns with FedAvg's defense level.

ii) Attacks using GANs to construct class representatives~\cite{hitaj2017deep} utilize the client-uploaded model as a discriminator and its output as labels to train a generator to generate similar data. The additional statistical information we provide might be used to constrain the distribution of inputs for GANs, specifically their mean values. Since the statistical information of the dataset may contain some common features among samples, it might potentially aid in speeding up the convergence of training GANs but may not significantly enhance the accuracy of generated data after GAN optimization. It's worth noting that if the BN layer parameters uploaded by FedAvg could be used to constrain the statistical information of GANs' inputs, they would be equivalent to the information provided by our method.

Additionally, these attack methods against FedAvg only yield favorable results when class members are similar, meaning the dataset has clear common features that allow the constructed representatives to resemble the training data. When class members are dissimilar, these shared features tend to be confounded, rendering the constraints imposed by the sample mean ineffective, hence not enhancing the effectiveness of GANs attacks.

In summary, our method exhibits a security level consistent with FedAvg against these types of attacks. Even in cases where the BN layer is not required, our method's security is similar to that of FedAvg.

Membership inference attacks against aggregate statistics~\cite{fredrikson2015model,hitaj2017deep} and Membership inference attacks against ML models~\cite{dwork2015robust,pyrgelis2017knock,shokri2017membership,hayes202588705membership,long2018understanding,truex2018towards} aim to infer whether a sample belongs to the training dataset using appropriate prior distributions and statistical data. These attack methods impose specific requirements on the dataset. In such attack scenarios, whether the sample mean information our method can provide is exploitable by the attacker depends on whether this information can reveal the inherent distribution correlations within the dataset. However, for high-dimensional complex data, sample mean information often falls short in achieving this. 

The inference attack towards client model is a complex topic. Other inference attack methods and defense mechanisms against them fall outside this paper’s scope. It is an interesting topic to explore more robust measures to prevent such breaches in future works.

Therefore, in the case of these attacks we mentioned, our method exhibits the same level of security as FedAvg (since FedAvg requires uploading statistically equivalent information within the BN layer). For scenarios without a BN layer, whether our method reduces security depends on the characteristics of the dataset itself. Real-world data is often high-dimensional and complex, making successful attacks challenging.

\section{Additional experiments}
\label{appendx:exp}


\subsection{t-SNE visualization}
\label{append:tsne}
\begin{figure*}[t]
\subfloat[local client \#1]{
  \includegraphics[width=.21\textwidth]{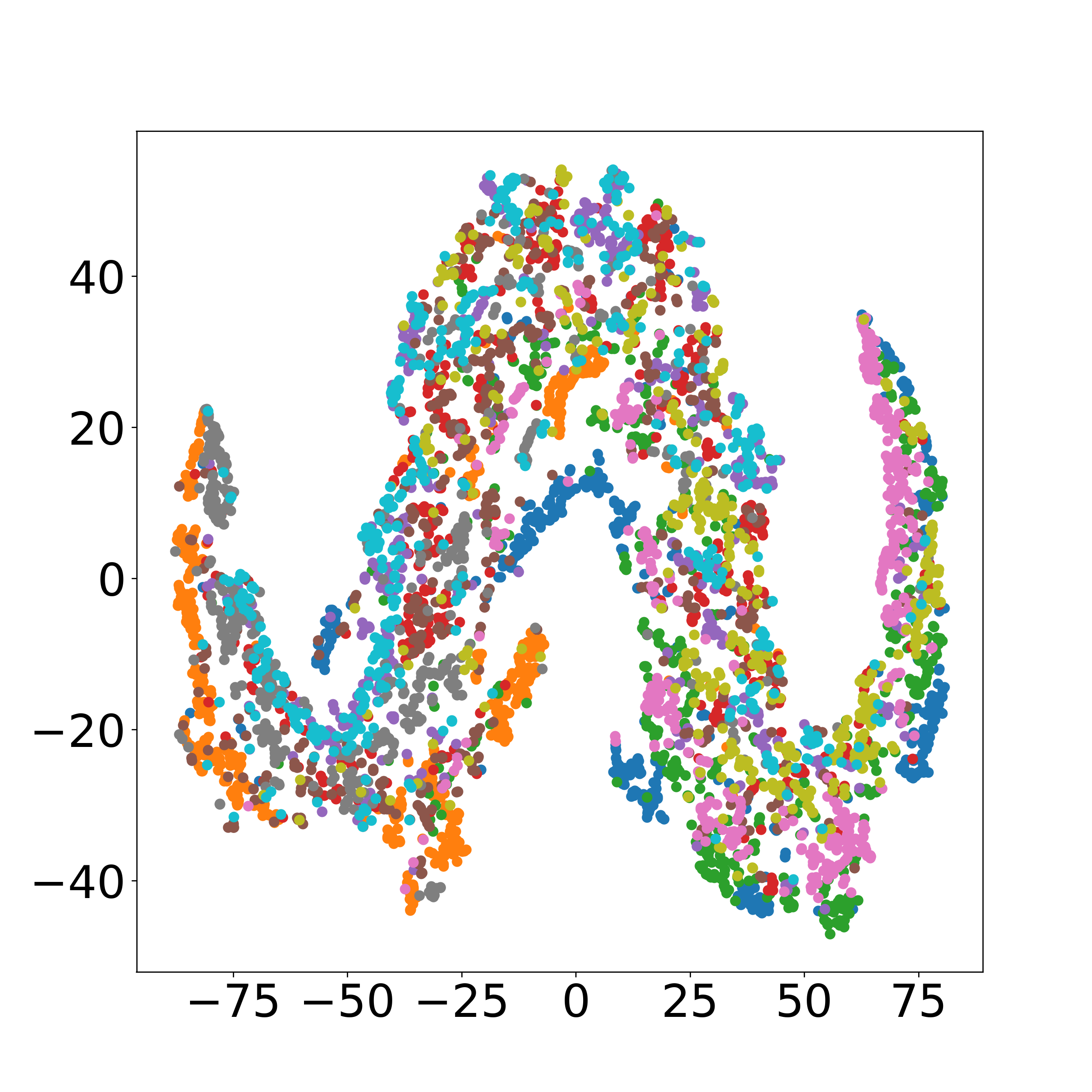}
    \label{pic1}}\hspace{-5.4mm}
\subfloat[local client \#2]{
  \includegraphics[width=.21\textwidth]{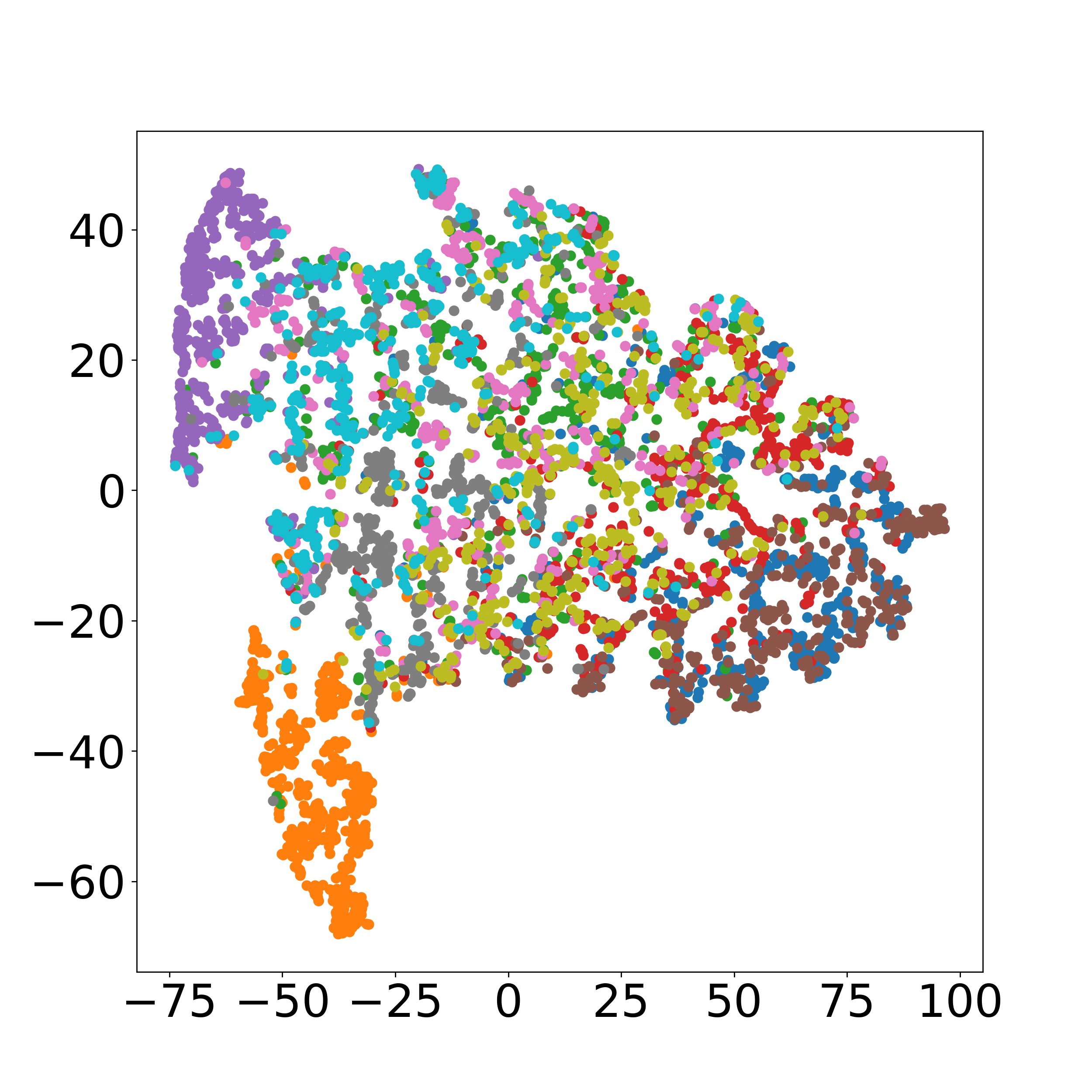}
  \label{pic2}}\hspace{-5.4mm}
\subfloat[local client \#3]{
  \includegraphics[width=.21\textwidth]{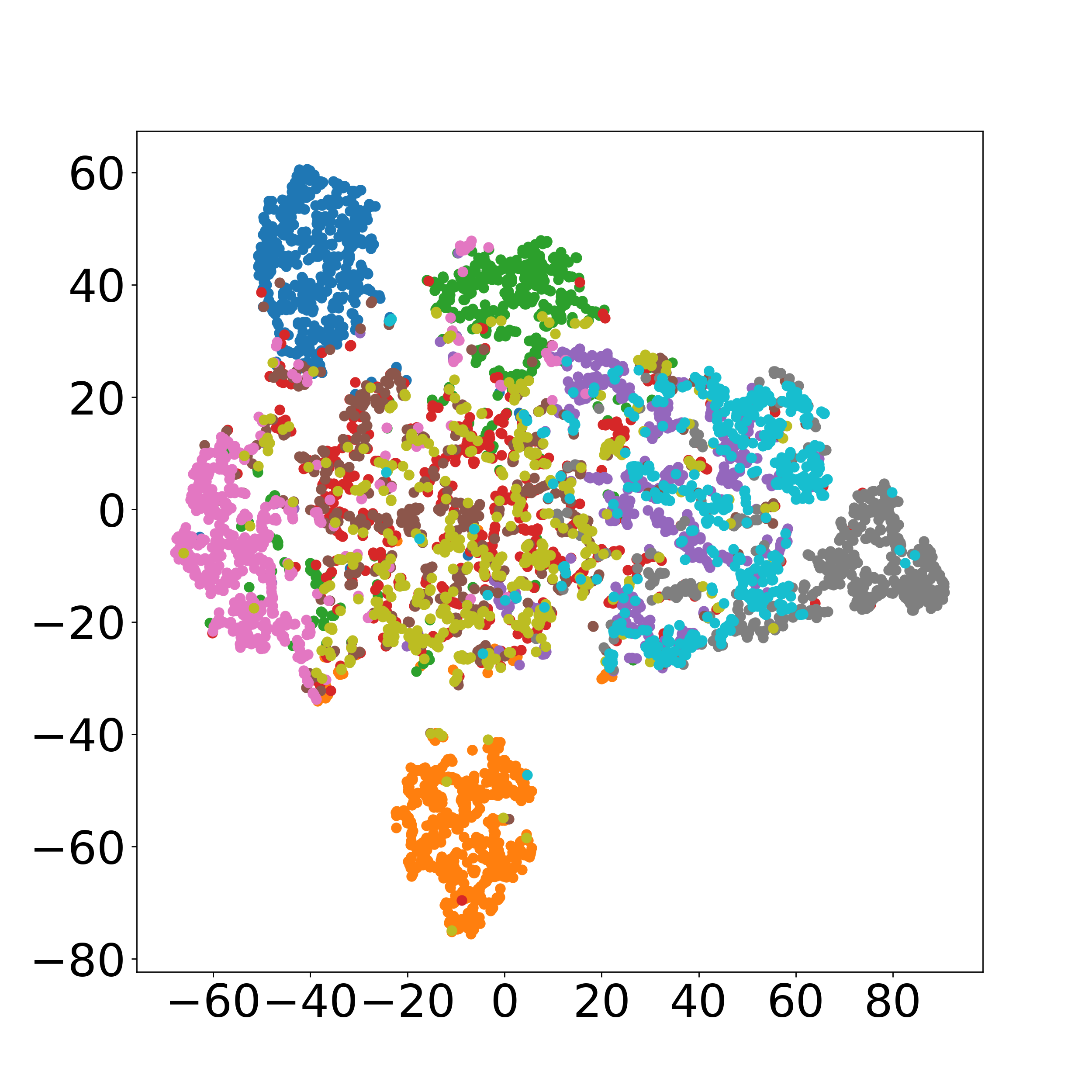}
    \label{pic3}}\hspace{-5.4mm}
\subfloat[local client \#4]{
  \includegraphics[width=.21\textwidth]{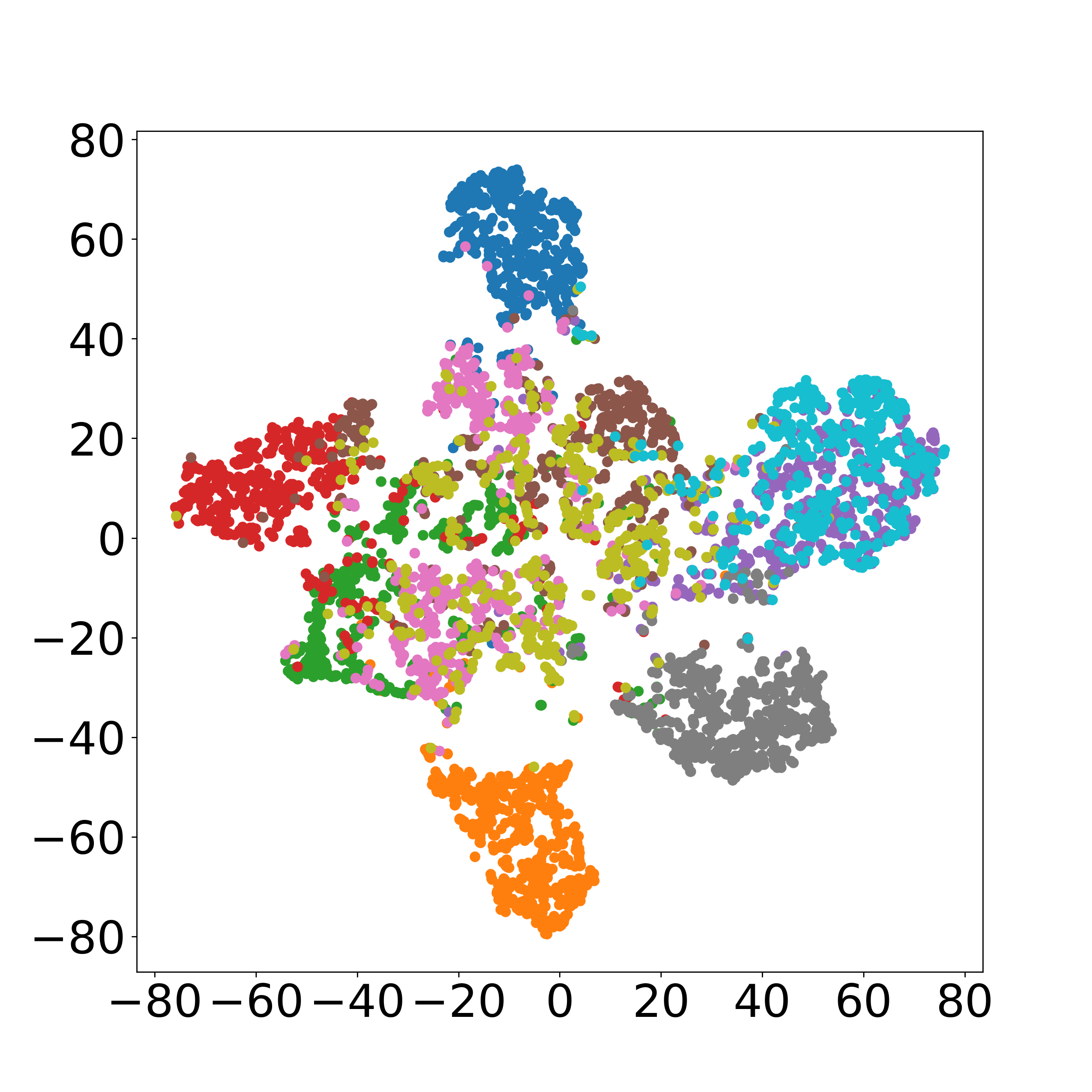}
  \label{pic4}}\hspace{-5.4mm}
\subfloat[local client \#5]{
  \includegraphics[width=.21\textwidth]{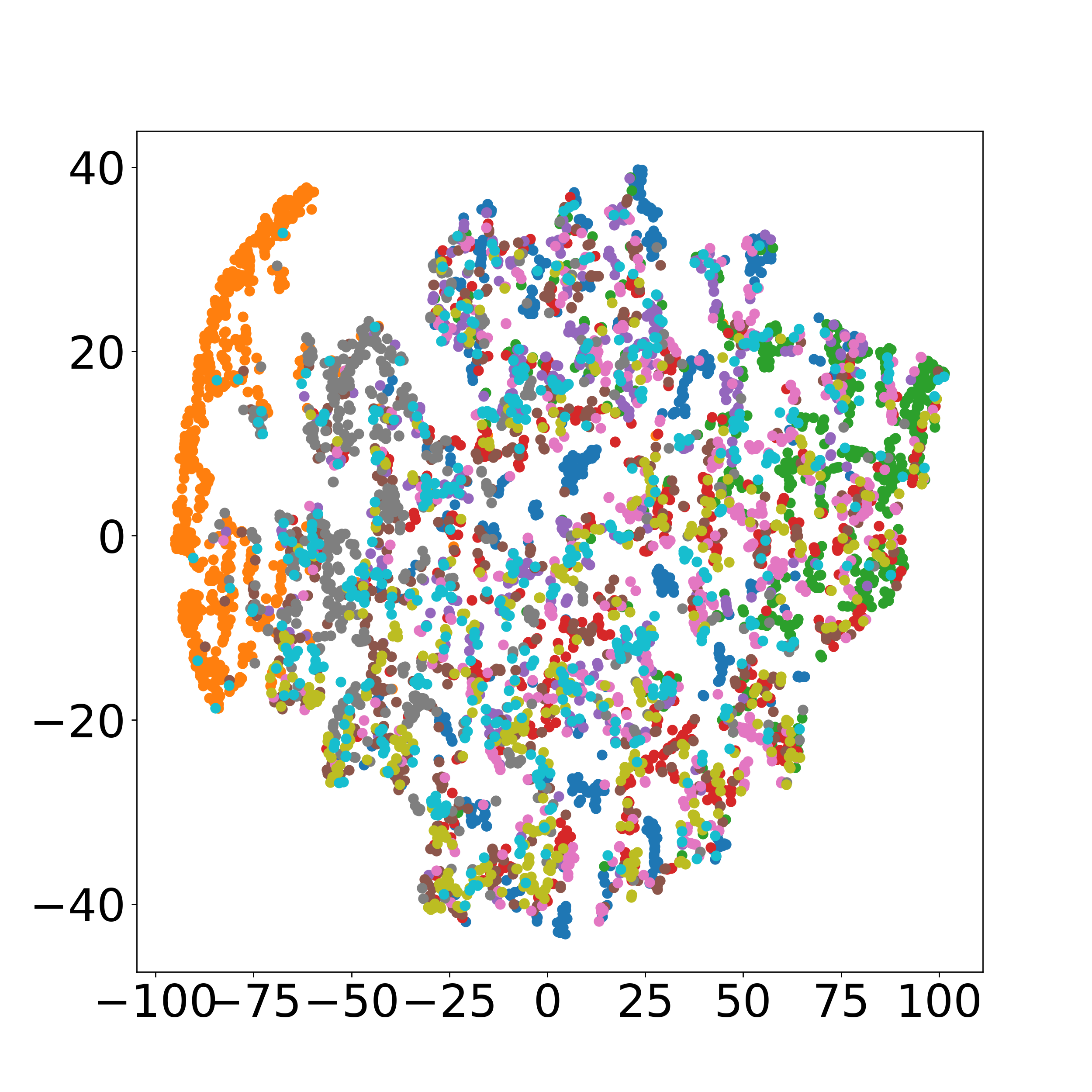}
    \label{pic5}}
    \\
\subfloat[local client \#6]{
  \includegraphics[width=.21\textwidth]{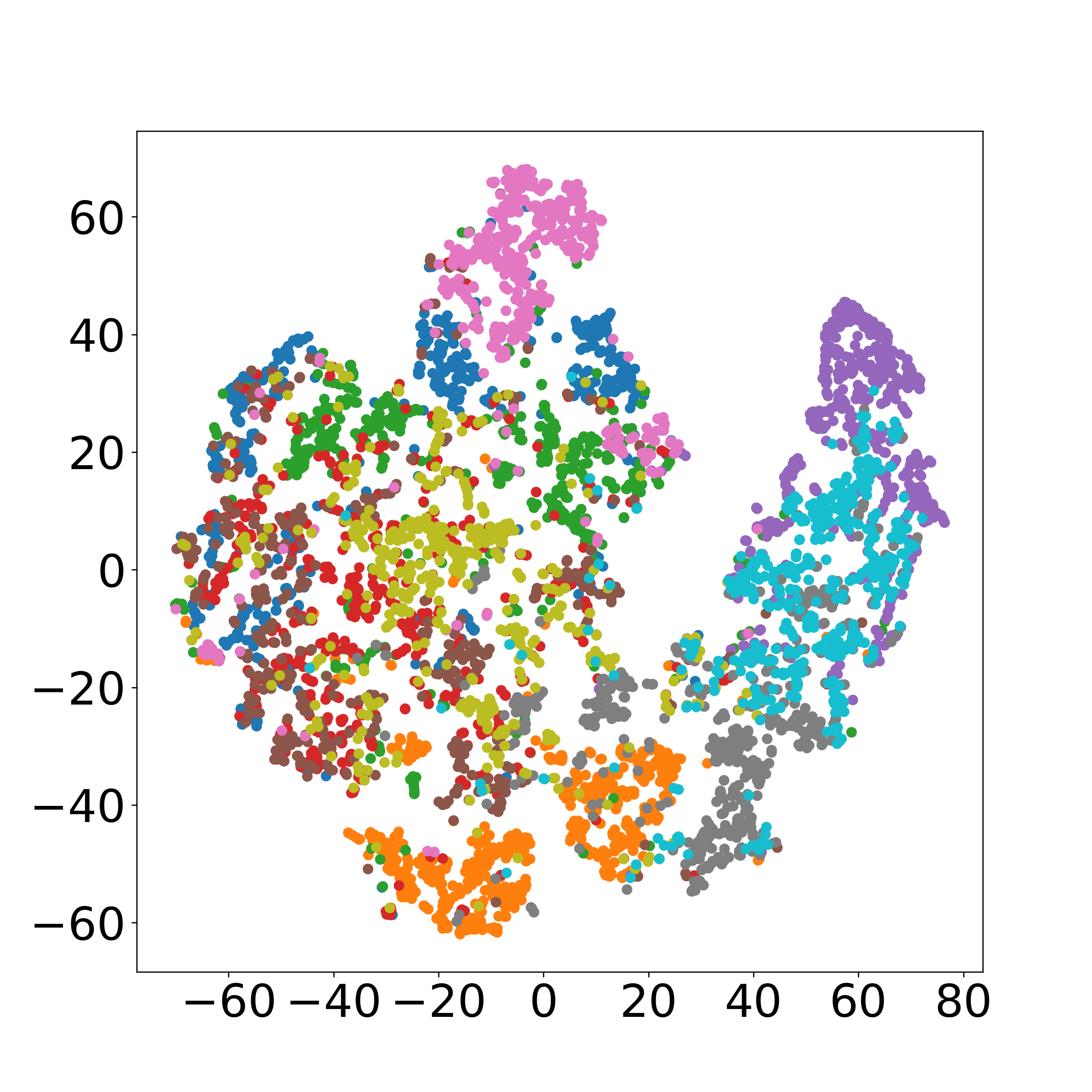}
  \label{pic6}}\hspace{-5.4mm}
\subfloat[local client \#7]{
  \includegraphics[width=.21\textwidth]{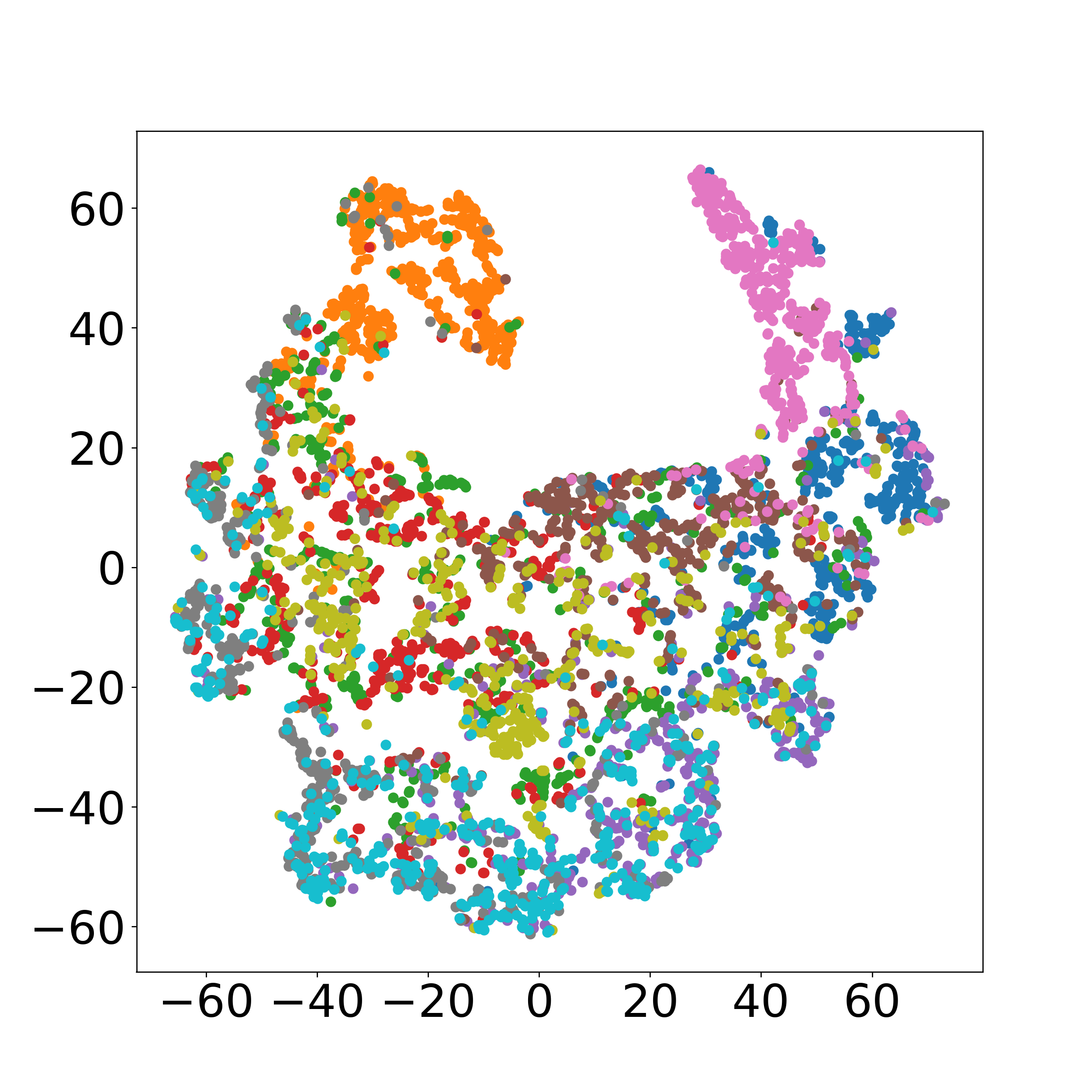}
    \label{pic7}}\hspace{-5.4mm}
\subfloat[local client \#8]{
  \includegraphics[width=.21\textwidth]{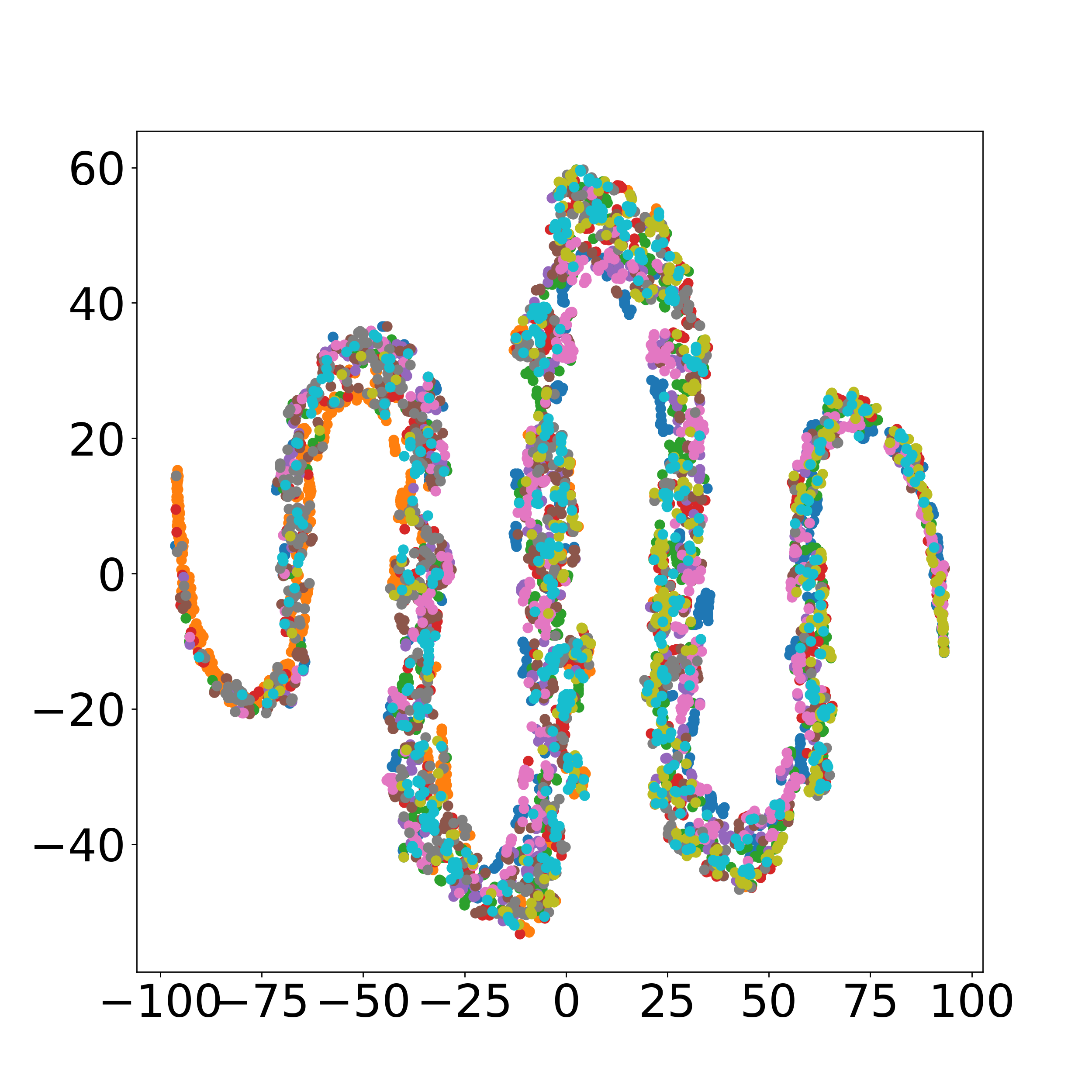}
  \label{pic8}}\hspace{-5.4mm}
\subfloat[local client \#9]{
  \includegraphics[width=.21\textwidth]{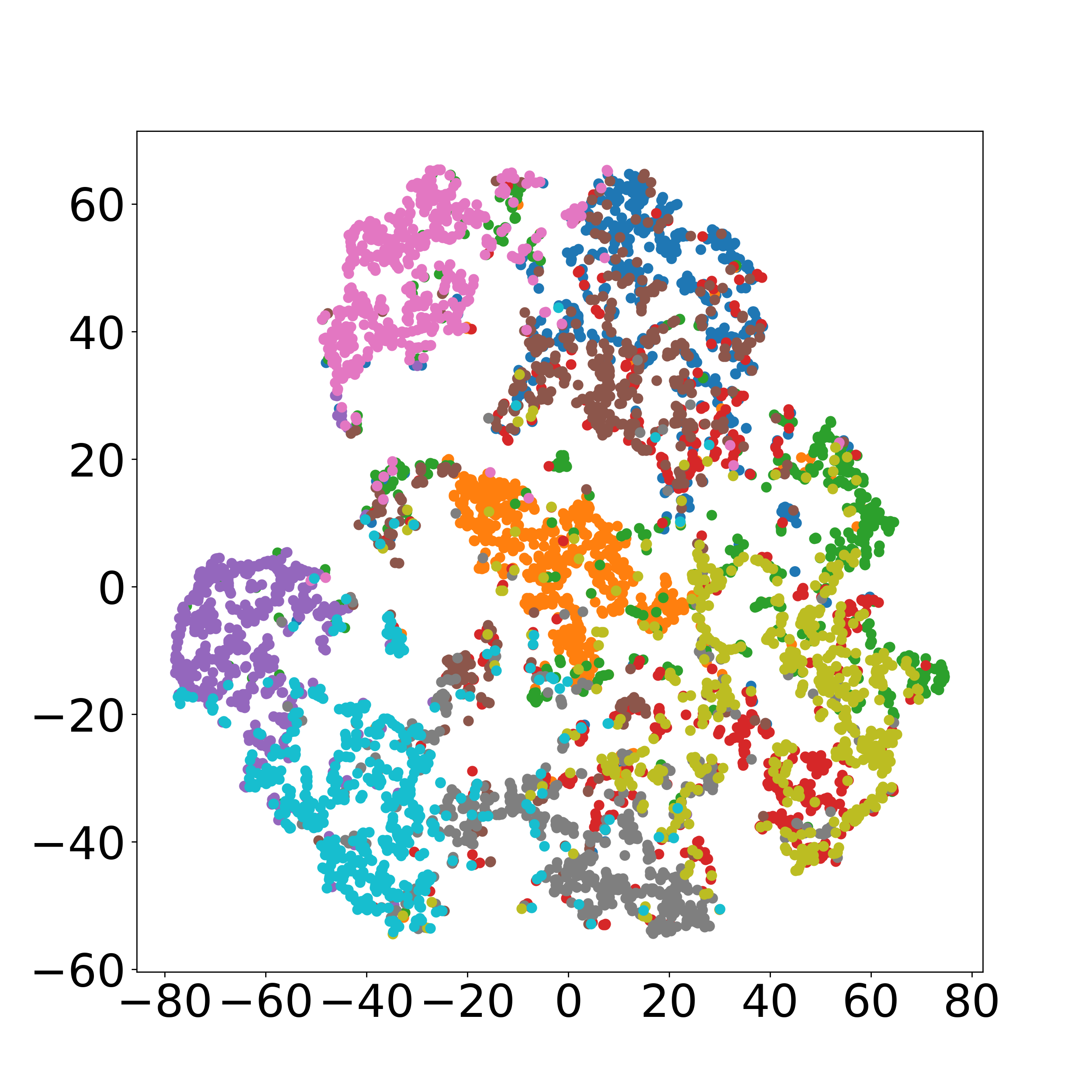}
    \label{pic9}}\hspace{-5.4mm}
\subfloat[local client \#10]{
  \includegraphics[width=.213\textwidth]{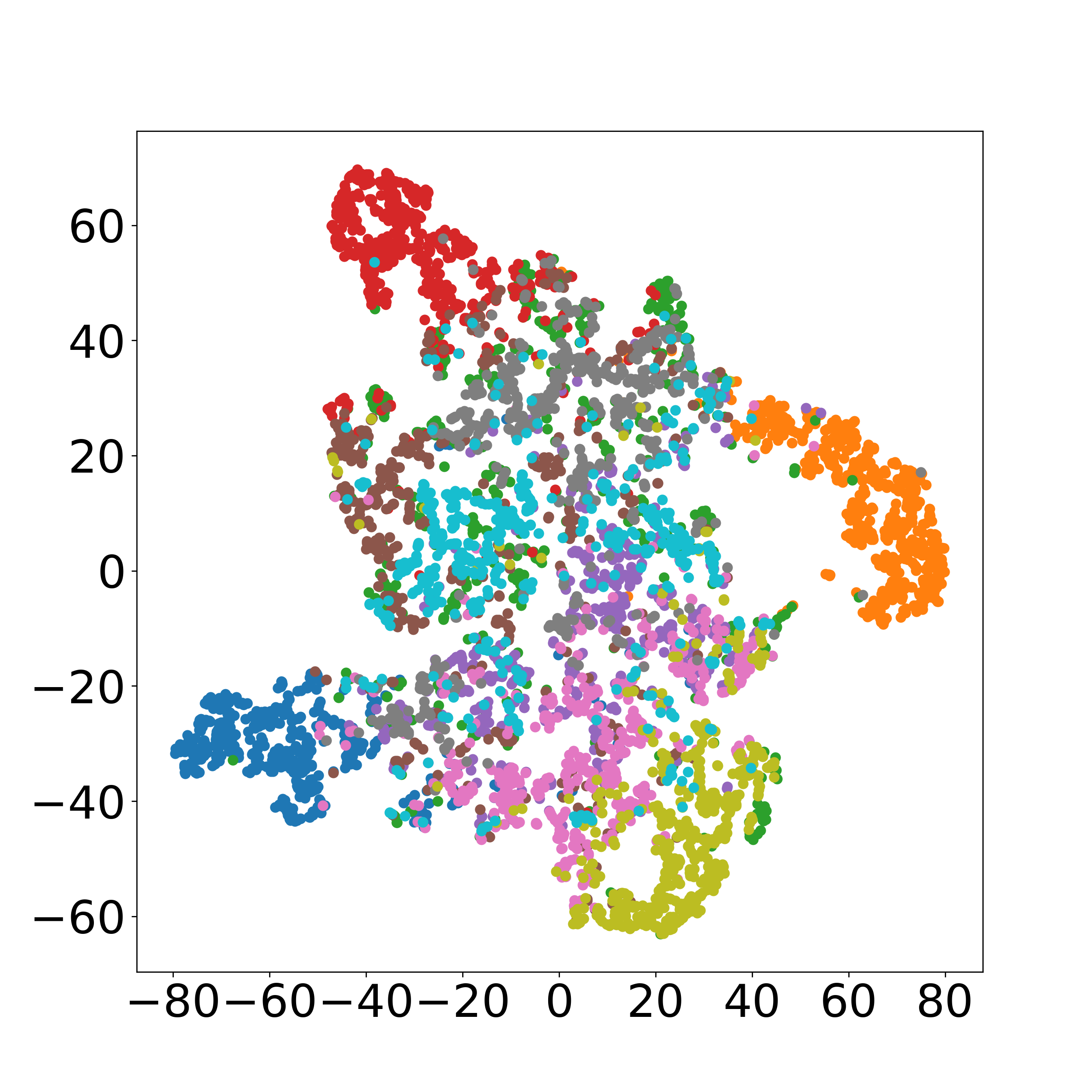}
  \label{pic10}}
\caption{t-SNE visualizations of 10 local clients.}
\label{fig:tsne_c}
\end{figure*}

\begin{figure*}[h]
  \subfloat[FedAvg]{
  \includegraphics[width=.21\textwidth]{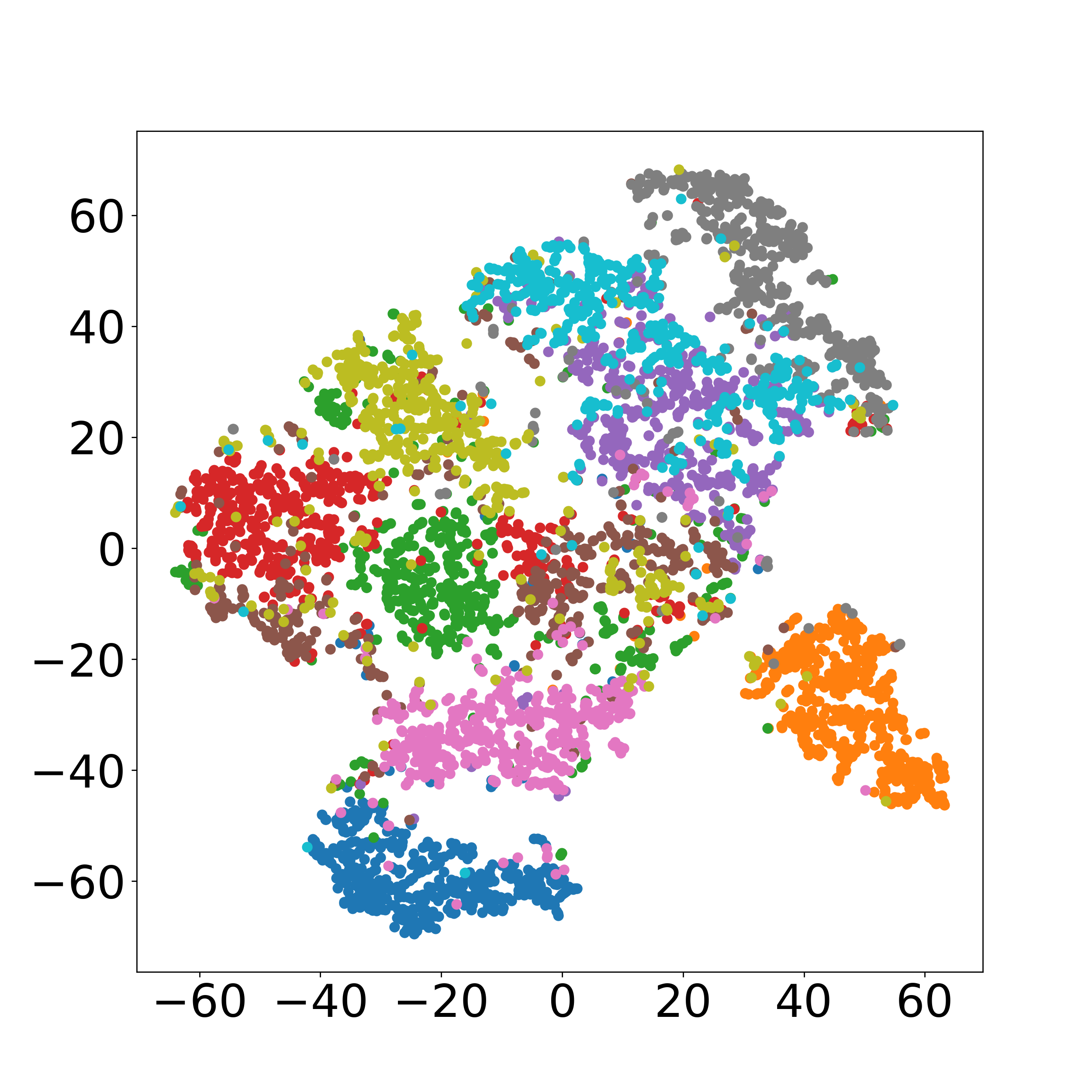}
  \label{picavg}}\hspace{-5.4mm}
\subfloat[FedNova]{
  \includegraphics[width=.21\textwidth]{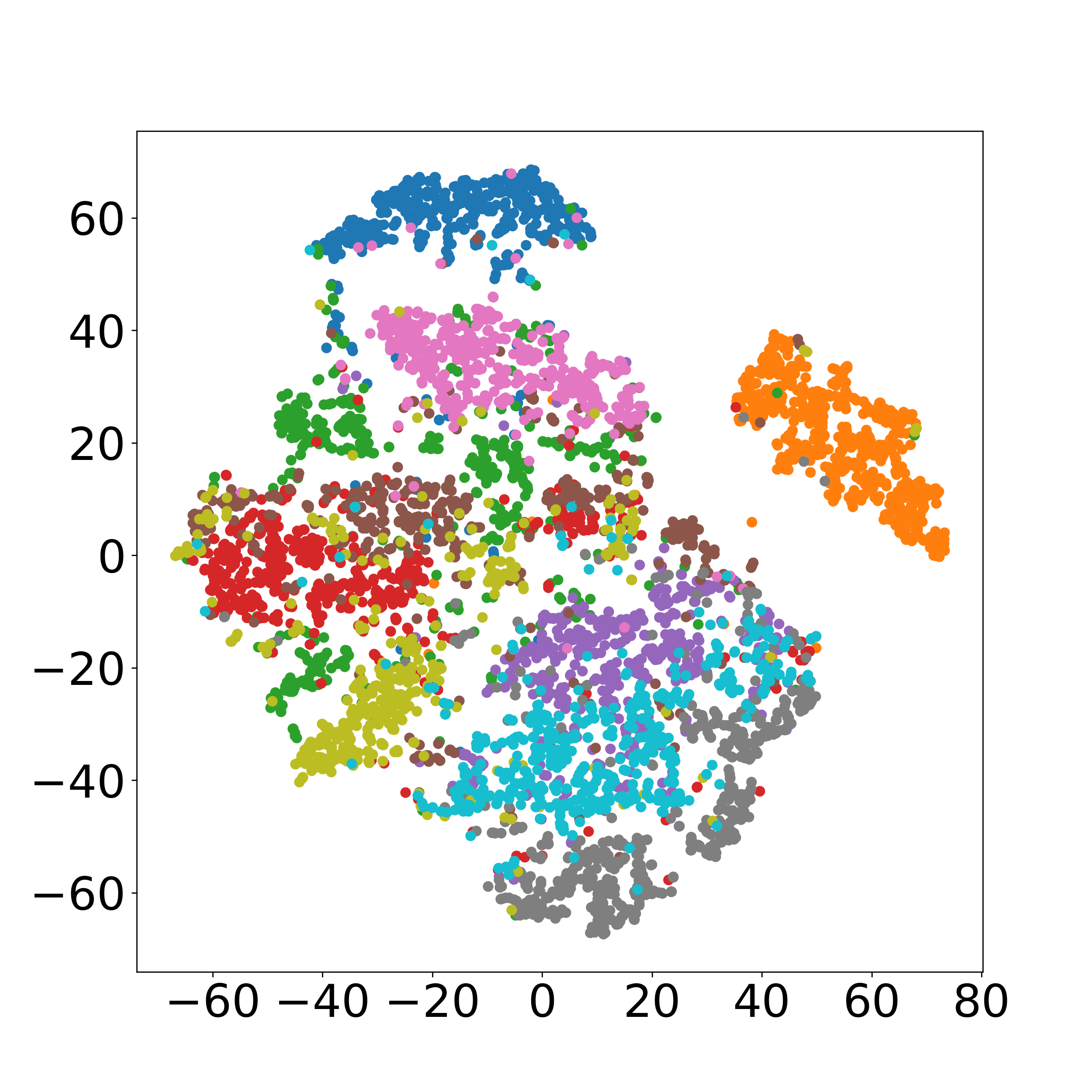}
  \label{picnova}}\hspace{-5.4mm}
\subfloat[SCAFFOLD]{
  \includegraphics[width=.21\textwidth]{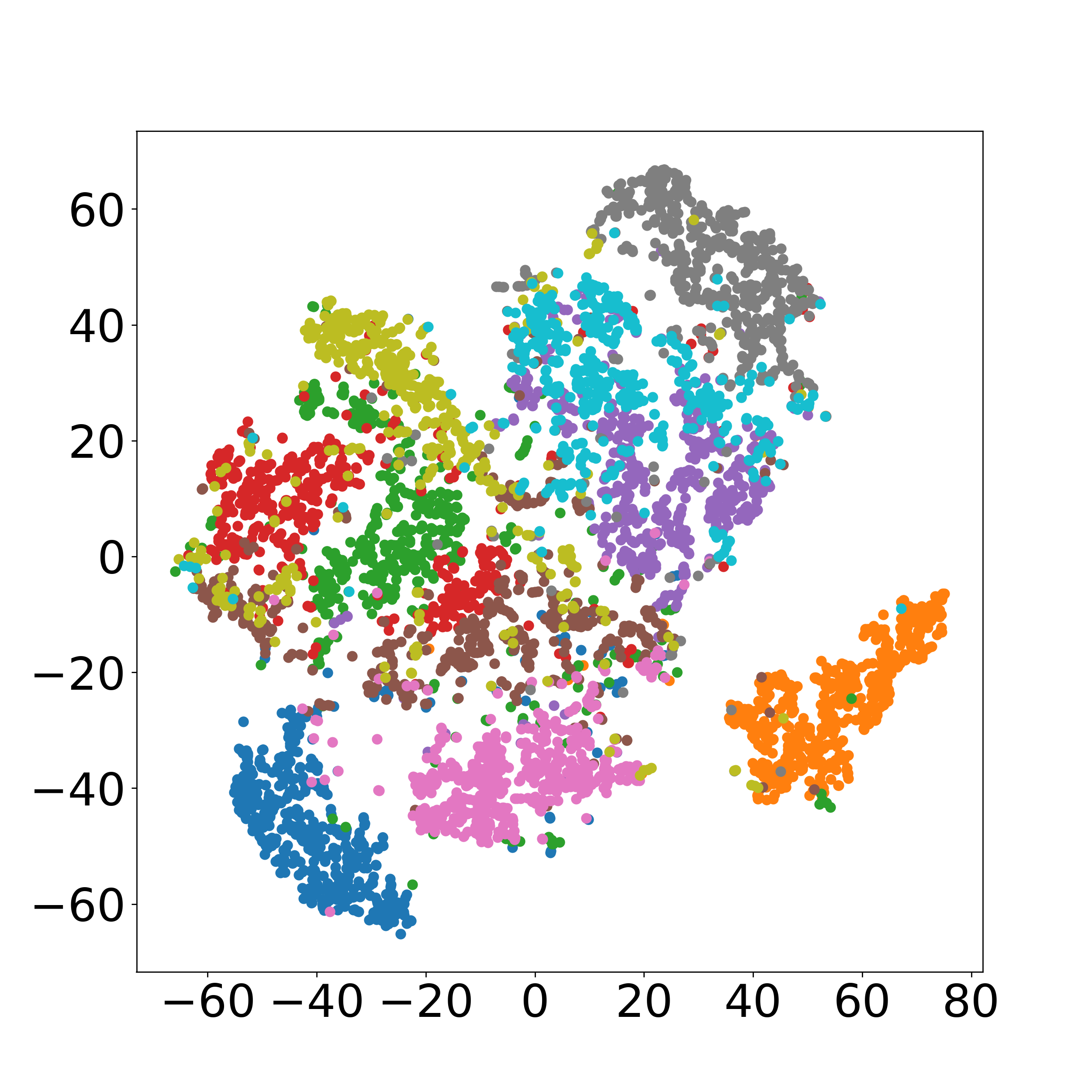}
  \label{picscaffold}}\hspace{-5.4mm}
\subfloat[FedProx]{
  \includegraphics[width=.21\textwidth]{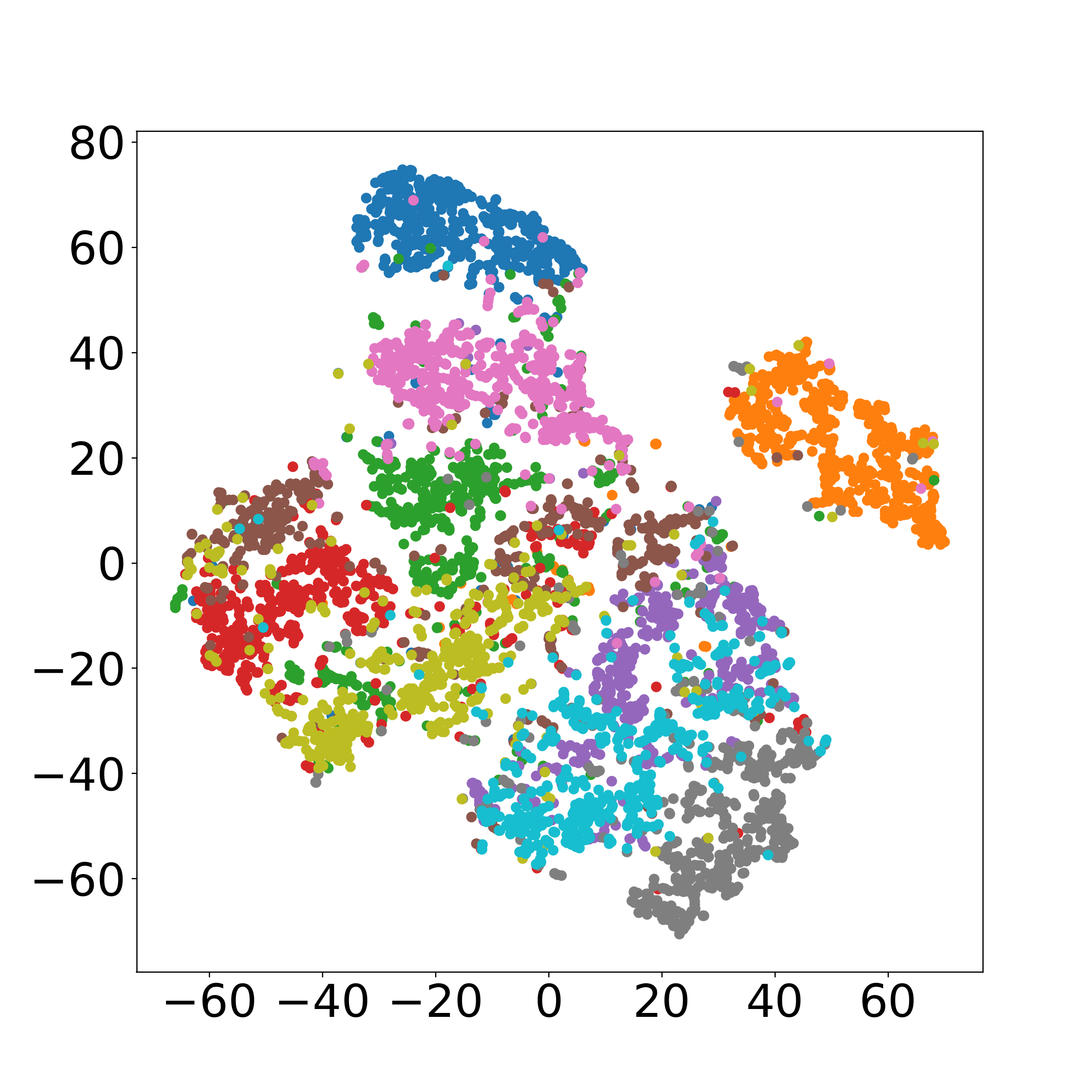}
  \label{picprox}}\hspace{-5.4mm}
\subfloat[DENSE]{
  \includegraphics[width=.21\textwidth]{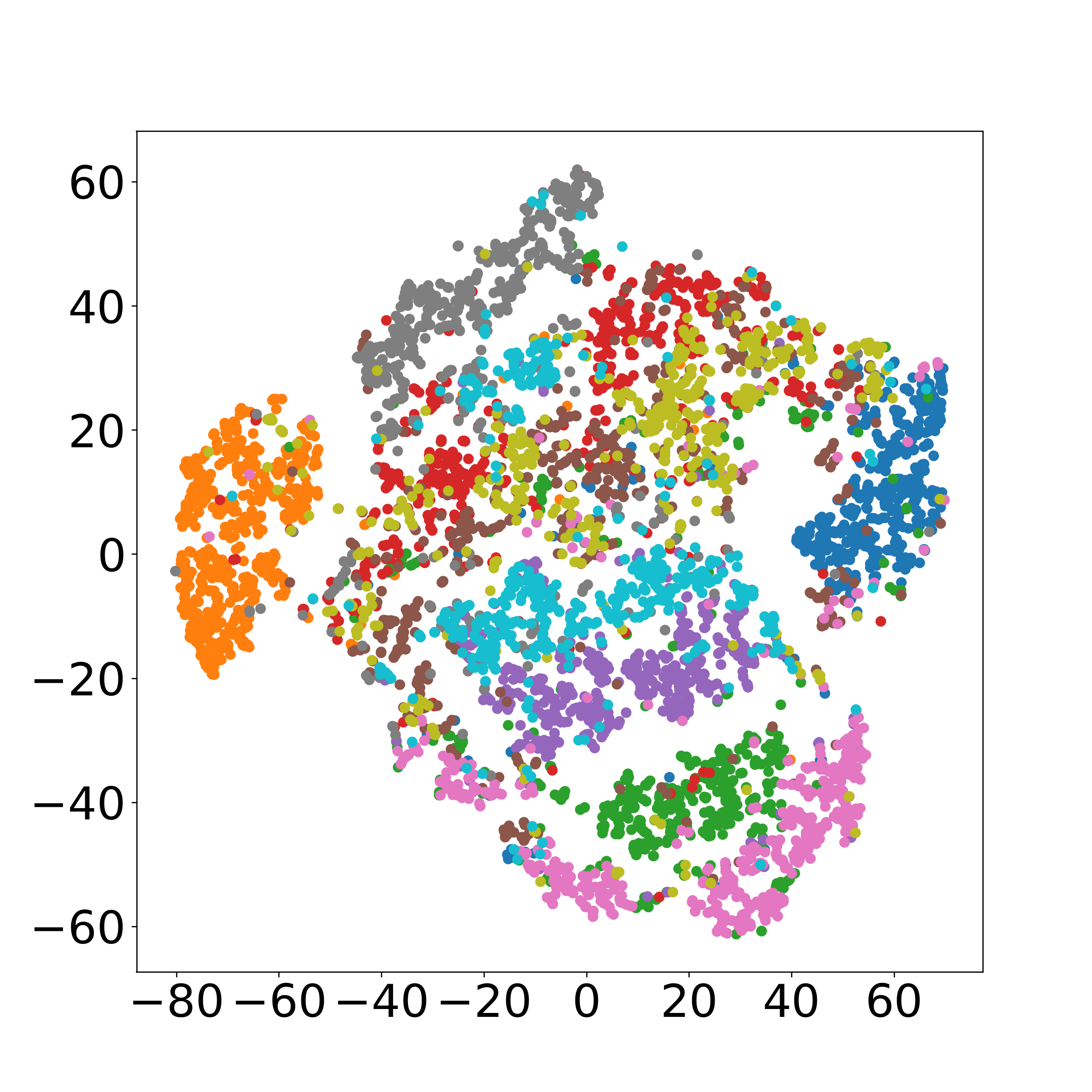}
  \label{picdense}}
\caption{t-SNE visualizations of the baseline approaches on the global model.}
\label{fig:tsne_g}
\end{figure*}

We conduct experiments using MNIST dataset with a $\beta$ value of 0.05, training 10 local clients over 200 local epochs with random seed 0. In this biased local dataset setting, local clients could only distinguish a subset of the classes, as illustrated in Figure~\ref{fig:tsne_c}. 

Based on seed 0, we partition the training data for the 10 local clients with the following form (label:\# of the data) as:

        local client \#1: \{4: 2, 5: 12, 6: 2847, 9: 16\}\\
        local client \#2: \{1: 20, 4: 189, 5: 5349\}\\
        local client \#3: \{0: 669, 1: 476, 2: 67, 6: 15, 7: 6068\}\\
        local client \#4: \{0: 266, 1: 375, 3: 3956, 7: 196, 9: 5932\}\\
        local client \#5: \{0: 4, 1: 418, 2: 5862\}\\
        local client \#6: \{1: 2, 2: 25, 4: 5195, 5: 24, 6: 80, 8: 28\}\\
        local client \#7: \{1: 5034, 2: 3, 4: 22, 5: 6, 6: 2669\}\\
        local client \#8: \{0: 4914\}\\
        local client \#9: \{4: 433, 5: 29, 6: 307, 8: 5373\}\\
        local client \#10: \{0: 70, 1: 417, 2: 1, 3: 2175, 4: 1, 5: 1, 7: 1, 8: 450, 9: 1\}\\

It is worth noting that local client \#2 has the training data mostly with label number 5, and as the corresponding t-SNE visualization shows in Figure~\ref{pic2}, the local train model could mainly cluster the data with label 5 (marked as purple). As data for label 1 (marked as orange) is different from other data with all other labels, some local clients may be able to cluster the data with label 1 with good results. Other local clients, such as local client \#3, \#4, \#6, \#7, \#9, \#10, show the similar results like local client \#2. 

Figure~\ref{fig:tsne_g} displays the t-SNE visualization for the global models of FedAvg, FedNova, SCAFFOLD, FedProx, and DENSE using the training data, with the figure legends identical to those in Figure~\ref{fig:tsne_our}. It's evident from Figure~\ref{fig:tsne_our} that FedLPA outperforms the baselines in classifying the ten classes. FedLPA's superiority is not only demonstrated by its ability to cluster the ten classes but also by the distinct separation between classes, as observed in Figure~\ref{fig:tsne_our}, compared to the baselines.

\subsection{Experiments on different local epoch numbers}
\label{appendx:exp:epochs}

\begin{table}[!h]\tiny 
\centering
\caption{ Comparison with various FL algorithms in one round with 10 local epochs settings.}
\label{table:10}
\setlength\tabcolsep{3.7mm}
\begin{tabular}{c|c|c|c|c|c|c|c}
\toprule[0.15em]
Dataset&Partition & FedLPA & FedNova & SCAFFOLD & FedAvg & FedProx &DENSE \\
\midrule[0.1em]
\multirow{9}{*}{FMNIST} &$\beta$=0.01& \cellcolor{green!25}24.47$\pm$1.02 &11.43$\pm$0.04& 13.37$\pm$0.11& 11.83$\pm$0.03& 12.30$\pm$0.10& 10.00$\pm$0.00\\
&$\beta$=0.05  & \cellcolor{green!25} 30.77$\pm$0.94&15.67$\pm$0.28&	20.07$\pm$0.46&	19.93$\pm$0.30&	20.07$\pm$0.37&	23.60$\pm$0.14 \\
& $\beta$=0.1& \cellcolor{green!25}42.83$\pm$0.33 &25.10$\pm$1.32&	21.03$\pm$1.08&	22.57$\pm$1.08&	22.20$\pm$1.17&	34.83$\pm$0.16 \\
&$\beta$=0.3 & \cellcolor{green!25}61.43$\pm$0.17 &40.90$\pm$0.05&	40.70$\pm$0.09&	38.20$\pm$0.10&	37.50$\pm$0.03&	43.17$\pm$0.05 \\
& $\beta$=0.5& \cellcolor{green!25}67.63$\pm$0.36 &52.43$\pm$0.60&	51.77$\pm$0.46&	54.67$\pm$0.73&	54.33$\pm$0.64&	54.30$\pm$0.07\\
&$\beta$=1.0& \cellcolor{green!25} 71.90$\pm$0.09&51.03$\pm$0.62&	52.30$\pm$0.53&	51.50$\pm$0.62&	50.90$\pm$0.57&	52.30$\pm$0.15 \\
&\#C=1 &\cellcolor{green!25}13.03$\pm$0.04 &10.90$\pm$0.02&	11.27$\pm$0.03&	10.90$\pm$0.02&	11.43$\pm$0.04&	10.00$\pm$0.00\\
&\#C=2 &\cellcolor{green!25} 28.93$\pm$0.74&16.93$\pm$0.19&	22.07$\pm$0.01&	23.83$\pm$0.20&	23.33$\pm$0.03&	22.60$\pm$0.88\\
&\#C=3 & \cellcolor{green!25} 37.73$\pm$0.09&22.20$\pm$0.23&	26.60$\pm$0.03&	23.67$\pm$0.27&	22.80$\pm$0.40&	23.03$\pm$0.86\\
\midrule[0.1em]
\multirow{9}{*}{CIFAR-10} &$\beta$=0.01 & \cellcolor{green!25}15.80$\pm$0.00 &10.07$\pm$0.00&	12.13$\pm$0.09&	11.90$\pm$0.07&	11.93$\pm$0.07&	10.00$\pm$0.00 \\
&$\beta$=0.05  & \cellcolor{green!25}20.23$\pm$0.01 &10.90$\pm$0.02&	10.00$\pm$0.00&	10.00$\pm$0.00&	10.00$\pm$0.00&	13.33$\pm$0.04\\
& $\beta$=0.1& \cellcolor{green!25}20.20$\pm$0.07 &10.27$\pm$0.00&	10.93$\pm$0.02&	10.37$\pm$0.00&	10.27$\pm$0.00&	14.77$\pm$0.09 \\
&$\beta$=0.3 & \cellcolor{green!25}25.60$\pm$0.01 &18.13$\pm$0.33&	14.97$\pm$0.05&	14.77$\pm$0.05&	15.67$\pm$0.03&	20.33$\pm$0.06\\
& $\beta$=0.5& \cellcolor{green!25} 25.60$\pm$0.08&14.87$\pm$0.05&	16.77$\pm$0.01&	15.73$\pm$0.01&	13.93$\pm$0.08&	23.20$\pm$0.16 \\
&$\beta$=1.0& \cellcolor{green!25}28.93$\pm$0.01 &15.63$\pm$0.03&	19.10$\pm$0.14&	15.30$\pm$0.05&	15.43$\pm$0.05&	22.30$\pm$0.46 \\
&\#C=1 &\cellcolor{green!25}11.00$\pm$0.01 &10.30$\pm$0.00&	10.23$\pm$0.00&	10.30$\pm$0.00&	10.33$\pm$0.00&	10.00$\pm$0.00\\
&\#C=2 &\cellcolor{green!25}20.40$\pm$0.03 &11.37$\pm$0.04&	11.67$\pm$0.06&	11.00$\pm$0.02&	11.80$\pm$0.06&	10.40$\pm$0.00\\
&\#C=3 & \cellcolor{green!25} 22.30$\pm$0.03&12.23$\pm$0.04&	14.37$\pm$0.13&	14.10$\pm$0.14&	14.00$\pm$0.12&	18.50$\pm$0.14\\
\midrule[0.1em]
\multirow{9}{*}{MNIST}  &$\beta$=0.01 & \cellcolor{green!25}32.20$\pm$0.50 &9.53$\pm$0.00&	9.37$\pm$0.00&	9.00$\pm$0.01&	9.40$\pm$0.00&	9.53$\pm$0.00 \\
&$\beta$=0.05  & \cellcolor{green!25}60.60$\pm$0.07 &20.80$\pm$0.13&	35.17$\pm$0.66&	35.10$\pm$0.87&	34.13$\pm$0.91&	50.37$\pm$1.57 \\
& $\beta$=0.1& \cellcolor{green!25} 78.07$\pm$0.09&45.07$\pm$0.37&	43.23$\pm$0.10&	43.83$\pm$0.13&	44.27$\pm$0.21&	65.53$\pm$0.85 \\
&$\beta$=0.3 & \cellcolor{green!25}85.60$\pm$0.17 &64.40$\pm$0.24&	64.03$\pm$0.11&	64.17$\pm$0.09&	64.07$\pm$0.11&	75.53$\pm$0.22 \\
& $\beta$=0.5& \cellcolor{green!25}91.77$\pm$0.00 &79.43$\pm$0.13&	77.37$\pm$0.22&	78.17$\pm$0.25&	77.90$\pm$0.30&	87.93$\pm$0.17 \\
&$\beta$=1.0& \cellcolor{green!25}94.70$\pm$0.00 &85.00$\pm$0.10&	85.10$\pm$0.06&	84.40$\pm$0.08&	84.63$\pm$0.08&	89.30$\pm$0.03\\
&\#C=1 &\cellcolor{green!25} 11.87$\pm$0.00&10.43$\pm$0.02&	10.13$\pm$0.01&	10.13$\pm$0.01&	10.13$\pm$0.01&	9.93$\pm$0.00\\
&\#C=2  &\cellcolor{green!25}47.93$\pm$0.89 &13.20$\pm$0.09&	16.47$\pm$0.21&	12.97$\pm$0.16&	12.23$\pm$0.07&	32.57$\pm$0.26\\
&\#C=3 & \cellcolor{green!25}65.97$\pm$0.98 &26.70$\pm$2.24&	31.67$\pm$2.60&	31.63$\pm$3.03&	31.20$\pm$3.24&	53.80$\pm$0.09\\
\midrule[0.1em]
\multirow{9}{*}{SVHN} &$\beta$=0.01 & \cellcolor{green!25} 17.00$\pm$0.03&13.93$\pm$0.16&	16.57$\pm$0.15&	16.27$\pm$0.22&	13.30$\pm$0.20&	13.97$\pm$0.17\\
&$\beta$=0.05  & \cellcolor{green!25}20.23$\pm$0.05 &15.40$\pm$0.11&	15.53$\pm$0.12&	15.53$\pm$0.12&	15.53$\pm$0.12&	17.90$\pm$1.01\\
& $\beta$=0.1& \cellcolor{green!25}32.57$\pm$0.53 &15.17$\pm$0.18&	18.37$\pm$0.03&	18.37$\pm$0.03&	18.33$\pm$0.03&	24.20$\pm$0.28 \\
&$\beta$=0.3 & \cellcolor{green!25}35.47$\pm$0.54 &18.23$\pm$0.29&	20.77$\pm$0.02&	21.63$\pm$0.03&	21.17$\pm$0.01&	29.23$\pm$0.03\\
& $\beta$=0.5& \cellcolor{green!25}41.17$\pm$0.01 &26.07$\pm$0.13&	27.40$\pm$0.00&	26.27$\pm$0.00&	27.80$\pm$0.00&	36.80$\pm$0.13\\
&$\beta$=1.0& \cellcolor{green!25}44.33$\pm$0.01 &30.77$\pm$0.01&	32.27$\pm$0.01&	30.43$\pm$0.03&	31.97$\pm$0.00&	29.47$\pm$2.86\\
&\#C=1 &\cellcolor{green!25}19.60$\pm$0.00 &10.10$\pm$0.03&	16.60$\pm$0.11&	16.77$\pm$0.13&	15.53$\pm$0.12&	8.90$\pm$0.02\\
&\#C=2  &\cellcolor{green!25}31.20$\pm$0.01 &11.80$\pm$0.33&	15.77$\pm$0.29&	15.67$\pm$0.31&	15.60$\pm$0.32&	14.00$\pm$0.24\\
&\#C=3 & \cellcolor{green!25}34.43$\pm$0.40 &8.93$\pm$0.01&	22.03$\pm$0.05&	18.03$\pm$0.12&	17.50$\pm$0.17&	23.57$\pm$0.90\\
\hline
\bottomrule[0.1em]
\end{tabular}
\end{table}

\begin{table}[!h]\tiny
\centering
\caption{ Comparison with various FL algorithms in one round with 20 local epochs settings.}
\label{table:20}
\setlength\tabcolsep{3.7mm}
\begin{tabular}{c|c|c|c|c|c|c|c}
\toprule[0.15em]
Dataset&Partition & FedLPA & FedNova & SCAFFOLD & FedAvg & FedProx &DENSE \\
\midrule[0.1em]
\multirow{9}{*}{FMNIST} &$\beta$=0.01 & \cellcolor{green!25} 24.17$\pm$1.13&11.90$\pm$0.07&	15.33$\pm$0.15&	13.40$\pm$0.09&	10.57$\pm$0.00&	12.67$\pm$0.14\\
&$\beta$=0.05  & \cellcolor{green!25}36.97$\pm$0.78 &18.83$\pm$0.55&	19.93$\pm$0.11&	19.37$\pm$0.14&	19.70$\pm$0.17&	33.13$\pm$0.47 \\
& $\beta$=0.1& \cellcolor{green!25}41.83$\pm$0.02 &28.13$\pm$1.34&	23.00$\pm$0.41&	24.63$\pm$0.75&	24.10$\pm$1.29&	36.40$\pm$0.02 \\
&$\beta$=0.3 & \cellcolor{green!25} 60.83$\pm$0.38&42.50$\pm$0.12&	42.83$\pm$0.01&	40.47$\pm$0.22&	40.63$\pm$0.11&	40.67$\pm$0.04\\
& $\beta$=0.5& \cellcolor{green!25} 67.80$\pm$0.25&53.17$\pm$0.25&	55.23$\pm$0.67&	53.27$\pm$0.34&	54.20$\pm$0.52&	64.60$\pm$0.45\\
&$\beta$=1.0& \cellcolor{green!25} 75.47$\pm$0.03&55.47$\pm$0.57&	54.53$\pm$0.52&	53.57$\pm$0.46&	54.73$\pm$0.41&	70.97$\pm$0.02 \\
&\#C=1 &\cellcolor{green!25}14.07$\pm$0.02 &10.43$\pm$0.00&	11.03$\pm$0.02&	10.43$\pm$0.00&	11.13$\pm$0.03&	10.00$\pm$0.00\\
&\#C=2 &\cellcolor{green!25} 29.67$\pm$0.37&16.83$\pm$0.21&	22.67$\pm$0.27&	23.27$\pm$0.24&	25.43$\pm$0.17&	24.77$\pm$0.14 \\
&\#C=3 & \cellcolor{green!25} 34.37$\pm$0.79&24.93$\pm$0.02&	26.17$\pm$0.01&	27.70$\pm$0.02&	27.70$\pm$0.03&	29.43$\pm$0.22\\
\midrule[0.1em]
\multirow{9}{*}{CIFAR-10} &$\beta$=0.01 & \cellcolor{green!25}15.13$\pm$0.01 &10.10$\pm$0.00&	12.50$\pm$0.11&	11.90$\pm$0.07&	11.83$\pm$0.07&	10.50$\pm$0.00 \\
&$\beta$=0.05  & \cellcolor{green!25}23.37$\pm$0.01 &11.33$\pm$0.04&	10.20$\pm$0.00&	10.00$\pm$0.00&	10.77$\pm$0.01&	14.67$\pm$0.02\\
& $\beta$=0.1& \cellcolor{green!25} 25.07$\pm$0.00&11.60$\pm$0.05&	12.33$\pm$0.10&	12.67$\pm$0.14&	13.23$\pm$0.21&	18.50$\pm$0.06 \\
&$\beta$=0.3 & \cellcolor{green!25} 26.00$\pm$0.02&16.63$\pm$0.13&	13.10$\pm$0.01&	13.87$\pm$0.01&	14.43$\pm$0.02&	24.97$\pm$1.13 \\
& $\beta$=0.5& \cellcolor{green!25} 30.60$\pm$0.00&14.20$\pm$0.02&	13.30$\pm$0.04&	13.13$\pm$0.03&	14.23$\pm$0.04&	27.60$\pm$0.06\\
&$\beta$=1.0& \cellcolor{green!25}26.77$\pm$0.10 &17.60$\pm$0.01&	17.50$\pm$0.05&	18.13$\pm$0.04&	18.20$\pm$0.01&	26.07$\pm$0.18\\
&\#C=1 &\cellcolor{green!25} 10.67$\pm$0.01&10.20$\pm$0.00&	10.23$\pm$0.00&	10.20$\pm$0.00&	10.27$\pm$0.00&	10.00$\pm$0.00\\
&\#C=2 &\cellcolor{green!25}22.00$\pm$0.00 &12.03$\pm$0.08&	10.23$\pm$0.00&	11.10$\pm$0.02&	11.40$\pm$0.04&	15.33$\pm$0.14\\
&\#C=3 & \cellcolor{green!25}23.60$\pm$0.05 &11.97$\pm$0.03&	14.93$\pm$0.17&	13.37$\pm$0.09&	13.63$\pm$0.09&	21.17$\pm$0.05 \\
\midrule[0.1em]
\multirow{9}{*}{MNIST}  &$\beta$=0.01  & \cellcolor{green!25}32.43$\pm$0.86 &11.00$\pm$0.06&	9.30$\pm$0.00&	9.37$\pm$0.00&	10.33$\pm$0.02&	12.40$\pm$0.19 \\
& $\beta$=0.05& \cellcolor{green!25} 68.73$\pm$0.45&26.73$\pm$0.24&	37.77$\pm$0.68&	37.70$\pm$0.84&	36.57$\pm$0.75&	62.03$\pm$0.54 \\
&$\beta$=0.1 & \cellcolor{green!25}71.77$\pm$0.20 &48.57$\pm$0.60&	45.67$\pm$0.22&	46.63$\pm$0.23&	45.83$\pm$0.23&	66.93$\pm$0.25 \\
& $\beta$=0.3& \cellcolor{green!25}90.83$\pm$0.01 &68.17$\pm$0.33&	66.90$\pm$0.03&	66.60$\pm$0.15&	66.03$\pm$0.21&	85.37$\pm$0.11 \\
&$\beta$=0.5& \cellcolor{green!25}89.43$\pm$0.09 &80.90$\pm$0.14&	76.63$\pm$0.13&	79.57$\pm$0.22&	79.47$\pm$0.27&	86.07$\pm$0.22 \\
&1.0 &\cellcolor{green!25}96.17$\pm$0.01 &86.60$\pm$0.13&	85.90$\pm$0.14&	86.03$\pm$0.14&	86.57$\pm$0.15&	88.40$\pm$0.00 \\
&\#C=1  &\cellcolor{green!25}11.47$\pm$0.01 &10.27$\pm$0.01&	10.13$\pm$0.01&	10.13$\pm$0.01&	10.13$\pm$0.01&	9.87$\pm$0.00\\
&\#C=2 & \cellcolor{green!25}53.37$\pm$0.61 &17.47$\pm$0.48&	20.70$\pm$0.58&	14.77$\pm$0.14&	13.47$\pm$0.02&	43.33$\pm$0.10 \\
&\#C=3 & \cellcolor{green!25}72.27$\pm$0.44 &28.63$\pm$1.61&	32.93$\pm$2.76&	31.40$\pm$2.01&	30.97$\pm$2.50&	44.30$\pm$0.62 \\
\midrule[0.1em]
\multirow{9}{*}{SVHN} &$\beta$=0.01 & \cellcolor{green!25}19.03$\pm$0.00 &14.83$\pm$0.13&	9.33$\pm$0.02&	9.30$\pm$0.02&	9.30$\pm$0.02&	18.23$\pm$0.03\\
&$\beta$=0.05  & \cellcolor{green!25}26.27$\pm$0.19 &13.37$\pm$0.04&	15.53$\pm$0.12&	15.53$\pm$0.12&	15.57$\pm$0.12&	24.63$\pm$0.41\\
& $\beta$=0.1& \cellcolor{green!25}28.8$\pm$0.70 &17.47$\pm$0.05&	19.33$\pm$0.00&	19.30$\pm$0.01&	19.70$\pm$0.06&	26.63$\pm$0.42\\
&$\beta$=0.3 & \cellcolor{green!25}45.03$\pm$0.17 &27.83$\pm$0.04&	26.53$\pm$0.05&	26.90$\pm$0.00&	27.57$\pm$0.01&	38.27$\pm$4.74\\
& $\beta$=0.5& \cellcolor{green!25}48.00$\pm$0.21 &30.80$\pm$0.19&	32.20$\pm$0.33&	30.07$\pm$0.21&	30.97$\pm$0.14&	43.33$\pm$0.08 \\
&$\beta$=1.0& \cellcolor{green!25}62.23$\pm$0.05 &49.07$\pm$0.25&	47.83$\pm$0.32&	48.03$\pm$0.25&	48.53$\pm$0.10&	60.03$\pm$0.55 \\
&\#C=1 &\cellcolor{green!25} 16.23$\pm$0.23&9.83$\pm$0.03&	17.07$\pm$0.11&	16.60$\pm$0.12&	15.50$\pm$0.12&	7.70$\pm$0.03 \\
&\#C=2  &\cellcolor{green!25}27.87$\pm$0.49 &11.83$\pm$0.32&	20.57$\pm$0.02&	19.17$\pm$0.01&	16.00$\pm$0.27&	18.53$\pm$0.57\\
&\#C=3 & \cellcolor{green!25}42.97$\pm$0.02 &14.10$\pm$0.12&	23.70$\pm$0.20&	25.80$\pm$0.39&	23.37$\pm$0.00&	36.73$\pm$0.07 \\
\bottomrule[0.1em] 
\end{tabular}
\end{table}


\begin{table}[!h]\tiny
\centering
\caption{ Comparison with various FL algorithms in one round with 50 local epochs settings.}
\label{table:50}
\setlength\tabcolsep{3.7mm}
\begin{tabular}{c|c|c|c|c|c|c|c}
\toprule[0.15em]
Dataset&Partition & FedLPA & FedNova & SCAFFOLD & FedAvg & FedProx &DENSE \\
\midrule[0.1em]
\multirow{9}{*}{FMNIST} &$\beta$=0.01 & \cellcolor{green!25}19.33$\pm$0.43 &10.13$\pm$0.00&	15.87$\pm$0.16&	18.53$\pm$0.35&	12.97$\pm$0.15 &10.70$\pm$0.01\\
&$\beta$=0.05  & \cellcolor{green!25} 32.70$\pm$0.40&19.47$\pm$0.53&	24.10$\pm$0.02&	23.93$\pm$0.11&	22.63$\pm$0.20 &31.33$\pm$1.34\\
& $\beta$=0.1& \cellcolor{green!25}40.00$\pm$0.01 &30.40$\pm$1.05&	27.37$\pm$0.23&	25.83$\pm$0.35&	25.50$\pm$0.72 & 39.93$\pm$1.10\\
&$\beta$=0.3 & \cellcolor{green!25}62.80$\pm$0.41 &43.67$\pm$0.01&	42.50$\pm$0.09&	41.50$\pm$0.10&	42.23$\pm$0.11 &57.80$\pm$0.04\\
& $\beta$=0.5& \cellcolor{green!25}68.27$\pm$0.00 &55.97$\pm$0.23&	55.27$\pm$0.38&	53.95$\pm$0.27&	55.00$\pm$0.26 &63.50$\pm$0.11\\
&$\beta$=1.0& \cellcolor{green!25}73.47$\pm$0.27 &61.20$\pm$0.09&	60.67$\pm$0.19&	60.77$\pm$0.12&	61.40$\pm$0.15 &66.03$\pm$0.06\\
&\#C=1 &\cellcolor{green!25}13.30$\pm$0.03 &10.50$\pm$0.00&	11.03$\pm$0.02&	10.50$\pm$0.00&	11.87$\pm$0.07 &10.00$\pm$0.00\\
&\#C=2 &\cellcolor{red!25}27.60$\pm$0.02 &16.37$\pm$0.08&	23.00$\pm$0.12&	18.37$\pm$0.38&	18.60$\pm$0.32&\cellcolor{green!25}29.33$\pm$0.44 \\
&\#C=3 & \cellcolor{green!25}38.13$\pm$0.01 &25.47$\pm$0.02&	25.23$\pm$0.29&	26.70$\pm$0.05&	26.40$\pm$0.18 &37.53$\pm$0.17 \\
\midrule[0.1em]
\multirow{9}{*}{CIFAR-10} &$\beta$=0.01 & \cellcolor{green!25}16.23$\pm$0.01 &10.23$\pm$0.00&	12.27$\pm$0.07&	13.07$\pm$0.08&	12.17$\pm$0.09 &10.33$\pm$0.00 \\
&$\beta$=0.05  & \cellcolor{green!25} 17.93$\pm$0.06&11.00$\pm$0.02&	10.33$\pm$0.00&	10.13$\pm$0.00&	11.20$\pm$0.03 &7.63$\pm$0.01 \\
& $\beta$=0.1& \cellcolor{green!25} 19.20$\pm$0.02&13.17$\pm$0.09&	13.63$\pm$0.21&	12.00$\pm$0.03&	12.43$\pm$0.06 &19.13$\pm$0.09\\
&$\beta$=0.3 & \cellcolor{green!25} 27.57$\pm$0.03&12.53$\pm$0.05&	12.33$\pm$0.02&	11.93$\pm$0.01&	12.90$\pm$0.01 &26.03$\pm$0.52 \\
& $\beta$=0.5& \cellcolor{green!25} 27.57$\pm$0.29&13.47$\pm$0.03&	12.30$\pm$0.00&	12.47$\pm$0.02&	13.47$\pm$0.02 &26.40$\pm$0.23 \\
&$\beta$=1.0& \cellcolor{green!25}30.27$\pm$0.02 &15.47$\pm$0.12&	15.30$\pm$0.14&	15.23$\pm$0.13&	15.53$\pm$0.10 &29.17$\pm$2.06\\
&\#C=1 &\cellcolor{green!25}10.90$\pm$0.00 &10.30$\pm$0.00&	10.30$\pm$0.00&	10.30$\pm$0.00&	10.33$\pm$0.00 &10.00$\pm$0.00 \\
&\#C=2 &\cellcolor{green!25}21.17$\pm$0.06 &10.13$\pm$0.00&	11.93$\pm$0.02&	10.57$\pm$0.01&	11.27$\pm$0.01& 15.87$\pm$0.08 \\
&\#C=3 & \cellcolor{green!25}23.80$\pm$0.01 &12.00$\pm$0.06&	12.07$\pm$0.02&	12.97$\pm$0.04&	11.90$\pm$0.02 &21.53$\pm$0.29 \\
\midrule[0.1em]
\multirow{9}{*}{MNIST}  &$\beta$=0.01 & \cellcolor{green!25}34.10$\pm$0.88& 10.57$\pm$0.02&	9.50$\pm$0.00&9.33$\pm$0.02&	10.13$\pm$0.01& 12.53$\pm$0.19 \\
&$\beta$=0.05  & \cellcolor{green!25}66.23$\pm$0.32 &32.00$\pm$0.78&	39.70$\pm$0.50&	39.60$\pm$0.31&	39.87$\pm$0.15& 56.63$\pm$0.65\\
& $\beta$=0.1& \cellcolor{green!25}72.90$\pm$0.27 &49.17$\pm$0.62&	47.20$\pm$0.22&	47.07$\pm$0.23&	46.30$\pm$0.10 &69.93$\pm$0.27 \\
&$\beta$=0.3 & \cellcolor{green!25} 87.03$\pm$0.02&68.30$\pm$0.33&	66.40$\pm$0.16&	67.10$\pm$0.09&	66.17$\pm$0.22& 82.47$\pm$0.01\\
& $\beta$=0.5& \cellcolor{green!25}90.43$\pm$0.07 &80.70$\pm$0.13&	78.13$\pm$0.19&	79.37$\pm$0.19&	79.50$\pm$0.22& 88.30$\pm$0.05 \\
&$\beta$=1.0& \cellcolor{green!25} 94.47$\pm$0.04&86.73$\pm$0.15&	85.43$\pm$0.11&	86.07$\pm$0.12&	86.20$\pm$0.16 &89.23$\pm$0.01 \\
&\#C=1 &\cellcolor{green!25} 11.37$\pm$0.01&10.17$\pm$0.01&	10.13$\pm$0.01&	10.13$\pm$0.01&	10.10$\pm$0.01 &9.80$\pm$0.00 \\
&\#C=2  &\cellcolor{green!25} 71.07$\pm$0.02&23.53$\pm$1.00&	22.93$\pm$0.80&	22.63$\pm$0.99&	17.53$\pm$0.12& 42.97$\pm$0.17 \\
&\#C=3 & \cellcolor{green!25} 76.17$\pm$0.38&29.60$\pm$1.80&	33.50$\pm$2.91&	32.77$\pm$2.27&	23.40$\pm$3.38 &57.30$\pm$0.12 \\
\midrule[0.1em]
\multirow{9}{*}{SVHN} &$\beta$=0.01 & \cellcolor{green!25}19.60$\pm$0.00	&13.93$\pm$0.16&	13.57$\pm$0.19	&9.50$\pm$0.03&	9.27$\pm$0.02&19.10$\pm$0.52 \\
&$\beta$=0.05  & \cellcolor{green!25}22.97$\pm$0.01 &14.87$\pm$0.12&	15.83$\pm$0.13&	15.67$\pm$0.12&	14.67$\pm$0.14& 19.97$\pm$0.54 \\
& $\beta$=0.1& \cellcolor{green!25}45.83$\pm$1.70 &22.40$\pm$0.21&	22.97$\pm$0.12&	22.47$\pm$0.01&	24.30$\pm$0.04 &41.47$\pm$4.63 \\
&$\beta$=0.3 & \cellcolor{green!25} 36.30$\pm$2.02&33.90$\pm$0.15&	33.87$\pm$0.20&	33.50$\pm$0.13&	34.43$\pm$0.21& 29.90$\pm$0.56 \\
& $\beta$=0.5& \cellcolor{green!25} 51.77$\pm$0.01&39.70$\pm$0.19&	39.93$\pm$0.13&	38.03$\pm$0.14&	38.33$\pm$0.12 &50.10$\pm$1.71\\
&$\beta$=1.0& \cellcolor{green!25}57.97$\pm$0.10 &56.70$\pm$0.15&	54.03$\pm$0.28&	55.33$\pm$0.23&	55.80$\pm$0.15 &47.80+8.91 \\
&\#C=1 &\cellcolor{green!25} 19.37$\pm$0.00&9.90$\pm$0.03&	16.57$\pm$0.12&	16.53$\pm$0.12&	15.53$\pm$0.12 &10.00$\pm$0.00 \\
&\#C=2  &\cellcolor{green!25}36.93$\pm$0.02 &12.53$\pm$0.25&	20.30$\pm$0.07&	20.70$\pm$0.12&	15.57$\pm$0.22& 40.77$\pm$2.21 \\
&\#C=3 & \cellcolor{green!25} 42.43$\pm$0.02&21.07$\pm$0.31&	29.63$\pm$0.16&	27.10$\pm$0.06&	24.73$\pm$0.00 &38.50$\pm$0.60 \\
\bottomrule[0.1em]
\end{tabular}
\end{table}

\begin{table}[!h]\tiny
\centering
\caption{ Comparison with various FL algorithms in one round with 100 local epochs settings.}
\label{table:100}
\setlength\tabcolsep{3.7mm}
\begin{tabular}{c|c|c|c|c|c|c|c}
\toprule[0.15em]
Dataset&Partition & FedLPA & FedNova & SCAFFOLD & FedAvg & FedProx &DENSE \\
\midrule[0.1em]
\multirow{9}{*}{FMNIST} &$\beta$=0.01 & \cellcolor{green!25} 19.17$\pm$0.01&	11.73$\pm$0.06&	16.10$\pm$0.18&	19.00$\pm$0.27&	12.67$\pm$0.12	&10.07$\pm$0.00 \\
&$\beta$=0.05  & \cellcolor{green!25} 36.77$\pm$0.51&	18.07$\pm$0.35&	22.67$\pm$0.03&	22.73$\pm$0.04&	21.20$\pm$0.17&31.77$\pm$1.14 \\
& $\beta$=0.1& \cellcolor{green!25}35.90$\pm$0.12	&32.83$\pm$0.58&	29.87$\pm$0.49&	30.80$\pm$0.28&	29.33$\pm$0.60&33.23$\pm$1.22 \\
&$\beta$=0.3 & \cellcolor{green!25} 64.07$\pm$0.28&	47.77$\pm$0.07&	42.20$\pm$0.01&	43.33$\pm$0.06&	46.03$\pm$0.06&60.30$\pm$0.17\\
& $\beta$=0.5& \cellcolor{green!25}68.73$\pm$0.10&	57.03$\pm$0.20&	55.87$\pm$0.38&	56.10$\pm$0.31&	58.60$\pm$0.43&64.60$\pm$0.01  \\
&$\beta$=1.0& \cellcolor{green!25} 76.27$\pm$0.00&	65.00$\pm$0.05&	61.67$\pm$0.35&	65.13$\pm$0.14&	65.03$\pm$0.14&75.80$\pm$0.05 \\
&\#C=1 &\cellcolor{green!25}13.37$\pm$0.04&	10.87$\pm$0.02&	10.47$\pm$0.00&	10.87$\pm$0.02&	13.23$\pm$0.21&10.00$\pm$0.00 \\
&\#C=2 &\cellcolor{green!25} 31.40$\pm$1.16&	20.93$\pm$0.23&	24.97$\pm$0.19&	23.13$\pm$0.25&	21.50$\pm$0.29&26.30$\pm$1.56 \\
&\#C=3 & \cellcolor{green!25} 49.73$\pm$0.24&	26.97$\pm$0.00&	25.57$\pm$0.27&	26.17$\pm$0.22&	25.50$\pm$0.12&46.87$\pm$0.10 \\
\midrule[0.1em]
\multirow{9}{*}{CIFAR-10} &$\beta$=0.01 & \cellcolor{green!25} 16.93$\pm$0.01&	10.33$\pm$0.00&	10.97$\pm$0.02&	9.57$\pm$0.41&	11.10$\pm$0.02&11.23$\pm$0.01\\
&$\beta$=0.05  & \cellcolor{green!25} 19.07$\pm$0.01&	12.33$\pm$0.11&	12.50$\pm$0.12&	10.33$\pm$0.00&	12.60$\pm$0.13&18.63$\pm$0.11\\
& $\beta$=0.1& \cellcolor{green!25}20.80$\pm$0.08&	12.53$\pm$0.05&	10.33$\pm$0.00&	10.67$\pm$0.00&	11.87$\pm$0.03&24.30$\pm$0.05  \\
&$\beta$=0.3 & \cellcolor{green!25}28.33$\pm$0.00&	11.63$\pm$0.02&	11.03$\pm$0.01&	11.07$\pm$0.00&	11.70$\pm$0.01&28.23$\pm$0.36 \\
& $\beta$=0.5& \cellcolor{green!25} 29.37$\pm$0.01&	12.07$\pm$0.01&	12.13$\pm$0.01&	11.80$\pm$0.01&	13.17$\pm$0.01&28.90$\pm$0.49 \\
&$\beta$=1.0& \cellcolor{green!25} 30.57$\pm$0.00&	14.53$\pm$0.09&	13.93$\pm$0.01&	13.97$\pm$0.10&	15.93$\pm$0.11&29.37$\pm$1.73 \\
&\#C=1 &\cellcolor{green!25}11.03$\pm$0.02&	10.23$\pm$0.00&	10.23$\pm$0.00&	10.23$\pm$0.00&	10.57$\pm$0.01&10.00$\pm$0.00 \\
&\#C=2 &\cellcolor{green!25} 16.70$\pm$0.13	&10.00$\pm$0.00&	12.90$\pm$0.03&	11.00$\pm$0.01&	11.97$\pm$0.03&13.67$\pm$0.03 \\
&\#C=3 & \cellcolor{green!25}18.87$\pm$0.01&	11.33$\pm$0.03&	10.70$\pm$0.00&	11.77$\pm$0.02&	11.67$\pm$0.02&15.97$\pm$0.10 \\
\midrule[0.1em]
\multirow{9}{*}{MNIST}  &$\beta$=0.01 & \cellcolor{green!25}34.10$\pm$0.66&	13.60$\pm$0.32&	9.33$\pm$0.00&	9.30$\pm$0.00&	9.30$\pm$0.00&16.63$\pm$0.33 \\
&$\beta$=0.05  & \cellcolor{green!25}72.47$\pm$0.07&	32.30$\pm$0.66&	41.37$\pm$0.37&	38.57$\pm$0.35&	40.70$\pm$0.49&55.30$\pm$1.88\\
& $\beta$=0.1& \cellcolor{green!25}78.53$\pm$0.20&	48.20$\pm$0.39&	47.87$\pm$0.26&	47.57$\pm$0.19&	46.93$\pm$0.04&76.47$\pm$0.20 \\
&$\beta$=0.3 & \cellcolor{green!25}85.83$\pm$0.04&	68.77$\pm$0.28&	67.43$\pm$0.11&	67.13$\pm$0.12&	65.67$\pm$0.36&84.23$\pm$0.08  \\
& $\beta$=0.5& \cellcolor{green!25} 89.03$\pm$0.12&	80.53$\pm$0.19&	79.13$\pm$0.23&	79.00$\pm$0.28&	79.50$\pm$0.30&88.30$\pm$0.31\\
&$\beta$=1.0& \cellcolor{green!25}94.13$\pm$0.03&	86.53$\pm$0.09&	85.87$\pm$0.09&	85.63$\pm$0.08&	86.17$\pm$0.14&92.57$\pm$0.02 \\
&\#C=1 &\cellcolor{green!25}11.27$\pm$0.01&	10.30$\pm$0.02&	10.10$\pm$0.01&	10.10$\pm$0.01&	10.13$\pm$0.01&9.93$\pm$0.00  \\
&\#C=2  &\cellcolor{green!25}71.07$\pm$0.35&	21.00$\pm$0.61&	22.47$\pm$0.89&	18.83$\pm$0.55&	14.50$\pm$0.12&45.47$\pm$0.14\\
&\#C=3 & \cellcolor{green!25} 76.83$\pm$0.32&	29.63$\pm$2.43&	35.17$\pm$2.54&	32.47$\pm$3.15&	29.2$\pm$2.22&67.33$\pm$0.95 \\
\midrule[0.1em]
\multirow{9}{*}{SVHN} &$\beta$=0.01 & \cellcolor{green!25} 19.50$\pm$0.00&	13.90$\pm$0.16&	9.37$\pm$0.02&	12.57$\pm$0.22&	11.60$\pm$0.09&19.10$\pm$0.13 \\
&$\beta$=0.05  & \cellcolor{green!25} 32.90$\pm$0.05&	13.50$\pm$0.03&	16.03$\pm$0.15&	15.90$\pm$0.18&	16.83$\pm$0.10&25.80$\pm$1.64\\
& $\beta$=0.1& \cellcolor{green!25} 36.63$\pm$0.27&	22.37$\pm$0.62&	24.17$\pm$0.20&	24.83$\pm$0.07&	25.93$\pm$0.09&26.97$\pm$0.23\\
&$\beta$=0.3 & \cellcolor{green!25} 56.40$\pm$0.01&	35.43$\pm$0.10&	34.40$\pm$0.05&	35.17$\pm$0.10&	34.40$\pm$0.07&55.67$\pm$1.85 \\
& $\beta$=0.5& \cellcolor{green!25}55.63$\pm$0.16&	39.07$\pm$0.03&	40.33$\pm$0.05&	37.47$\pm$0.01&	37.07$\pm$0.12&55.53$\pm$0.62 \\
&$\beta$=1.0& \cellcolor{green!25}65.57$\pm$0.01&	55.87$\pm$0.27&	55.30$\pm$0.26&	54.80$\pm$0.19&	54.17$\pm$0.34&62.50$\pm$0.12  \\
&\#C=1 &\cellcolor{green!25} 16.27$\pm$0.22&	10.33$\pm$0.00&	13.83$\pm$0.17&	15.67$\pm$0.11&	15.63$\pm$0.11&12.10$\pm$0.30\\
&\#C=2  &\cellcolor{green!25}  41.87$\pm$0.01&	14.80$\pm$0.12&	22.53$\pm$0.07&	20.77$\pm$0.06&	13.87$\pm$0.86&41.43$\pm$1.77\\
&\#C=3 & \cellcolor{green!25} 48.70$\pm$0.02&	23.50$\pm$0.04&	30.20$\pm$0.08&	29.20$\pm$0.10&	25.30$\pm$0.02&48.60$\pm$0.49 \\
\bottomrule[0.1em]
\end{tabular}
\end{table}

Here, we present experiments similar to those in Table~\ref{table:main} but with different numbers of epochs (10, 20, 50, 100). The performance of our methods outperforms other approaches, as shown in Table~\ref{table:10}, Table~\ref{table:20}, Table~\ref{table:50}, and Table~\ref{table:100}. Without tuning the number of local epochs, our method consistently achieves high performance compared to other baselines. In almost all the settings, our method can outperform the state-of-the-art baseline approach DENSE. We also note that DENSE consumes more computing resources, as shown in Table~\ref{table:runcom}. Besides, it needs an extra data generation stage and an extra model distillation stage. Our method could get better results and consume fewer resources. What’s more, in Section~\ref{sec:extend}, we also show that our method has the potential to extend to multiple-round settings, while it is hard to extend the DENSE into multi-round settings.

\subsection{Extreme setting, 5 clients}

\begin{table}[ht]\scriptsize
\centering
\caption{ Comparison with various FL algorithms in one round when client number is 5.}
\label{table:client5}
\setlength\tabcolsep{3.0mm}
\begin{tabular}{c|c|c|c|c|c|c|c}
\toprule[0.15em]
Dataset&Partition & FedLPA & FedNova & SCAFFOLD & FedAvg & FedProx &DENSE \\
\midrule[0.1em]
\multirow{6}{*}{FMNIST} &$\beta$=0.01 & \cellcolor{green!25} 48.13$\pm$0.28&26.03$\pm$0.07&30.77$\pm$0.49 & 30.80$\pm$0.34& 17.83$\pm$0.07& 44.23$\pm$0.14 \\
&$\beta$=0.05  & \cellcolor{green!25} 55.20$\pm$0.17&23.40$\pm$0.16& 30.80$\pm$0.67&29.90$\pm$0.12 &20.43$\pm$0.16 & 46.17$\pm$0.09 \\
& $\beta$=0.1  & \cellcolor{green!25}59.27$\pm$0.12 &33.47$\pm$0.16&37.77$\pm$0.45& 35.43$\pm$0.86& 32.57$\pm$0.98& 58.73$\pm$0.15 \\
&$\beta$=0.3  & \cellcolor{green!25}73.13$\pm$0.00 &53.13$\pm$0.42&52.57$\pm$0.46 &52.03$\pm$0.59 & 49.90$\pm$0.33& 63.40$\pm$0.06 \\
& $\beta$=0.5 & \cellcolor{green!25}74.17$\pm$0.02 &60.27$\pm$0.53& 60.13$\pm$0.57& 59.97$\pm$1.14&61.67$\pm$0.35 &72.03$\pm$0.05  \\
&$\beta$=1.0& \cellcolor{green!25}75.30$\pm$0.00 &63.00$\pm$0.05&60.87$\pm$0.24 &62.63$\pm$0.05 & 60.37$\pm$0.01& 74.93$\pm$0.04 \\
\bottomrule[0.1em]
\end{tabular}
\end{table}

When the number of clients is set to 5, the experimental results for the FMNIST dataset are shown in Table~\ref{table:client5}. These results demonstrate that our framework performs well even in extreme situations when the number of clients is relatively small.

\subsection{Extreme setting, $\beta=0.001$}

\begin{table}[ht]\scriptsize
\centering
\caption{ Comparison with various FL algorithms in one round with different epoch numbers and $\beta=0.001$.} 
\label{table:0001}
\setlength\tabcolsep{2.5mm}
\begin{tabular}{c|c|c|c|c|c|c|c}
\toprule[0.15em]
Dataset&epochs number & FedLPA & FedNova & SCAFFOLD & FedAvg & FedProx &DENSE \\
\midrule[0.1em]
\multirow{5}{*}{FMNIST} &10 & \cellcolor{green!25}14.57$\pm$0.04 &10.60$\pm$0.01 & 10.53$\pm$0.01&10.60$\pm$0.01 & 13.10$\pm$0.01& 10.00$\pm$0.01 \\
&20 & \cellcolor{green!25}15.33$\pm$0.04 &10.13$\pm$0.00 & 10.23$\pm$0.00&10.13$\pm$0.00 &12.87$\pm$0.16 & 10.00$\pm$0.00 \\
&50 & \cellcolor{green!25}13.77$\pm$0.02 &10.57$\pm$0.01 &10.17$\pm$0.00 &10.57$\pm$0.01 &12.30$\pm$0.11 &10.00$\pm$0.00   \\
&100 & \cellcolor{green!25} 15.83$\pm$0.03&10.17$\pm$0.00 &10.73$\pm$0.01 &10.17$\pm$0.00 &13.23$\pm$0.21 & 10.00$\pm$0.00  \\
&200 & \cellcolor{green!25} 14.53$\pm$0.00& 10.07$\pm$0.00& 10.10$\pm$0.00& 10.07$\pm$0.00&12.50$\pm$0.12 & 10.00$\pm$0.00  \\
\midrule[0.1em]
\multirow{5}{*}{CIFAR-10}  &10 & \cellcolor{green!25}11.50$\pm$0.00 &10.27$\pm$0.00 &10.17$\pm$0.00 & 10.27$\pm$0.00&10.33$\pm$0.00 &10.00$\pm$0.00   \\
&20 & \cellcolor{green!25}10.57$\pm$0.01 &10.27$\pm$0.00 &10.13$\pm$0.00 & 10.27$\pm$0.00& 10.30$\pm$0.00& 10.00$\pm$0.00 \\
&50 & \cellcolor{green!25}10.77$\pm$0.01 &10.23$\pm$0.00 &10.33$\pm$0.00 &10.23$\pm$0.00 &10.33$\pm$0.00 & 10.00$\pm$0.00 \\
&100 & \cellcolor{green!25}10.90$\pm$0.01 & 10.20$\pm$0.00& 10.30$\pm$0.00&10.23$\pm$0.00 &10.57$\pm$0.01 &10.00$\pm$0.00  \\
&200 & \cellcolor{green!25}10.87$\pm$0.02 &10.27$\pm$0.00 & 10.23$\pm$0.00&10.27$\pm$0.00 &10.37$\pm$0.01 & 10.00$\pm$0.00 \\
\midrule[0.1em]
\multirow{5}{*}{MNIST}&10 & \cellcolor{green!25}24.10$\pm$0.17 &10.07$\pm$0.01 &12.17$\pm$0.07 &11.83$\pm$0.05 & 12.17$\pm$0.12&  9.90$\pm$0.00\\
&20 & \cellcolor{green!25}19.53$\pm$0.33 & 10.07$\pm$0.01&12.07$\pm$0.07&13.37$\pm$0.08 &12.37$\pm$0.12 &9.27$\pm$0.00  \\
&50 & \cellcolor{green!25}16.93$\pm$0.37 &10.07$\pm$0.01 & 10.80$\pm$0.04&13.17$\pm$0.09 & 13.13$\pm$0.25&  11.40$\pm$0.08\\
&100 & \cellcolor{green!25} 19.07$\pm$0.41& 10.13$\pm$0.01& 10.97$\pm$0.00&11.37$\pm$0.02 &12.90$\pm$0.13 &12.83$\pm$0.17  \\
&200 & \cellcolor{green!25}15.63$\pm$0.03 & 10.07$\pm$0.01& 11.13$\pm$0.06& 12.50$\pm$0.04& 11.83$\pm$0.11&  9.27$\pm$0.00\\
\midrule[0.1em]
\multirow{5}{*}{SVHN}  &10 & \cellcolor{green!25}17.50$\pm$0.02 &15.90$\pm$0.00 & 15.53$\pm$0.12& 15.53$\pm$0.12&15.53$\pm$0.12 & 17.13$\pm$0.03 \\
&20 & \cellcolor{green!25} 20.10$\pm$0.21&15.90$\pm$0.00 &15.53$\pm$0.12 &15.53$\pm$0.12 &14.00$\pm$0.11 & 17.13$\pm$0.03 \\
&50 & \cellcolor{green!25}20.07$\pm$0.71 &16.30$\pm$0.00 & 15.50$\pm$0.12&15.13$\pm$0.16 & 14.03$\pm$0.07&15.17$\pm$0.16  \\
&100 & \cellcolor{green!25} 19.70$\pm$0.00&15.90$\pm$0.00 &15.10$\pm$0.16 &15.53$\pm$0.12 &13.77$\pm$0.10 & 18.47$\pm$0.05 \\
&200 & \cellcolor{green!25}19.13$\pm$0.00 &13.90$\pm$0.11 &14.90$\pm$0.19 &15.13$\pm$0.16 & 13.27$\pm$0.06& 15.23$\pm$0.16 \\
\bottomrule[0.1em]
\end{tabular}
\end{table}

Here, we demonstrate that even when $\beta=0.001$ and with different dataset and local epoch number settings, FedLPA has the potential to aggregate models effectively in extreme situations and produce superior results. These results are presented in Table~\ref{table:0001}.



\subsection{Experiments with FedOV and Co-Boosting}
\label{append:fedovcoboosting}

We compare with FedOV\footnote{https://github.com/Xtra-Computing/FedOV}, the state-of-the-art method which addresses label skews in one-shot federated learning. We run the experiments with fair comparison (same model size) on MNIST dataset with \#C=2 partition setting. Table~\ref{table:fedov} shows that our method could be comparable with FedOV in some scenarios even when FedOV transmits the unknown label information to the clients and utilizes the knowledge distillation. As the epoch number of local clients equals to 50,100,200, FedLPA outperforms FedOV.

We also compare with Co-Boosting\footnote{https://github.com/rong-dai/Co-Boosting}, the state-of-the-art distillation method.  We run the experiments with fair comparison (same model size) on the FMNIST dataset, and the rest of the settings are the same as the default. The results in Table~\ref{table:coboosting} show that when the $\beta$ is smaller than 0.1, our method outperforms Co-Boosting. Thus, with the increment of skewness, FedLPA shows significantly superior results.

In conclusion, our method could be comparable with FedOV and Co-Boosting in some settings, even when they consume more computational resources as shown in Appendix~\ref{append:run}.

\begin{table}
\centering
\caption{Comparison with FedOV on MNIST with \#C=2.}
\label{table:fedov}
\setlength\tabcolsep{2.7mm}
\begin{tabular}{c|c|c|c|c|c}
\toprule[0.15em]
epoch number &10&20&50&100&200\\
\midrule[0.1em]
FedLPA & 47.93$\pm$0.89 & 53.37$\pm$0.61 & \cellcolor{green!25}71.07$\pm$0.02 & \cellcolor{green!25} 71.07$\pm$0.35 & \cellcolor{green!25} 69.63$\pm$0.29 \\
\hline
FedOV &  \cellcolor{green!25}71.0$\pm$0.25 & \cellcolor{green!25} 70.27$\pm$0.39 & 69.23$\pm$0.31 & 65.83$\pm$0.23 & 64.50$\pm$0.38 \\
\bottomrule[0.1em]
\end{tabular}
\end{table}

\begin{table}
\centering
\caption{Comparison with Co-Boosting on FMNIST.}
\label{table:coboosting}
\setlength\tabcolsep{2.7mm}
\begin{tabular}{c|c|c|c|c|c}
\toprule[0.15em]
$\beta$ &0.01&0.05&0.1&0.3&0.5\\
\midrule[0.1em]
FedLPA & \cellcolor{green!25}21.20±0.67  &\cellcolor{green!25} 54.27±0.38 & 55.33±0.06  & 68.20±0.04 & 73.33±0.06\\
\hline
Co-Boosting &  17.31$\pm$0.24 &  48.97$\pm$1.44 & \cellcolor{green!25} 73.15$\pm$1.86 & \cellcolor{green!25} 83.37$\pm$0.44 &\cellcolor{green!25} 86.21$\pm$0.31 \\
\bottomrule[0.1em]
\end{tabular}
\end{table}

\subsection{Communication overhead evaluation}
\label{append:com}
Table~\ref{table:runcom} shows the communication overhead evaluation of a simple CNN with 5 layers on CIFAR-10 dataset. The results are given based on the experiments. In this section, we will give a concrete example to show the details.

The communication bits are the number of bits that are transmitted between a server and a client in a directed communication. It reflects the communication efficiency of federated learning algorithms. Better algorithms should have lower communication bits. The default floating point precision is 32 bits in Pytorch. 

\textbf{A fully-connected neural network model example: }We use a fully-connected neural network model with architecture 784-256-64-10 as an example to show the calculation, which has $784 \cdot256+256+256 \cdot64+64+64 \cdot10+10=217930$ floating point numbers, which is $6973760$ bits or around $0.831$ MB.

For a single directed communication from a client to the server or vice versa, the cost for FedAvg, FedProx, FedNova, and DENSE is $0.831$ MB each. SCAFFOLD costs $1.662$ MB for the same communication, which is double the amount of the others.

For a single communication from a client to the server, our method requires additional upload of $\A_k$ and $\B_k$, which contain $785 \cdot785+256 \cdot256+257 \cdot257+64 \cdot64+65 \cdot65+10 \cdot10=756231$ floating point numbers in total. Note, as $\A_k$ and $\B_k$ are symmetric matrices, we only need to upload the upper triangular part of them, reducing the total to roughly $756231/2=378115.5$ floating point numbers as about $1.442$ MB. Therefore, our approach costs $2.272$ MB for the one directed communication, which is $2.734$ times as FedAvg, FedProx, and DENSE, and $1.367$ times as SCAFFOLD. 


\textbf{A CNN model example: } We use another example using CNN to show the communication overhead. For example, we have one model, the first layer is nn.Conv2d(1, 6, 5), means there are 3 input channels, 6 output channels, and a 5x5 kernel size; the second layer is nn.Conv2d(6, 8, 5), means there are 6 input channels, 8 output channels, and a 5x5 kernel size.
 
The parameter count for the first layer is 1x6x5x5+6=156. Note that $\A$ and $\B$ are both symmetric matrices. Thus, the additional parameters for $\A$ and $\B$ for each kernel would be 5x5, and estimating the covariance for biases without decomposition results in a size of 6x6. Therefore, the additional parameters for this layer are 201 ($\A_{k_1}$= 6x((5x5-5)/2+5), $\B_{k_1}$=6x((5x5-5)/2+5)+(6x6-6)/2+6. 

The parameter count for the second layer is 8x6x5x5+8=1208. Therefore, the additional parameters for this layer are 1476 ($\A_{k_2}$= 6x8x((5x5-5)/2+5),$\B_{k_2}$=6x8x((5x5-5)/2+5)+(8x8-8)/2+8). 

These two examples all follow the theory: the communicated parameters $\A$, $\B$ and $\M$ are approximately 2x of the number of all parameters in the model $\bm{\theta}$.

The communication overhead will increase linearly using FedLPA when we change the client numbers to 20 and 50.


However, as Figure~\ref{fig:multiple_round} demonstrates, to achieve the same performance as FedLPA, FedAvg, FedNova, SCAFFOLD, and FedProx require more communication rounds, resulting in a heavier data transfer burden on the system.

\subsection{Running time and computation overhead evaluation}
\label{append:run}

\begin{table}
\centering
\caption{Running time and computation overhead evaluation.}
\label{table:running}
\setlength\tabcolsep{3.5mm}
\resizebox{\linewidth}{!}{
\begin{tabular}{c|c|c|c|c|c}
\toprule[0.15em]
FedLPA & FedNova/SCAFFOLD/FedAvg & FedProx & DENSE & FedOV & Co-Boosting\\
\midrule[0.1em]
65mins & 50mins & 75mins & 400mins & 150mins & 700mins\\
\bottomrule[0.1em]
\end{tabular}
}
\end{table}

The running times of different algorithms, using a simple CNN on the CIFAR-10 dataset, are summarized in Table~\ref{table:running}. In this experiment, there are 10 clients, each running 200 local epochs with only one communication round. Our device is a single 2080Ti GPU. Compared to the state-of-the-art methods FedOV, DENSE and Co-Boosting, our method is efficient and slightly slower than the fastest algorithm. Notably, DENSE consumes almost 7 times the computational resources, as the knowledge distillation method is computationally intensive and resource-demanding. Co-Boosting even uses more time. It's important to note that while our method is efficient, it also yields almost always the best results. In our paper, we mainly adopt the most-cited non-IID FL benchmark (https://github.com/Xtra-Computing/NIID-Bench) to get a fair comparison of FedLPA and other baselines. The reason why the computation cost of FedProx is higher than FedAvg may be that the FedProx adds a $l_2$ regular term to make local updates around the global mode, which adds more computing overhead. Besides, using the original codebase (https://github.com/litian96/FedProx) from FedPorx also consumes more time than FedAvg and FedNova, under the above non-IID FL benchmark. 

 Our methods guarantee that the computation result overhead will increase almost linearly using FedLPA when we change the client numbers to 20 and 50.

\subsection{Experiments with more complex neural network structure}
\label{appendix:structure}
We do the experiments with FedLPA on the same experiment setting in the paper using the more complex network, ResNet-18~\cite{he2016deep}, with five random seeds on the CIFAR-10 dataset. We set the parameters with $\beta$=0.1, 0.3 and 0.5 with 10 clients. The results are shown in Table~\ref{table:resnet18}.  From the results, we could see that using ResNet-18, our method still gets better performance compared to other baselines.

We do the experiments with FedLPA on the same experiment setting in the paper using the more complex network, VGG-9~\cite{simonyan2014very}, with five random seeds on the FMNIST dataset. We set the parameters with $\beta$=0.1, 0.3 and 0.5 with 10 clients. The results are shown in Table~\ref{table:vgg9}.  From the results, we could see that using VGG-9, our method still performs better than other baselines.

Based on the results of ResNet-18 and VGG-9, Our method has the potential to be applied to more complex models.

\begin{table}
\centering
\caption{Experiments with ResNet-18.}
\label{table:resnet18}
\setlength\tabcolsep{3.5mm}
\resizebox{\linewidth}{!}{
\begin{tabular}{c|c|c|c|c|c|c}
\toprule[0.15em]
$\beta$  & FedLPA  & FedNova& SCAFFOLD & FedAvg&FedProx & Dense \\
\midrule[0.1em]
0.1 &  \cellcolor{green!25}23.62$\pm$0.51 &12.16$\pm$0.23& 10.07$\pm$0.04 & 13.87$\pm$0.26 & 12.04$\pm$0.16 & 21.45$\pm$0.60 \\
\hline
0.3 &  \cellcolor{green!25}27.43$\pm$0.04 & 11.75$\pm$0.09 &10.08$\pm$0.32& 10.01$\pm$0.07 & 12.97$\pm$0.17 & 27.10$\pm$0.25\\
\hline
0.5 &  \cellcolor{green!25}31.70$\pm$0.14 &  13.81$\pm$0.31 & 12.75$\pm$0.11& 10.61$\pm$0.21&  11.45$\pm$0.23 & 29.04$\pm$0.30\\
\bottomrule[0.1em]
\end{tabular}
}
\end{table}

\begin{table}
\centering
\caption{Experiments with VGG-9.}
\label{table:vgg9}
\setlength\tabcolsep{3.5mm}
\resizebox{\linewidth}{!}{
\begin{tabular}{c|c|c|c|c|c|c}
\toprule[0.15em]
$\beta$  & FedLPA  & FedNova& SCAFFOLD & FedAvg&FedProx & Dense \\
\midrule[0.1em]
0.1 &  \cellcolor{green!25}58.48$\pm$1.33 &28.77$\pm$2.03& 32.45$\pm$0.12 & 33.71$\pm$0.16 & 31.78$\pm$0.40 & 51.76$\pm$0.28 \\
\hline
0.3 &  \cellcolor{green!25}75.98$\pm$1.72 & 53.78$\pm$0.32 &55.00$\pm$1.07& 54.61$\pm$0.02 & 52.79$\pm$0.80 & 70.10$\pm$1.45\\
\hline
0.5 &  \cellcolor{green!25}79.02$\pm$0.81 &  62.31$\pm$0.90 & 61.75$\pm$0.34& 63.18$\pm$1.70&  62.55$\pm$0.17 & 76.20$\pm$1.10\\
\bottomrule[0.1em]
\end{tabular}
}
\end{table}

\subsection{Experiments with more complex datasets}
\label{appendix:dataset}
We do the experiments with FedLPA on the same experiment setting in the paper using ResNet-18 with five random seeds on the CIFAR-100~\cite{cifar10} dataset. The results are shown in Table~\ref{table:cifar100}. We can see that even with the complicated dataset CIFAR-100, our method could also get satisfactory results in the federated one-shot setting. Besides, we also have added the experiments on EMNST~\cite{cohen2017emnist} using simple-CNN with 10 clients and five random seeds. We do the experiments on EMNIST-mnist and EMNIST-letters. The results are shown in Table~\ref{table:emnist}.

In addition to these, we conduct experiments with Tiny-ImangeNet~\cite{Le2015TinyIV} with ResNet-18 with 10 clients and five random seeds. The results are shown in Table~\ref{table:tinyimagenet}.

\begin{table}
\centering
\caption{Experiments with CIFAR-100 using FedLPA.}
\label{table:cifar100}
\setlength\tabcolsep{3.5mm}
\begin{tabular}{c|c|c}
\toprule[0.15em]
$\beta$   & FedAvg&  FedLPA \\
\midrule[0.1em]
0.1	& 	1.31$\pm$0.05 &15.11$\pm$0.38 \\
0.3 & 1.75$\pm$0.10 & 	18.82$\pm$0.71\\
0.5& 1.38$\pm$0.11  &	21.77$\pm$0.03\\
\bottomrule[0.1em]
\end{tabular}
\end{table}

\begin{table}
\centering
\caption{Experimental with EMNIST.}
\label{table:emnist}
\setlength\tabcolsep{3.4mm}
\begin{tabular}{c|c|c|c}
\toprule[0.15em]
Dataset & Partitions & FedLPA &  FedAvg  \\
\midrule[0.1em]
\multirow{3}{*}{EMNIT-mnist (10 classes)} &$\beta$=0.1& 74.23$\pm$3.10 &57.63$\pm$2.30 \\
&$\beta$=0.3 & 86.55$\pm$0.24& 62.32$\pm$1.77 \\
& $\beta$=0.5& 91.75$\pm$0.26 & 82.71$\pm$0.96\\
\midrule[0.1em]
\multirow{3}{*}{EMNIT-letters (37 classes)} &$\beta$=0.1& 26.34$\pm$0.71 &16.22$\pm$0.38 \\
&$\beta$=0.3 &31.75$\pm$0.03 &25.51$\pm$0.44  \\
& $\beta$=0.5&33.78$\pm$0.14 &26.34$\pm$0.07  \\
\bottomrule[0.1em]
\end{tabular}
\end{table}

\begin{table}
\centering
\caption{Experiments with Tiny-ImangeNet using ResNet-18.}
\label{table:tinyimagenet}
\setlength\tabcolsep{3.5mm}
\begin{tabular}{c|c|c|c}
\toprule[0.15em]
$\beta$   & FedLPA& Dense &FedAvg \\
\midrule[0.1em]
0.1	& 	17.02$\pm$1.40&15.88$\pm$1.96 &3.72$\pm$1.44 \\
0.3 & 27.80$\pm$2.10 &24.91 $\pm$1.65	&8.41$\pm$0.87\\
0.5& 30.14$\pm$1.25  &	29.43$\pm$0.72 &12.07 $\pm$1.92\\
\bottomrule[0.1em]
\end{tabular}
\end{table}

\subsection{Ablation experiments analyzing the number of approximation iterations of FedLPA}

The proposed method is composed of multiple approximations: 1) empirical Fisher to approximate the Hessian, 2) block-diagonal Fisher matrix instead of full, 3) approximating global model parameter $\bar{\M}$ with optimization problem in  Eq. \ref{EQ:MU:obj:solution}. 

1) Empirical Fisher to approximate the Hessian:

Although empirical Fisher has been successfully applied in many methods and yielded good results, discussions concerning the approximation error of empirical Fisher are limited. Fortunately, previous work~\cite{kunstner2019limitations} provides a detailed critical discussion of the empirical Fisher approximation.

\textbf{i. Fisher to approximate the Hessian:}

When the loss function represents an exponential family distribution, the Fisher is a well-justified approximation of the Hessian, and its approximation error can be bounded in terms of residuals. The accuracy of this approximation improves as the residuals diminish and is exact when the data is perfectly fitted.

\textbf{ii. Empirical Fisher to Fisher:}

It's noted that the Fisher and empirical Fisher coincide near minima of the loss function under two conditions:

A. The model distribution closely approximates the data distribution.

B. A sufficiently large number of samples allows both the Fisher and empirical Fisher to converge to their respective average values in the population.

In practical environments, especially condition 1, might not hold, causing bias between empirical Fisher and Fisher. However, empirical Fisher still contains effective covariance information. In second-order optimization methods, the covariance information in empirical Fisher can adapt to the gradient noise in stochastic optimization. Nevertheless, referencing the work~\cite{martens2016second}, we can use the model’s predictive distribution to obtain an unbiased estimate of the true Fisher at the same computational cost as empirical Fisher.

(2) Block-diagonal Fisher matrix to approximate the full one:
The work~\cite{martens2018kronecker} provides a detailed evaluation and testing of using block-diagonal Fisher to approximate the full one. Firstly, Chapter 6.3.1 ``Interpretations of this approximation" in the paper~\cite{martens2018kronecker} indicates that using a block-wise Kronecker-factored Fisher closely approximates the full Fisher. Although there is a bias term, this term approximates zero when there are sufficient samples. Furthermore, the paper examines the approximation quality of block-diagonal Fisher compared with the true Fisher and suggests that block-diagonal Fisher captures the main correlations, while the remaining correlations have a minimal impact on the experimental results.

(3) Besides, we have added some experiments for more ablation studies with our method on the same experiment setting in the paper with five random seeds with 10 clients. We conducted experiments on FMNIST datasets with $\beta$=0.1, 0.3 and 0.5. The results are shown in Table~\ref{table:approximation}. We show the experiments analyzing the number of approximation iterations. With the experiment results, we could know that 5000 iterations are enough to get the ideal results. By default, we use 10000 iterations. 

We also show that the computation time for the approximation is linear with the number of approximation iterations in the last column of Table~\ref{table:approximation}.

\begin{table}
\centering
\caption{Experiments for the approximation study.}
\label{table:approximation}
\setlength\tabcolsep{3.5mm}
\resizebox{\linewidth}{!}{
\begin{tabular}{c|c|c|c|c}
\toprule[0.15em]
Number of iterations	& Accuracy($\beta$=0.1)  & Accuracy($\beta$=0.3) & Accuracy($\beta$=0.5) & Computation(s) \\
\midrule[0.1em]
1000&	52.81$\pm$0.71&	60.31$\pm$0.23&	72.11$\pm$0.57 & 3\\
5000&	59.70$\pm$0.32&	68.09$\pm$0.30&	74.27$\pm$0.12 & 15\\
10000&	55.33$\pm$0.06&	68.20$\pm$0.04&	73.33$\pm$0.06 & 30\\
20000&	58.41$\pm$0.05&	68.11$\pm$0.07&	73.51$\pm$0.02& 60\\
\bottomrule[0.1em]
\end{tabular}
}
\end{table}


Additionally, it's worth noting that concerning Laplace approximation, the analysis~\cite{bilodeau2023tightness} suggests that the error of Laplace approximation is inversely proportional to the input dimension $n$ with $O(n^{-1})$. According to this conclusion, it can be inferred that in our method, for each layer of the neural network, the error of Laplace approximation is inversely proportional to its width. When the neural network is infinitely wide, the approximation error tends towards zero.

\subsection{Artifact details}
\label{appendx:exp:codebase}
We have uploaded the codebase containing all the methods compared in our paper. Setting up the environment is relatively straightforward with the provided readme file. If you refer to the scripts folder, you will find all the bash scripts necessary to reproduce the tables and figures from our experiments.

The experiments.sh script covers the experiments in Table~\ref{table:main}, Table~\ref{table:10}, Table~\ref{table:20}, Table~\ref{table:50}, and Table~\ref{table:100}. Running these experiments on a single 2080Ti GPU will take approximately 81 days. Specifically, Table~\ref{table:main} itself will take about 35 days.

The experiments\_client.sh script covers the experiments in Table~\ref{table:clients}, requiring approximately 40 days on a single 2080Ti GPU.

The experiments\_coor.sh script covers the experiments in Table~\ref{table:lambda}, which can be completed in 2 days.

The experiments\_dp.sh script covers the experiments in Table~\ref{table:dp_all}, requiring approximately 1 day on a single 2080Ti GPU.

The experiments\_fedavg\_with\_attack.py and experiments\_fedlpa\_with\_attack.py covers the experiments in Table~\ref{table:idlg} requiring approximately 1 day on a single 2080Ti GPU.

The experiments\_extreme\_clients.sh script covers the experiments in Table~\ref{table:client5} and requires approximately 4 days of GPU processing.

The experiments\_extreme.sh script reproduces the experiments in Table~\ref{table:0001} and takes about 10 days.

The experiments of Table~\ref{table:fedov} and Table~\ref{table:coboosting} take about 8 days.

The experiments\_emnist.sh script covers the experiments in Table~\ref{table:emnist} and takes about 1 day.

Running experiments\_multiple\_round.sh will yield the results as shown in Figure~\ref{fig:multiple_round}, and this process takes about 1 day.

The experiments for Table~\ref{table:compare1} and Table~\ref{table:compare} will take about 1 day. The experiments for Table~\ref{table:approximation} and Table~\ref{table:proportion} will also take about 1 day.

To generate the t-SNE visualizations shown in Figure~\ref{fig:tsne_our}, Figure~\ref{fig:tsne_c}, and Figure~\ref{fig:tsne_g}, you can use the experiments.py script with the ``alg=tsne" option.

In total, to reproduce all the experiment results in this paper, it will require about 185 days for GPU processing.


\clearpage
\section*{NeurIPS Paper Checklist}


\begin{enumerate}

\item {\bf Claims}
    \item[] Question: Do the main claims made in the abstract and introduction accurately reflect the paper's contributions and scope?
    \item[] Answer: \answerYes{}.
    \item[] Justification: The main claims made in the abstract and introduction accurately reflect the paper's contributions and scope.

\item {\bf Limitations}
    \item[] Question: Does the paper discuss the limitations of the work performed by the authors?
    \item[] Answer: \answerYes{}.
    \item[] Justification: The paper discusses the limitations of the work performed by the authors.

\item {\bf Theory Assumptions and Proofs}
    \item[] Question: For each theoretical result, does the paper provide the full set of assumptions and a complete (and correct) proof?
    \item[] Answer: \answerYes{}.
    \item[] Justification: The paper provides the full set of assumptions and a complete (and correct) proof. 

    \item {\bf Experimental Result Reproducibility}
    \item[] Question: Does the paper fully disclose all the information needed to reproduce the main experimental results of the paper to the extent that it affects the main claims and/or conclusions of the paper (regardless of whether the code and data are provided or not)?
    \item[] Answer: \answerYes{}.
    \item[] Justification: The paper fully discloses all the information needed to reproduce the main experimental results of the paper to the extent that it affects the main claims and/or conclusions of the paper.

\item {\bf Open access to data and code}
    \item[] Question: Does the paper provide open access to the data and code, with sufficient instructions to faithfully reproduce the main experimental results, as described in supplemental material?
    \item[] Answer: \answerYes{}.
    \item[] Justification:  The paper provides open access to the data and code, with sufficient instructions to faithfully reproduce the main experimental results, as described in supplemental material. 

\item {\bf Experimental Setting/Details}
    \item[] Question: Does the paper specify all the training and test details (e.g., data splits, hyperparameters, how they were chosen, type of optimizer, etc.) necessary to understand the results?
    \item[] Answer: \answerYes{}.
    \item[] Justification:  The paper specifies all the training and test details (e.g., data splits, hyperparameters, how they were chosen, type of optimizer, etc.) necessary to understand the results.

\item {\bf Experiment Statistical Significance}
    \item[] Question: Does the paper report error bars suitably and correctly defined or other appropriate information about the statistical significance of the experiments?
    \item[] Answer: \answerYes{}.
    \item[] Justification:  The results are accompanied by error bars, confidence intervals, or statistical significance tests, at least for the experiments that support the main claims of the paper.

\item {\bf Experiments Compute Resources}
    \item[] Question: For each experiment, does the paper provide sufficient information on the computer resources (type of compute workers, memory, time of execution) needed to reproduce the experiments?
    \item[] Answer:  \answerYes{}.
    \item[] Justification: For each experiment, the paper provides sufficient information on the computer resources (type of compute workers, memory, time of execution) needed to reproduce the experiments.
    
\item {\bf Code Of Ethics}
    \item[] Question: Does the research conducted in the paper conform, in every respect, with the NeurIPS Code of Ethics \url{https://neurips.cc/public/EthicsGuidelines}?
    \item[] Answer: \answerYes{}.
    \item[] Justification: The research conducted in the paper conforms, in every respect, with the NeurIPS Code of Ethics.

\item {\bf Broader Impacts}
    \item[] Question: Does the paper discuss both potential positive societal impacts and negative societal impacts of the work performed?
    \item[] Answer:   \answerNA{}.
    \item[] Justification: There is no societal impact of the work performed.
    
\item {\bf Safeguards}
    \item[] Question: Does the paper describe safeguards that have been put in place for responsible release of data or models that have a high risk for misuse (e.g., pretrained language models, image generators, or scraped datasets)?
    \item[] Answer: \answerNA{}.
    \item[] Justification: The paper poses no such risks.

\item {\bf Licenses for existing assets}
    \item[] Question: Are the creators or original owners of assets (e.g., code, data, models), used in the paper, properly credited and are the license and terms of use explicitly mentioned and properly respected?
    \item[] Answer: \answerNA{}.
    \item[] Justification: The paper does not use existing assets.

\item {\bf New Assets}
    \item[] Question: Are new assets introduced in the paper well documented and is the documentation provided alongside the assets?
    \item[] Answer:  \answerNA{}.
    \item[] Justification: The paper does not release new assets.

\item {\bf Crowdsourcing and Research with Human Subjects}
    \item[] Question: For crowdsourcing experiments and research with human subjects, does the paper include the full text of instructions given to participants and screenshots, if applicable, as well as details about compensation (if any)? 
    \item[] Answer: \answerNA{}.
    \item[] Justification: The paper does not involve crowdsourcing nor research with human subjects.

\item {\bf Institutional Review Board (IRB) Approvals or Equivalent for Research with Human Subjects}
    \item[] Question: Does the paper describe potential risks incurred by study participants, whether such risks were disclosed to the subjects, and whether Institutional Review Board (IRB) approvals (or an equivalent approval/review based on the requirements of your country or institution) were obtained?
    \item[] Answer:  \answerNA{}.
    \item[] Justification: The paper does not involve crowdsourcing nor research with human subjects.

\end{enumerate}

\end{document}